\documentclass{article}

\usepackage{arxiv}

\usepackage[utf8]{inputenc} 
\usepackage[T1]{fontenc}    
\usepackage{hyperref}       
\usepackage{url}            
\usepackage{booktabs}       
\usepackage{amsfonts}       
\usepackage{nicefrac}       
\usepackage{microtype}      
\usepackage{lipsum}
\usepackage{graphicx}
\graphicspath{ {./images/} }

\usepackage{latexsym}
\usepackage[linesnumbered,boxed,ruled,commentsnumbered]{algorithm2e}
\usepackage{multirow}
\usepackage{longtable}
\usepackage{graphicx}
\usepackage{lipsum}
\usepackage{tabularx}
\usepackage{algorithmic}
\usepackage{float}
\usepackage{pdflscape}
\usepackage{caption}
\usepackage{subcaption}
\usepackage{setspace}
\usepackage{amsmath}
\newcommand{\lastcon}{\item[\algorithmiclastcon]}

\newcommand{\algorithmiclastcon}{\textbf{Output:}}
\usepackage{hyperref}

\makeatletter
\let\normalsize\relax
\let\@currsize\normalsize
\makeatother

\usepackage{natbib}
 \bibpunct[, ]{(}{)}{,}{a}{}{,}%
\usepackage{ulem}
\usepackage{soul}
\usepackage{xcite}

\newtheorem{proposition}{Proposition}
\newtheorem{proof}{proof}

\allowdisplaybreaks

\title{An Intelligent End-to-End Neural Architecture Search Framework for Electricity Forecasting Model Development}

\author{
 Jin Yang \\
  School of Management\\
  Harbin Institute of Technology\\
  Harbin, Heilongjiang 150001, China \\
  \texttt{22B910011@stu.hit.edu.cn} \\
   \And
 Guangxin Jiang \\
  School of Management\\
  Harbin Institute of Technology\\
  Harbin, Heilongjiang 150001, China \\
  \texttt{gxjiang@hit.edu.cn} \\
  \And
 Yinan Wang \\
  Department of Industrial and Systems Engineering\\
  Rensselaer Polytechnic Institute\\
  Troy, NY 12180, U.S. \\
  \texttt{{wangy88}@rpi.edu} \\
  \And
 Ying Chen* \\
  School of Management\\
  Harbin Institute of Technology\\
  Harbin, Heilongjiang 150001, China \\
  \texttt{yingchen@hit.edu.cn(Corresponding author)} \\
}

\begin{document}

\maketitle

\begin{abstract}
    Recent years have witnessed exponential growth in developing deep learning (DL) models for time-series electricity forecasting in power systems. However, most of the proposed models are designed based on the designers' inherent knowledge and experience without elaborating on the suitability of the proposed neural architectures. Moreover, these models cannot be self-adjusted to dynamically changed data patterns due to the inflexible design of their structures. Although several recent studies have considered the application of the neural architecture search (NAS) technique for obtaining a network with an optimized structure in the electricity forecasting sector, their training process is computationally expensive and their search strategies are not flexible, indicating that the NAS application in this area is still at an infancy stage. In this study, we propose an intelligent automated architecture search (IAAS) framework for the development of time-series electricity forecasting models. The proposed framework contains three primary components, i.e., network function-preserving transformation operation, reinforcement learning (RL)-based network transformation control, and heuristic network screening, which aim to improve the search quality of a network structure. After conducting comprehensive experiments on two publicly-available electricity load datasets and two wind power datasets, we demonstrate that the proposed IAAS framework significantly outperforms the ten existing models or methods in terms of forecasting accuracy and stability. Finally, we perform an ablation experiment to showcase the importance of critical components in the proposed IAAS framework in improving forecasting accuracy.

\end{abstract}

\section{Introduction}\label{section:intro}
\subsection{Overview}
Electricity market has reshaped the electricity trade mode since the introduction of competitive market and deregulation processes during the early 1990s \citep{weron2014electricity}. As electricity is a tradeable commodity that cannot be stored on a large scale, modern electricity markets require a balance between electricity production and consumption in real time (\citealt{gan2020balancing}). Consequently, this requirement plays an integral role in maintaining the stable operations and regulations of power systems \citep{huang2021multistage}. However, there are several factors that affect this balance. 

From the production perspective, many renewable energy sources, such as wind and solar, are increasingly contributing to the power systems with rapid growth \citep{solaun2019climate}. According to the fuel report from 2021 International Energy Agency, the overall global renewable electricity is predicted to be over 4800 gigawatts (GW) in 2026, which is equivalent to the total power capacity of fossil fuels and nuclear combined in 2020. Compared to traditional energy resources, e.g., coal and gas, renewable energy is sustainable and clean (\citealt{munawer2018human} and \citealt{nyashina2020effects}). However, its uncertain and intermittent nature brings significant challenges to the smooth and secure operation of power systems (\citealt{chen2023novel} and \citealt{pryor2020climate}). From the consumption perspective, electricity load can be correlated with various patterns related to industrial activities and weather conditions (\citealt{chen2018short} and \citealt{jalali2021novel}). For instance, industrial and commercial consumers usually consume more electricity during daytime than at night and utilize more electricity during summer than in spring or autumn. Nonetheless, these generic electricity usage patterns on any given day are still full of uncertainties (\citealt{billings2022industrial} and \citealt{li2022day}) since most consumption behaviors are uncontrollable. To enhance the secure and reliable operation, electricity forecasting has become one of the most effective techniques to minimize these uncertainties in modern power systems \citep{sunar2019strategic}.   

Contrasting with numerous other commodities, electricity entails being consumed immediately after being generated \citep{huang2021multistage}, which brings high requisites in forecasting accuracy. As artificial intelligence (AI) techniques are demonstrating outstanding performance in predicting and capturing uncertainties, researchers have turned to utilizing advanced machine learning (ML) methods instead of traditional statistical time-series methods (e.g., autoregressive moving average and exponential smoothing) for the electricity forecasting model development. For example, \cite{hu2015short} used support vector machine (SVM) to forecast wind power generation; \cite{liu2019random} employed random forest (RF) method to predict solar power generation; \cite{ul2014optimization} exploited artificial neural network (ANN) to forecast load demand. Various other relevant research studies can be found in the literature, such as \cite{chen2013wind, lahouar2015day,shepero2018residential,srivastava2019solar}, etc. On the other hand, the remarkable success of deep learning (DL) can be witnessed in pattern recognition, object detection, sales forecasting, and other applications (e.g., \citealt{bi2022improving,hu2022shedr,liu2022poolnet+,zhang2021detecting,zhou2021temporal}). Therefore, we have seen a dramatic trend in using DL methods in recent years to develop forecasting models for power systems. For instance, in wind power forecasting, \cite{shahid2021novel} introduced a novel genetic long short-term memory (LSTM) network model, and \cite{xiong2022short} proposed DL models based on attention mechanism. In solar power forecasting, \cite{heo2021multi} introduced a multi-channel convolutional neural network (CNN) model, and \cite{agga2021short} proposed two models: one is CNN+LSTM model, and the other is the convolutional LSTM model. In load forecasting, \cite{chen2018short} introduced two deep residual network models, and  \cite{jalali2021novel} proposed an evolutionary-based deep CNN model. Compared to traditional ML methods (e.g., SVM and RF), DL models have more flexible structures. Consequently, these models are likely to capture the hidden patterns within data and then construct powerful forecasting models. 

The primary goal of this research study is to provide a robust and end-to-end framework that can flexibly self-adjust the deep neural network structures for adapting to various datasets, in order to produce high-quality forecasting models for power systems. There are two motivations behind our considerations. First, numerous prior research studies have designed model structures based on their knowledge or experience \citep{elsken2019neural}, especially in the domain of energy forecasting. For example, \cite{shahid2021novel} designed a three-layer LSTM model for wind power forecasting; \cite{heo2021multi} developed a four-layer multi-channel CNN model for solar power forecasting; \cite{jalali2021novel} presented a model containing two CNN layers, a pooling layer and a fully connected network (FCN) layer, for load forecasting. Even though these research studies have conducted abundant experiments to demonstrate the advantages of their proposed models, they fail to clarify the rationales for designing each specific model structure. As noted, a larger (or deeper) network would usually achieve a better performance than a smaller network \citep{yue2015beyond} without considering the overfitting issue. In other words, whether a particular forecasting model could be further improved by using more layers (e.g., CNN or LSTM) or including more units in each layer is still unknown. As described in \cite{ren2021comprehensive}, designing an optimal neural architecture only based on the inherent knowledge of human beings is problematic, since it is difficult for people to jump out of their thinking paradigms. The second motivation is that although there are some initial trials on designing network structures, all these methods either only optimize the network structure incrementally (incapable of pruning) or stack the newly added layers in a certain paradigm. For example, in \cite{jalali2021new}, a heuristic algorithm is used to optimize the number of CNN and LSTM layers following the fixed structure that CNN layers are built on the top of LSTM layers; in \cite{shahid2021novel}, a genetic algorithm is used to only optimize the number of neurons in LSTM networks without considering other network structures (e.g., CNN and RNN) in the forecasting model. Such a paradigm simplifies the optimization problem, while inevitably decreasing the flexibility of the network structure search. Many factors, such as climate conditions and industrial activities, will introduce significant uncertainties and intermittency to power systems, resulting in a dynamic pattern. Therefore, an automated neural architecture optimization algorithm is required to be more flexible for power systems such that it can adaptively generate high-quality and self-adjusted forecasting models.

\subsection{Literature Review}

Neural architecture search (NAS) is a technique that can help automate the architecture design of neural networks \citep{elsken2019neural}. In the computer vision area, numerous NAS techniques have been proposed, which include intermittent-aware NAS \citep{mendis2021intermittent}, instance-aware NAS \citep{cheng2020instanas}, and platform-aware NAS \citep{tan2019mnasnet}. These techniques have exhibited promising and competitive results on some public benchmark datasets. However, in the domain of time-series forecasting, only a few works have been reported. Limited examples include a study by \cite{pan2021autostg}, which designed a NAS method for predictions of air quality and traffic speed. Based on \cite{pan2021autostg}, \cite{chen2021scale} developed a scale-aware NAS (SNAS) method for multivariate time-series forecasting, where the gradient-based optimization method was utilized in the search process. Regarding electricity forecasting, \cite{khodayar2017rough} developed an NAS strategy based on the rough set theory \citep{pawlak1982rough} for short-term wind speed forecasting, and \cite{torres2019random} used a random model based on the NAS strategy for load forecasting. Nonetheless, both of these NAS strategies were regarded as unintelligent and inefficient in \cite{jalali2021new}, which then proposed an improved evolutionary whale optimization algorithm to optimize the neural architecture for wind power forecasting. Even though the latest NAS techniques in \cite{chen2021scale} and \cite{jalali2021new} have achieved some success in the time-series model development, two issues in these studies have been ignored, namely high computational cost and limited search flexibility. In the following, we conduct a literature review in terms of these two issues in NAS, and then present the knowledge gaps. 

\subsection*{1.2.1. High Computational Cost}
In terms of high computational cost, reinforcement learning (RL)-based search methods are used to provide a more controllable search space (e.g., \citealt{baker2016designing, zoph2016neural}) than evolutionary-based methods (e.g., \citealt{real2017large,xie2017genetic, jalali2021new}), which means RL-based methods can reduce the computational cost. Nonetheless, such methods are also limited by the computational issues, because each sampled architecture has to be trained from the beginning \citep{BAYMURZINA202282}. Concerning this, many research works have proposed warm-started methods by inheriting the weights of the existing network. A representative example is the efficient architecture search (EAS) framework \citep{cai2018efficient}. Briefly, \cite{cai2018efficient} used the Net2Net function-preserving transformation framework \citep{chen2015net2net} and proposed an RL-based EAS framework to widen or deepen an existing network efficiently without modifying the mapping implemented by the network. As noted, the Net2Net framework \citep{chen2015net2net} is likely to rapidly transfer knowledge stored in one neural network to another and contains widening and deepening operations to conduct network function-preserving transformations for CNN and FCN. Despite the excellent performance of the EAS framework, directly applying it to time-series data is inapplicable for two reasons: (i) the CNN and FCN structures cannot well address the internal temporal dependency within time-series data; (ii) the function-preserving transformations for recurrent neural network (RNN) or LSTM (a special case of RNN) have not been investigated. 

\subsection*{1.2.2. Limited Search Flexibility}
In terms of limited search flexibility, existing NAS studies cannot search for neural architectures from two directions. Taking an example of the EAS framework \citep{cai2018efficient}, the updated models can only be enlarged (widened or deepened) so that the models obtained are vulnerable to overfitting the training data \citep{liu2018rethinking}. Contrary to the EAS framework, the N2N framework \citep{ashok2017n2n} proposed a reverse operation that compresses networks using RL by layer removal and layer shrinkage. However, this framework is designed to compress or prune existing networks such as ResNet-18 and VGG-19, rather than directly building the model from data. Due to both limitations of the existing works, how to develop a search strategy that can not only enlarge the networks but also shrink them becomes an effective way to increase the search flexibility in NAS. 

To fill in the research gaps, in this study, we adapt the EAS framework to time-series data to address the computational issue in NAS. Moreover, we develop a flexible search strategy that modifies the network structures from two directions, so as to identify a high-quality electricity time-series forecasting model. The contributions of our proposed framework are discussed in the following section. 

\subsection{Contributions}

In this study, we propose an intelligent automated architecture search (IAAS) framework, which contains three main components, i.e., network transformation operation, network transformation control, and network heuristic screening. The first two components jointly generate new architectures of the given model. The third component expands the search space by using multiple network structures as candidates and screens the generated architectures to only preserve the ones that lead to better performance. With such a framework, we can build a high-quality forecasting model for the given electricity time-series data. The key technical contributions of the proposed IAAS framework have three aspects.

Our first contribution is to adapt the EAS framework to time-series data. As discussed above, the EAS framework \citep{cai2018efficient} has solved the computational problem in NAS by adopting the Net2Net framework \citep{chen2015net2net}. However, the EAS framework targets image data, which is different from time-series electricity data. Therefore, to better analyze the electricity data and extract features from them, we consider RNN as well as CNN and FCN in network transformation operation. However, function-preserving transformation methods for RNN are lacking in the literature. Directly applying the transformation formulations \citep{chen2015net2net} for spatial networks (e.g., CNN and FCN) to RNN is not appropriate, since the current formulations for these networks do not consider the inherent temporal information. Therefore, we innovatively develop the RNN function-preserving transformation via the following steps: (i) we propose a general feature learning formulation for all types of neural networks; (ii) the RNN feature learning process is redesigned based on the proposed general formulation; (iii) we propose wider and deeper function-preserving transformation methods for the reformulated RNN structures. Additionally, considering the outstanding performance of LSTM in developing the time-series forecasting models (see \citealt{shahid2021novel} and \citealt{agga2021short}), we also introduce the function-preserving transformation for LSTM following a similar logic.

Our second contribution is to improve the search flexibility of network structures. The EAS framework \citep{cai2018efficient} exploits the RL-based meta-controllers to modify the existing network. In the transformation control, the RL-based meta-controllers in EAS are pre-determined to take five steps of deeper transformation and four steps of wider transformation at each search episode. This hard-coded transformation order cannot be intelligently changed, and the search process is not flexible. As a comparison, the network operation control in our IAAS framework has two advantages: (i) besides the wider actor and deeper actor used to enlarge the networks, the IAAS framework has a pruning process to shrink the network structures; (ii) we introduce a selector actor to automatically determine whether to widen, deepen, prune or keep the network unchanged at each search episode instead of using the fixed order of the network transformation operations as conducted in EAS. The first advantage is that the search is now bi-directional (increase/decrease), improving the search flexibility. The second advantage is that it avoids the hard-coded transformation, which enables intelligent and adaptive network structure transformation. In addition, we also modify the meta-controllers (i.e., wider actor and deeper actor) from the EAS framework so that the IAAS framework can widen and deepen the network more intelligently and flexibly without following a fixed paradigm that the CNN layers have to be built on the top of FCN layers (see \citealt{cai2018efficient}). 

Our third contribution is to expand the search space of network structures so as to improve the possibility of obtaining high-quality solutions. In EAS, the new network architecture generated by the transformations is directly used in the next-round transformation without screening. Such an implementation is like using one trajectory to train the RL-based meta-controllers. This potentially has one problem: one fixed-length trajectory may not collect enough samples to train the RL-based meta-controllers within limited computational resources. Therefore, some wrong actions of RL may occur during the search procedure. To remedy this, we propose a net pool module in IAAS. Benefiting from this module, we can expand the search space using multiple network structure candidates in the net pool. Moreover, we propose a heuristic screening algorithm to manage the network structures in the net pool so that IAAS can iteratively transform a set of networks for continuous performance improvement. 

We perform comprehensive experiments to evaluate our approach based on two different cases. One is for wind power forecasting from the electricity production side, and the other is for load forecasting from the electricity consumption side. The wind power datasets have been acquired from a wind power company in China, and the load datasets have been collected from ISO New England Inc$^1$. The experiments mainly contain two parts: comparison experiments and ablation experiments. In the comparison experiment, we use ten existing models or methods with standard performance metrics to demonstrate the effectiveness of our framework. As to the ablation experiments, the selector actor and net pool are new compared to the EAS framework. Therefore, we develop three variants of the IAAS framework to showcase the significance of proposing selector actor and net pool components in the IAAS framework. 

All in all, we make the following contributions to the literature:
\begin{enumerate}
    \item[(i)] We propose an RL-based IAAS framework, aiming to build a high-quality electricity forecasting model;
    \item[(ii)] We reformulate the feature learning operation of RNN and LSTM structures, and propose the function-preserving transformation methods for each of them;
    \item[(iii)] We propose a network transformation control system with RL so that the network structures can both be intelligently and flexibly enlarged and shrunk;
    \item[(iv)] We propose a heuristic screening algorithm for the net pool component to iteratively improve the quality of the searched neural architectures;
    \item[(v)] We conduct extensive comparison experiments and ablation experiments on existing electricity datasets to explore the performance of the proposed IAAS framework.
\end{enumerate}

\subsection{Organization of this Article}

The remainder of this paper has been organized as follows. Section \ref{section: Function-preserving-transformation} provides the network transformation methods, including function-preserving transformations for RNN and LSTM and a shrinkage transformation. Section \ref{section:IAAS-framework} proposes the IAAS framework and details its key components. Section \ref{section:Numerical-Results} demonstrates the advantages of the proposed IAAS framework via comparison and ablation experiments. Finally, in Section \ref{section:conclusion}, we discuss the extension of the IAAS framework and provide the concluding remarks.

\section{Network Transformation }\label{section: Function-preserving-transformation}
In the process of searching for a high-quality network structure, we consider two methods to transform the network: one is a function-preserving transformation method to enlarge the network and the other is a pruning method to shrink the network. 

Function-preserving transformation indicates that the change in the network architecture should not influence its functionality \citep{chen2015net2net}. Thus, the initial values for the new set of network parameters ${\boldsymbol{\theta }}^{\boldsymbol{'}}$ in the student neural network $G(\boldsymbol{x}{;}{\boldsymbol{\theta }}^{\boldsymbol{'}})$ (i.e., the extended neural network) should be determined to ensure it has the same outputs as the teacher neural network $F(\boldsymbol{x}{;}\boldsymbol{\theta })$ (i.e., the original neural network) with the same input $\boldsymbol{x}$. The function-preserving transformation is formulated by the following equation:
\begin{equation}
    {\forall }\boldsymbol{x},F(\boldsymbol{x}{;}\boldsymbol{\theta })=G(\boldsymbol{x}{;}{\boldsymbol{\theta }}^{\boldsymbol{'}}).
    \label{eq:function-preserving}
\end{equation}
Eq. \eqref{eq:function-preserving} is an essential characteristic in our proposed framework, indicating that after the network's wider or deeper transformations, the outputs of the student network are the same as those of the teacher network. This property ensures the student network inherits the experience learned by the teacher network and thus enables the warm start of its training process to improve the training efficiency. We first reformulate RNN and LSTM into a two-phase feature learning process in Appendix \ref{section:RNN and LSTM} of the online supplement. Subsequently, we investigate the function-preserving transformation to widen and deepen the RNN and LSTM
 in Sections \ref{subsection-Wider Transformation} and \ref{subsection:Deeper transformation}, respectively.
After that, we present the detailed information of how to prune the network in Section \ref{subsection:shrinkage-transformation}.

\subsection{Wider Transformation} \label{subsection-Wider Transformation}
The wider transformation of a neural network influences at least two layers, which makes it complicated. We will first introduce the wider transformation in RNN and then extend it to LSTM. Suppose that an RNN layer $i$ has $p_i$ input and $p_{i+1}$ output features, and RNN layer $i+1$ has ${p_{i+2}}$ output features, then the parameters of layers $i$ and $i+1$ are ${\boldsymbol{W}}^{(i)}\in {\mathbb{R}}^{p_i \times p_{i+1}}$, ${\boldsymbol{H}}^{(i)}\in {\mathbb{R}}^{p_{i+1}\times p_{i+1}}$ and ${\boldsymbol{W}}^{(i{+1)}}\in {\mathbb{R}}^{p_{i+1}\times {p_{i+2}}}$, ${\boldsymbol{H}}^{(i{+1)}}\in {\mathbb{R}}^{{p_{i+2}}\times {p_{i+2}}}$, respectively. If we widen layer $i$ to layer $i'$ with ${p'_{i+1}}$ output features (${p'_{i+1}}>p_{i+1}$), then the original network parameters of layers $i$ and $i+1$ are replaced by ${\boldsymbol{W}}^{(i')}\in {\mathbb{R}}^{p_i\times {p'_{i+1}}}$, ${\boldsymbol{H}}^{(i')}\in {\mathbb{R}}^{{p'_{i+1}}\times {p'_{i+1}}}$ and ${\boldsymbol{W}}^{(i'{+1)}}\in {\mathbb{R}}^{{p'_{i+1}}\times {p_{i+2}}}$, ${\boldsymbol{H}}^{(i'{+1)}}\in {\mathbb{R}}^{{p_{i+2}}\times {p_{i+2}}}$, separately.

To derive the wider transformation for RNN, we follow \cite{chen2015net2net} to specify a random mapping function $g:\{1,2,\dots,p'_{i+1}\}\to \{1,2,\dots,p_{i+1}\}$ that satisfies:

\begin{equation}
g( k ) = \begin{cases} k & {k \leq p_{i+1}},\\
\text{random sample from}\ {1,2, \ldots ,p_{i+1}} & {k > p_{i+1}}.
\end{cases}
\label{eq:random-mapping}
\end{equation}
Eq. \eqref{eq:random-mapping} indicates how the features in a new student network layer have been represented. The first $p_{i+1}$ features are exactly the same as those in its teacher network layer, while from the $(p_{i+1}+1)^{th}$ feature, they have been randomly replicated from the original features in the teacher network layer. To keep the function-preserving transformation, we utilize a replication factor, $f_w$, similar to that in \cite{chen2015net2net}:
\begin{equation}
    f_w(k)=\frac{1}{\left|\{l|g(k)=g(l)\}\right|},l=1,2,\dots,p_{i+1},\ k=1,2,\dots,{p'_{i+1}},
    \label{eq:replication-factor}
\end{equation}
where $\left|\{l|g(k)=g(l)\}\right|$ is the cardinality of set $\{l|g(k)=g(l)\}$. And this term represents the replication number of one output feature in the original network. For example, if a feature is replicated once, there will be two exactly identical features in the output vector. Then, the replication factor is 1/2, which will be employed simultaneously as the replicated feature weights and its original feature. Subsequently, we can derive the RNN wider transformation with the new student network layer parameters specified by Proposition \ref{proposition: RNN-wider} and its corresponding proof can be found in Appendix \ref{sec:appx-proof-of-propositions} of the online supplement.

\begin{proposition} \label{proposition: RNN-wider}
When widening a layer $i$ to a layer $i'$ in RNN, we can set the parameters of layer $i'$and layer $i'+1$, ${\boldsymbol{W}}^{(i')}$, ${\boldsymbol{H}}^{(i')}$, ${\boldsymbol{W}}^{(i'+1)}$, and ${\boldsymbol{H}}^{(i'+1)}$, as:
\begin{align*}
    \quad \quad \quad \quad \quad W^{(i')}_{j,k}&=W^{(i)}_{j,g(k)},&j=1,2,\dots ,p_i,k=1,\dots,{p'_{i+1}} \quad \quad \quad \quad \quad
    \\
    H^{(i')}_{l,k}&=f_w(l)H^{(i)}_{g(l),g(k)},&l=1,2,\dots ,{p'_{i+1}},k=1,\dots,{p'_{i+1}} \quad \quad \quad \quad \quad
        \\
    W^{(i'+1)}_{k,h}&=f_w(k)W^{(i+1)}_{g(k),h},&k=1,2,\dots ,{p'_{i+1}},h=1,\dots,{p_{i+2}} \quad \quad \quad \quad \quad
        \\
    H^{(i'+1)}_{r,h}&=H^{(i+1)}_{r,h},&r=1,2,\dots ,{p_{i+2}},h=1,\dots,{p_{i+2}}. \quad \quad \quad \quad \quad
\end{align*}
Then, the wider function-preserving transformation for RNN layer $i$ is satisfied.
\end{proposition}

For LSTM, we also assume that LSTM layer $i$ has $p_i$ input and $p_{i+1}$ output features, and LSTM layer $i + 1$ has $p_{i+1}$ input and ${p_{i+2}}$ output features. Hence, the parameters of layer $i$ are $\boldsymbol{W}^{(i)}_f\in \mathbb{R}^{p_i \times p_{i+1}}, \boldsymbol{W}^{(i)}_i\in \mathbb{R}^{p_i \times p_{i+1}}$, $\boldsymbol{W}^{(i)}_o\in \mathbb{R}^{p_i \times p_{i+1}}$, $\boldsymbol{W}^{(i)}_c\in \mathbb{R}^{p_i \times p_{i+1}}$, $\boldsymbol{H}^{(i)}_f\in \mathbb{R}^{p_{i+1}\times p_{i+1}}$, $\boldsymbol{H}^{(i)}_i\in \mathbb{R}^{p_{i+1}\times p_{i+1}}$, $\boldsymbol{H}^{(i)}_o\in \mathbb{R}^{p_{i+1}\times p_{i+1}}$, and $\boldsymbol{H}^{(i)}_c\in \mathbb{R}^{p_{i+1}\times p_{i+1}}$; moreover, the parameters of layer $i+1$ are $\boldsymbol{W}^{(i+1)}_f\in \mathbb{R}^{p_{i+1}\times p_{i+2}}, \boldsymbol{W}^{(i+1)}_i\in \mathbb{R}^{p_{i+1}\times p_{i+2}}$, $\boldsymbol{W}^{(i+1)}_o\in \mathbb{R}^{p_{i+1}\times p_{i+2}}$, $\boldsymbol{W}^{(i+1)}_c\in \mathbb{R}^{p_{i+1}\times p_{i+2}}$, $\boldsymbol{H}^{(i+1)}_f\in \mathbb{R}^{{p_{i+2}}\times {p_{i+2}}}$, $\boldsymbol{H}^{(i+1)}_i\in \mathbb{R}^{{p_{i+2}}\times {p_{i+2}}}$, $\boldsymbol{H}^{(i+1)}_o\in \mathbb{R}^{{p_{i+2}}\times {p_{i+2}}}$, and $\boldsymbol{H}^{(i+1)}_c\in \mathbb{R}^{{p_{i+2}}\times {p_{i+2}}}$. If we widen layer $i$ to layer $i'$ with ${p'_{i+1}}$ output features (${p'_{i+1}}>p_{i+1}$), the original network parameters of layers $i$ are replaced by $\boldsymbol{W}^{(i')}_f\in \mathbb{R}^{p_i \times p'_{i+1}}, \boldsymbol{W}^{(i')}_i\in \mathbb{R}^{p_i \times p'_{i+1}}$, $\boldsymbol{W}^{(i')}_o\in \mathbb{R}^{p_i \times p'_{i+1}}$, $\boldsymbol{W}^{(i')}_c\in \mathbb{R}^{p_i \times p'_{i+1}}$, $\boldsymbol{H}^{(i')}_f\in \mathbb{R}^{{p'_{i+1}}\times {p'_{i+1}}}$, $\boldsymbol{H}^{(i')}_i\in \mathbb{R}^{{p'_{i+1}}\times {p'_{i+1}}}$, $\boldsymbol{H}^{(i')}_o\in \mathbb{R}^{{p'_{i+1}}\times {p'_{i+1}}}$, and $\boldsymbol{H}^{(i')}_c\in \mathbb{R}^{{p'_{i+1}}\times {p'_{i+1}}}$. In addition, the parameters of layer $i+1$ are replaced by $\boldsymbol{W}^{(i'+1)}_f\in \mathbb{R}^{p'_{i+1}\times p_{i+2}}, \boldsymbol{W}^{(i'+1)}_i\in \mathbb{R}^{p'_{i+1}\times p_{i+2}}$, $\boldsymbol{W}^{(i'+1)}_o\in \mathbb{R}^{p'_{i+1}\times p_{i+2}}$, $\boldsymbol{W}^{(i'+1)}_c\in \mathbb{R}^{p'_{i+1}\times p_{i+2}}$, $\boldsymbol{H}^{(i'+1)}_f\in \mathbb{R}^{{p_{i+2}}\times {p_{i+2}}}$, $\boldsymbol{H}^{(i'+1)}_i\in \mathbb{R}^{{p_{i+2}}\times {p_{i+2}}}$, $\boldsymbol{H}^{(i'+1)}_o\in \mathbb{R}^{{p_{i+2}}\times {p_{i+2}}}$, and $\boldsymbol{H}^{(i'+1)}_c\in \mathbb{R}^{{p_{i+2}}\times {p_{i+2}}}$. Therefore, we can derive the LSTM wider transformation with the new student network layer parameters specified by Proposition \ref{proposition: LSTM-wider}, and its proof can be found in  Appendix \ref{sec:appx-proof-of-propositions} of the online supplement.

\begin{proposition} \label{proposition: LSTM-wider}
When widening a layer $i$ to a layer $i'$ in LSTM, we can set the parameters of layer $i'$and layer $i'+1$ as:
\begin{align*}
    \quad \quad \quad \quad \quad W^{(i')}_{f;j,k}&=W^{(i)}_{f;j,g(k)},&j=1,2,\dots ,p_i,k=1,\dots,{p'_{i+1}} \quad \quad \quad \quad \quad
    \\
    \quad \quad \quad \quad \quad W^{(i')}_{i;j,k}&=W^{(i)}_{i;j,g(k)},&j=1,2,\dots ,p_i,k=1,\dots,{p'_{i+1}} \quad \quad \quad \quad \quad
    \\
    \quad \quad \quad \quad \quad W^{(i')}_{o;j,k}&=W^{(i)}_{o;j,g(k)},&j=1,2,\dots ,p_i,k=1,\dots,{p'_{i+1}} \quad \quad \quad \quad \quad
    \\
    \quad \quad \quad \quad \quad W^{(i')}_{c;j,k}&=W^{(i)}_{c;j,g(k)},&j=1,2,\dots ,p_i,k=1,\dots,{p'_{i+1}} \quad \quad \quad \quad \quad
    \\
    H^{(i')}_{f;l,k}&=f_w(l)H^{(i)}_{f;g(l),g(k)},&l=1,2,\dots ,{p'_{i+1}},k=1,\dots,{p'_{i+1}} \quad \quad \quad \quad \quad
        \\
    H^{(i')}_{i;l,k}&=f_w(l)H^{(i)}_{i;g(l),g(k)},&l=1,2,\dots ,{p'_{i+1}},k=1,\dots,{p'_{i+1}} \quad \quad \quad \quad \quad
        \\
    H^{(i')}_{o;l,k}&=f_w(l)H^{(i)}_{o;g(l),g(k)},&l=1,2,\dots ,{p'_{i+1}},k=1,\dots,{p'_{i+1}} \quad \quad \quad \quad \quad
        \\
    H^{(i')}_{c;l,k}&=f_w(l)H^{(i)}_{c;g(l),g(k)},&l=1,2,\dots ,{p'_{i+1}},k=1,\dots,{p'_{i+1}} \quad \quad \quad \quad \quad
        \\
    W^{(i'+1)}_{f;k,h}&=f_w(k)W^{(i+1)}_{f;g(k),h},&k=1,2,\dots ,{p'_{i+1}},h=1,\dots,{p_{i+2}} \quad \quad \quad \quad \quad
        \\
    W^{(i'+1)}_{i;k,h}&=f_w(k)W^{(i+1)}_{i;g(k),h},&k=1,2,\dots ,{p'_{i+1}},h=1,\dots,{p_{i+2}} \quad \quad \quad \quad \quad
        \\
    W^{(i'+1)}_{o;k,h}&=f_w(k)W^{(i+1)}_{o;g(k),h},&k=1,2,\dots ,{p'_{i+1}},h=1,\dots,{p_{i+2}} \quad \quad \quad \quad \quad
        \\
    W^{(i'+1)}_{c;k,h}&=f_w(k)W^{(i+1)}_{c;g(k),h},&k=1,2,\dots ,{p'_{i+1}},h=1,\dots,{p_{i+2}} \quad \quad \quad \quad \quad
        \\
    H^{(i'+1)}_{f;r,h}&=H^{(i+1)}_{f;r,h},&r=1,2,\dots ,{p_{i+2}},h=1,\dots,{p_{i+2}} \quad \quad \quad \quad \quad 
    \\
    H^{(i'+1)}_{i;r,h}&=H^{(i+1)}_{i;r,h},&r=1,2,\dots ,{p_{i+2}},h=1,\dots,{p_{i+2}} \quad \quad \quad \quad \quad 
    \\
    H^{(i'+1)}_{o;r,h}&=H^{(i+1)}_{o;r,h},&r=1,2,\dots ,{p_{i+2}},h=1,\dots,{p_{i+2}} \quad \quad \quad \quad \quad 
    \\
    H^{(i'+1)}_{c;r,h}&=H^{(i+1)}_{c;r,h},&r=1,2,\dots ,{p_{i+2}},h=1,\dots,{p_{i+2}}. \quad \quad \quad \quad \quad 
\end{align*}
Then, the wider function-preserving transformation for LSTM layer $i$ is satisfied.
\end{proposition}

\subsection{Deeper Transformation }\label{subsection:Deeper transformation}
The deeper transformations for RNN and LSTM are developed in a much more straightforward manner than their wider transformations. In the following, we will first use RNN to illustrate the essential process of deeper transformation, and then extend it to LSTM. Similar to \cite{chen2015net2net}, the deeper transformation should replace an RNN layer with two RNN layers using the identity mapping method. In other words, the input features of the new student RNN layer should be identical to those of the teacher RNN layer. Suppose that we deepen an RNN layer $i$ with a new student layer $i^{''}$ by inserting the student layer $i^{''}$ after the RNN layer $i$. If layer $i$ has $p_{i+1}$ output features, then layer $i^{''}$ is assumed to have $p_{i+1}$ input and $p_{i+1}$ output features for simplicity. Note that the input and output sizes of layer $i^{''}$ are not restricted to be $p_{i+1}$. Using the other value is also acceptable. However, this brings more complexities to guarantee the function-preserving transformation, which is beyond the scope of this study. The new student network parameters of the layer $i^{''}$ are ${\boldsymbol{W}}^{(i^{''})}\in {\mathbb{R}}^{p_{i+1}\times p_{i+1}}\ $and ${\boldsymbol{H}}^{(i^{''})}\in {\mathbb{R}}^{p_{i+1}\times p_{i+1}}$. According to \cite{chen2015net2net}, the deeper transformation is applicable when the activation function has this property, i.e., $f_a(f_a(\boldsymbol{z}))=f_a(\boldsymbol{z})$,  where $\boldsymbol{z}$ is a vector. This property means that even though a vector passes the activation function multiple times, the output is kept unchanged. Considering this, we employ the rectified linear unit (ReLU) activation function to satisfy this property. In addition, for a more general application that the network has no RNN layers, we are also able to insert a new student RNN layer into the network if the new student network parameters satisfy Proposition \ref{proposition: RNN-deeper}. For example, in a network, if layer\textit{ }$i$\textit{ }is a CNN layer and layer $i+1$ is an FCN layer, we can insert an RNN layer between layer $i$ and layer $i+1$.

\begin{proposition} \label{proposition: RNN-deeper}

When deepening layer $i$ with an RNN layer $i^{''}$ in a network, we can set the RNN layer parameters $W^{(i^{''})}_{l,k}$ and $H^{(i^{''})}_{l,k}$ as:
\begin{align*}
    \quad \quad \quad \quad \quad \quad W^{(i^{''})}_{l,k}&=
    \begin{cases}
    1&l=k \\ 
    0&l\neq k
    \end{cases}
    ,&l=1,2,\dots ,{p_{i+1}},k=1,\dots,{p_{i+1}}, \quad \quad \quad \quad \quad\\
    {H}^{(i^{''})}_{l,k}&=0,&l=1,2,\dots ,{p_{i+1}},k=1,\dots,{p_{i+1}}. \quad \quad \quad \quad \quad.
\end{align*}
Then, the deeper function-preserving transformation for RNN layer $i$ is satisfied.
\end{proposition}

For LSTM, we can insert a new student LSTM layer into the network if the parameters of this new student network fulfill Proposition \ref{proposition: LSTM-deeper}. Similar to RNN, LSTM can be inserted into any place of a network.
\begin{proposition} \label{proposition: LSTM-deeper}

When deepening layer $i$ with an LSTM layer $i^{''}$ in a network, we can set the LSTM layer parameters as:
\begin{align*}
    \quad \quad \quad \quad \quad \quad W^{(i^{''})}_{o;l,k}&=
    \begin{cases}
    1&l=k \\ 
    0&l\neq k
    \end{cases}
    ,&l=1,2,\dots ,p_{i+1},k=1,\dots,p_{i+1}, \quad \quad \quad \quad \quad\\
    {W}^{(i^{''})}_{f;l,k} ={W}^{(i^{''})}_{i;l,k} ={W}^{(i^{''})}_{c;l,k}&=0,&l=1,2,\dots ,p_{i+1},k=1,\dots,p_{i+1}, \quad \quad \quad \quad \quad\\
    {H}^{(i^{''})}_{f;l,k} ={H}^{(i^{''})}_{i;l,k} ={H}^{(i^{''})}_{o;l,k} ={H}^{(i^{''})}_{c;l,k}&=0,&l=1,2,\dots ,p_{i+1},k=1,\dots,p_{i+1}. \quad \quad \quad \quad \quad
\end{align*}
Then, the deeper function-preserving transformation for LSTM layer $i$  is satisfied.
\end{proposition}

\subsection{Shrinkage Transformation} \label{subsection:shrinkage-transformation}

In this study, we consider a score-based pruning method, i.e., movement pruning \citep{sanh2020movement}, to shrink the network structures. Score-based pruning is to prune the network based on the importance score. In the following, we use FCN as an example to illustrate how the movement pruning method performs. Let $\boldsymbol{W}^{(i)}\in \mathbb{R}^{p_i\times p_{i+1}}$ be the weight matrix for layer $i$ and $\boldsymbol{x}^{(i+1)} = \boldsymbol{W}^{(i)}\boldsymbol{x}^{(i)} \in \mathbb{R}^{p_{i+1}}$ be the output  for the input $\boldsymbol{x}^{(i)} \in \mathbb{R}^{p_i}$. To determine which weights to be pruned, we use the importance scores $\boldsymbol{\Lambda}^{(i)}\in \mathbb{R}^{p_i\times p_{i+1}}$ for $\boldsymbol{W}^{(i)}$. 

According to $\boldsymbol{\Lambda}^{(i)}$, the pruning strategy is built on a mask $\boldsymbol{M}^{(i)}\in \{0,1\}^{p_i\times p_{i+1}}$, which generates a new weight matrix $\boldsymbol{W}^{(i)}\odot \boldsymbol{M}^{(i)}$ ($\odot$ is the Hadamard product), to produce a new output $\boldsymbol{x}^{(i+1)}=(\boldsymbol{W}^{(i)}\odot \boldsymbol{M}^{(i)})\boldsymbol{x}^{(i)}$. Particularly, \cite{sanh2020movement} defines the $Top_v$ strategy to compute the mask, which selects the $v\%$ highest values in $\boldsymbol{\Lambda}^{(i)}$, as presented in Eq. (\ref{eq:top-v}): 
\begin{align}
    \label{eq:top-v}
    Top_v(\boldsymbol{\Lambda}^{(i)})_{l,k} =
    \begin{cases}
    1,  &\mathit{\Lambda ^{(i)}_{l,k}\ in\ top\ }v\% \ ,l =1,\cdots,p_i, k =1,\cdots,p_{i+1},\\ 
    0,  &\mathit{otherwise,}
    \end{cases}
\end{align}
where $ \mathit{\Lambda ^{(i)}_{l,k}}, l =1,\cdots,p_i, k =1,\cdots,p_{i+1}$, represent one element in $\boldsymbol{\Lambda}^{(i)}$.

During training, the movement pruning method uses the first-order information of the loss function $L$ to update the importance score $\Lambda ^{(i)}_{l,k}$ with the gradient:
\begin{equation} \label{eq:importance-score-gradient}
    \frac{\partial L}{\partial {\Lambda}^{(i)}_{l,k}} = \frac{\partial L}{\partial {x}_k^{(i+1)} }\frac{\partial {x}_k^{(i+1)}}{\partial {\Lambda}^{(i)}_{l,k}} = \frac{\partial L}{\partial {x}_k^{(i+1)} } {W}^{(i)}_{l,k} {x}^{(i)}_l  \ ,l =1,\cdots,p_i, k =1,\cdots,p_{i+1}.
\end{equation}
Eq. (\ref{eq:importance-score-gradient}) implies that the importance score $\Lambda ^{(i)}_{l,k}$ increases when $W^{(i)}_{l,k}$ moves away from 0, and decreases when $W^{(i)}_{l,k}$ moves towards to 0. Note that the gradient of importance score is calculated by automatic differentiation technique \citep{baydin2018automatic}. In this study, we consider the weight associated with a particular neuron to be the smallest unit for executing the pruning operation. We calculate the importance score of the output neuron $k$ as
\begin{align}
   {\lambda}^{(i+1)}_k = \sum_{l=1}^{p_i} \Lambda ^{(i)}_{l,k},
\end{align}
where ${\lambda}^{(i+1)}_k$ represent the $k^{th}$ element in vector $\boldsymbol{\lambda}^{(i+1)}$, the importance score for the output $\boldsymbol{x}^{(i+1)}$ is represented as $\boldsymbol{\lambda}^{(i+1)}$, and the mask for $\boldsymbol{\lambda}^{(i+1)}$ is represented as $\boldsymbol{m}^{(i+1)}$. We can then calculate the importance score of all neurons and employ the $Top_v$ strategy to eliminate neurons of less importance, as well as their corresponding weights. Consequently, we can obtain a smaller network structure. In   Appendix \ref{appendix:pruning-cnn-lstm} of the online supplement, we present the details of how to use the movement pruning method to prune FCN, RNN, CNN and LSTM.

\section{IAAS Framework Development}\label{section:IAAS-framework}

In this section, we propose an IAAS framework to intelligently and flexibly search the neural architecture. Primarily, this framework has three main components, i.e., network transformation operation, network transformation control, and network heuristic screening. The new model architecture is generated by conducting the network transformation operation determined by the network transformation control. Subsequently, the newly generated model architectures are further evaluated and selected by the network heuristic screening. In the following, we first describe the overview of the IAAS framework and then introduce three actor networks and pruning implementation details as well as the net pool component utilized in the framework.

\subsection{Overview of IAAS Framework}\label{subsection:overview-IAAS-framework}

We design the IAAS framework with an end-to-end mode. Specifically, we do not need to manually determine the number of layers and the size of each layer. Moreover, this framework can automatically determine when to widen, deepen, and prune the network. Lastly, after the maximum number of search episodes is reached, the framework can automatically output the best neural architecture. In other words, we just need to prepare the input and output data, and this framework can intelligently build a high-quality neural architecture for the given data.

In a neural network, each layer usually contains different parameters, which are critical information for the network transformation. According to \cite{zoph2016neural}, we specify the parameters of each layer in the form of a variable-length string. Subsequently, the proposed IAAS framework is applied to these strings for intelligent implementation of network transformation.

\begin{figure}[H]
\centering
\includegraphics[scale=0.4]{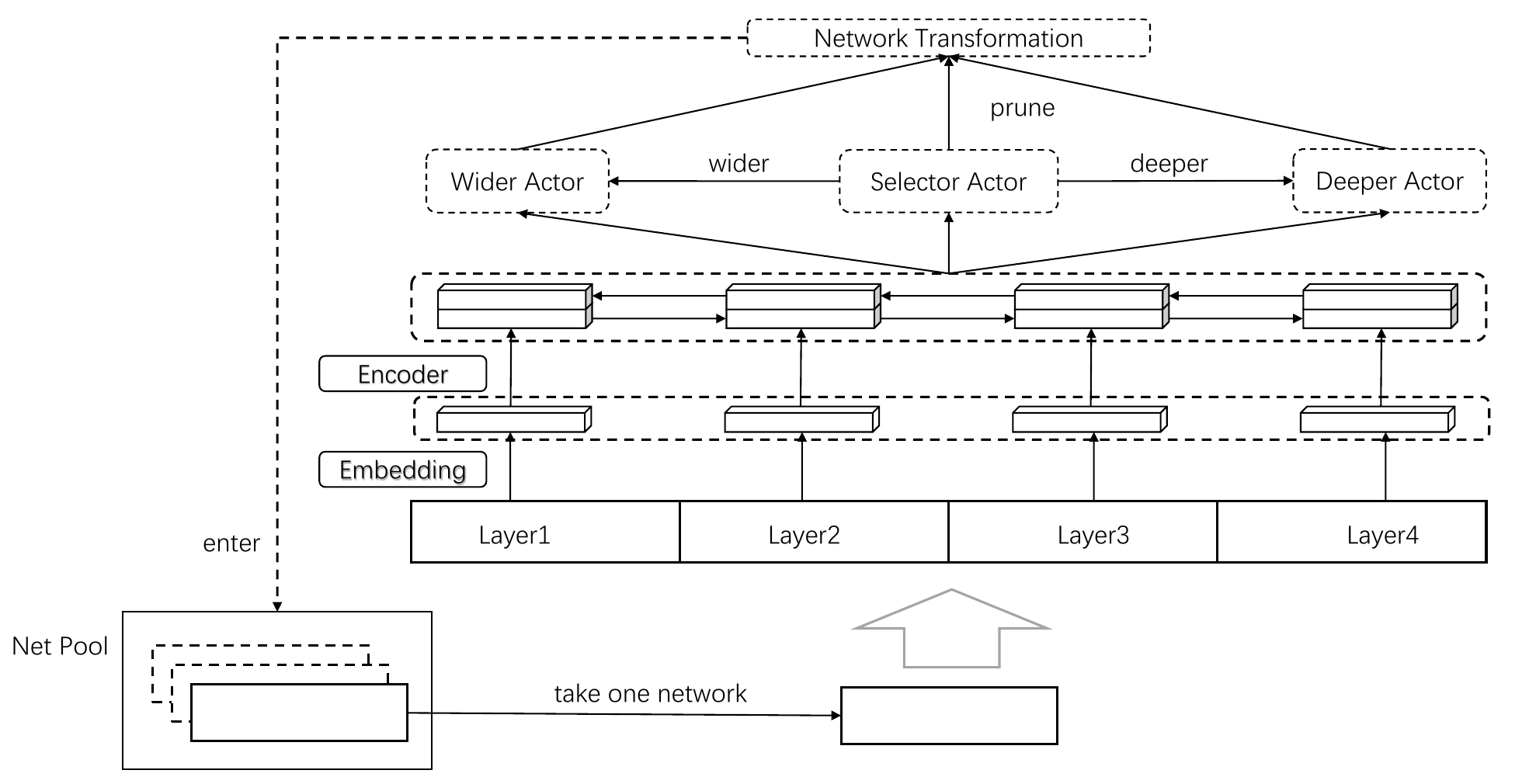}
\caption{Overview of the proposed IAAS framework with an example of a four-layer input architecture} 
\label{fig:IAAS-framework}
\end{figure}

With the seq2seq technique \citep{Bahdanau2015}, each string containing the information of the neural architecture layer is first processed by an embedding layer in the IAAS framework, making it a fixed-length vector. We follow the technique given in \cite{cai2018efficient} and set the vector length as 16 in this study. Hence, if the network architecture has $N$ layers, we then obtain a matrix with the size of $N\times 16$ to represent the architecture information. Similar to \cite{cai2018efficient}, an encoder network built with a bi-directional LSTM network is subsequently used to learn the representations of this matrix. Note that the fixed-length vector from the embedding layer is necessary for the encoder network. We present an illustrative example of the IAAS framework with a four-layer input architecture in Fig. \ref{fig:IAAS-framework}. It can be observed that after the encoder network, the learned features are simultaneously fed into three actors, namely wider actor, selector actor, and deeper actor, which are the cores of the network transformation control. Selector actor pioneers the decision-making process with four output action candidates, i.e., ``unchanged", ``pruning", ``widening", and ``deepening". Briefly, a pruning action is to shrink the network with the movement pruning algorithm \citep{sanh2020movement}, while a widening or deepening action is sent to the wider actor or deeper deeper for the network enlargement. It can be noted that the selector actor automates the process of widening or deepening the network, successfully avoiding the manual settings for the network transformation in the EAS framework \citep{cai2018efficient}. After the network transformation operation, the updated architecture will be entered into a net pool with a limited size. When the net pool receives all the transformed networks at one search episode, a heuristic screening algorithm is implemented to eliminate those networks with bad performance. The rest networks will then be implemented for the next-round transformation.

\subsection{Markov Decision Process}\label{subsection: MDP} 

In this research, the sequential optimization of network architecture is formulated as a Markov decision process (MDP), which is solved by reinforcement learning (RL). Specifically, the state $s_u$ is the network architecture at the $u^{th}$ transformation step (we use $u$ here since $t$ has been used as the notation for the time-series data). Three RL agents are defined to have the different action spaces so that they can adaptively select the appropriate transformation actions and intelligently conduct the specific network transformation to update the network architecture (see detailed information in Section \ref{subsection:actor-networks}). We evaluate the reward signal $r_u$ with the model forecasting performance on the testing datasets. In particular, we define $r_u$ as:
\begin{equation}
  r_u=\frac{1}{RMSE_u},
\label{eq:reward function}
\end{equation}
where $RMSE_u$ indicates the rooted mean squared error of forecasting accuracy at the $u^{th}$ transformation step. This new reward function with $RMSE$ can award the agent with a nonlinear rate so that the same magnitude reduction in a smaller $RMSE$ is more prominent than that in a higher $RMSE$. The goals of the three RL agents are all to maximize the discounted reward, i.e., $R_u=\sum_{i\ge 0} \gamma ^i r_{u+i}$, in expectation, where $\gamma \in [0,1)$ is the discount factor.

At the transformation step $u$, the three agents work together to make a transformation decision $a_u$ based on the parameters $\theta_u$ and the current network structure $s_u$. The policy $\mu(\cdot, s_u)$ generated by the agents is also recorded as the behavior policy to train the actor in the present and future steps. After the transformation, a new network structure $s_{u+1}$ is obtained. With a few training iterations, the forecasting accuracy is evaluated and the reward signal $r_u$ is calculated. This provides the data $(s_u,a_u,r_u,\mu(\cdot|s_u))$ for the RL algorithm to update the parameters $\theta_u$. After updating $\theta_u$, a new set of parameters $\theta_{u+1}$ is achieved for
 decision making at the next step. This process is repeated until the search budget is exhausted.

We aim to optimize the non-differentiable expected long-term reward $\mathbb{E}(R_u)$ by using a policy gradient method called actor-critic with experience replay (ACER) (\citealt{wang2017sample}). This algorithm is used to train three agents simultaneously, allowing us to update each agent with the most recent information based on the performance of all agents. For example, the selector agent will utilize the transformation actions taken from wider and deeper agents to prioritize its decision (widening, deepening, pruning or keeping network unchanged). Without such a design, the selector agent may not well evaluate the decisions made before, leading to an erratic decision possibly to transform the networks. After each step of the network transformation, a new sampling trajectory $\{s_0,a_0,r_0,\mu(\cdot s_0),\dots,s_u,a_u,r_u,\mu(\cdot s_u)\}$ is generated. ACER is then used to calculate the policy gradient and update the parameters $\theta_u$. Moreover, by transforming the network structures in net pool, multiple sampling trajectories are generated and utilized for parameter updating. Each of these trajectories is composed by a series of decisions. With these trajectories, we can optimize the parameters of actor and critic networks, leading to searching a quality network structure since the reward signal is based on the network's performance. Please refer to the detailed information of how to use the ACER algorithm to update the parameters of RL agents in   Appendix \ref{ACER} of the online supplement and how RL is involved in the IAAS framework in   Appendix \ref{appx-algorithms} of the online supplement.

\subsection{Actor Networks}\label{subsection:actor-networks}

In this section, we propose three actor networks, namely selector actor, wider actor, and deeper actor. Corresponding to each actor network, we construct a  critic network with two FCN layers. The actor network is used to output actions, while the critic network is exploited to evaluate the goodness of the actions. In the following, we introduce detailed information about the three actor networks.

\subsection*{3.3.1. Selector Actor}\label{subsubsection:selector-actor}
\begin{figure}[h]
    \centering
    \includegraphics[scale=0.4]{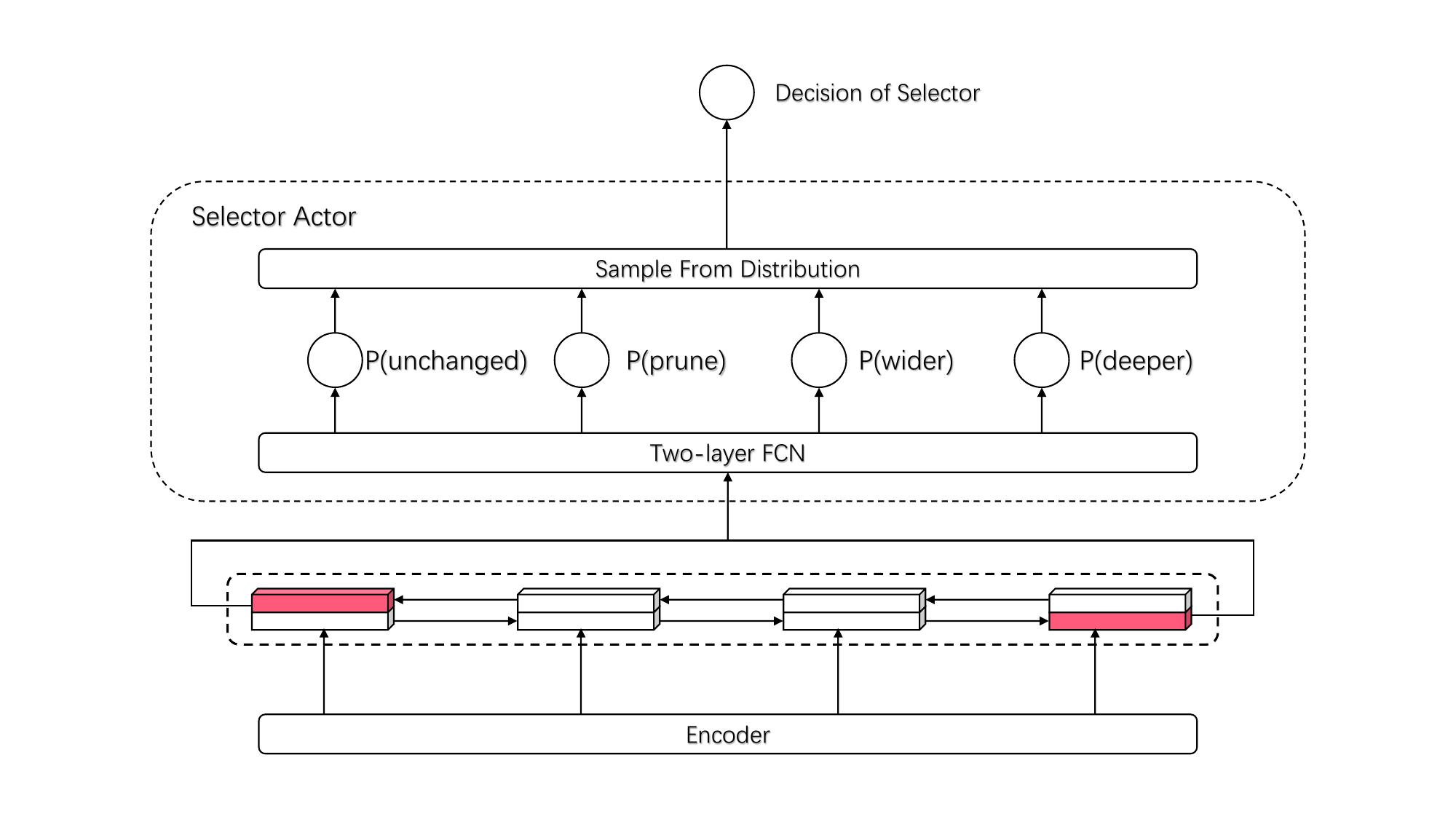}
    \caption{Illustration of selector actor net (P indicates the probability)\label{fig: selector}}
\end{figure}

The architecture of the selector actor is shown in Fig. \ref{fig: selector}. As seen, the embedded representation of the current network architecture is fed into the bi-LSTM encoder network for feature extraction. In bi-LSTM, the features are extracted from both the forward and backward information flows of the input. In our problem, the forward and backward information flows to capture the dependencies among different layers in the current network architecture. The objective of having the selector actor is to determine the specific transformation operation for the current model architecture. Thus, only the features at the end of these two information flows are fed into the selector actor (we mark them as red in Fig. \ref{fig: selector}) because they contain the accumulative information and represent the overall characteristics of the current architecture. Then, the output of the selector actor indicates the probabilities of four actions, i.e., keep the network unchanged, prune the network, widen the network, and deepen the network. In Fig. \ref{fig: selector}, they are denoted as P(unchanged), P(prune), P(wider), and P(deeper), respectively. Instead of selecting the action with the highest probability, we leverage a distribution sampling method. Specifically, we use the softmax function to normalize these four probabilities (${x_i}$, $i$=1,2,3,4):
\begin{equation}
\sigma({\boldsymbol{x}})_i=\frac{e^{x_i}}{\sum_je^{x_j}}, i={1,2,3,4}.
\label{normalization}
\end{equation}

We then obtain a discrete probability distribution with these normalized probabilities. An action is sampled from this discrete distribution as the final action of the selector actor. The distribution sampling method can both ensure the selector actor exploits the learned experience (preferably select the action with a higher probability) and maintain the ability to explore new strategies (still possibly select the action with a lower probability). Thus, it can prevent the selector actor from trapping into the local optimum. Given the advantage of the distribution sampling method, we also include it in the design of the wider actor and the deeper actor.

For each transformation step, we use the current selector actor to generate the target policy $\pi_s$ (see   Appendix \ref{ACER} of the online supplement), and exploit the policy gradient in Eq. (\ref{eq:ACER-gradient}) and Eq. (\ref{eq:critic-gradient}) to optimize the parameters in selector actor network and critic network with Adam optimizer. For the wider and deeper actors, we follow the similar procedure to optimize the parameters of the actor and critic networks.

\subsection*{3.3.2. Wider Actor}\label{subsubsection:wider-actor}
\begin{figure}
\centering
    \includegraphics[scale=0.4]{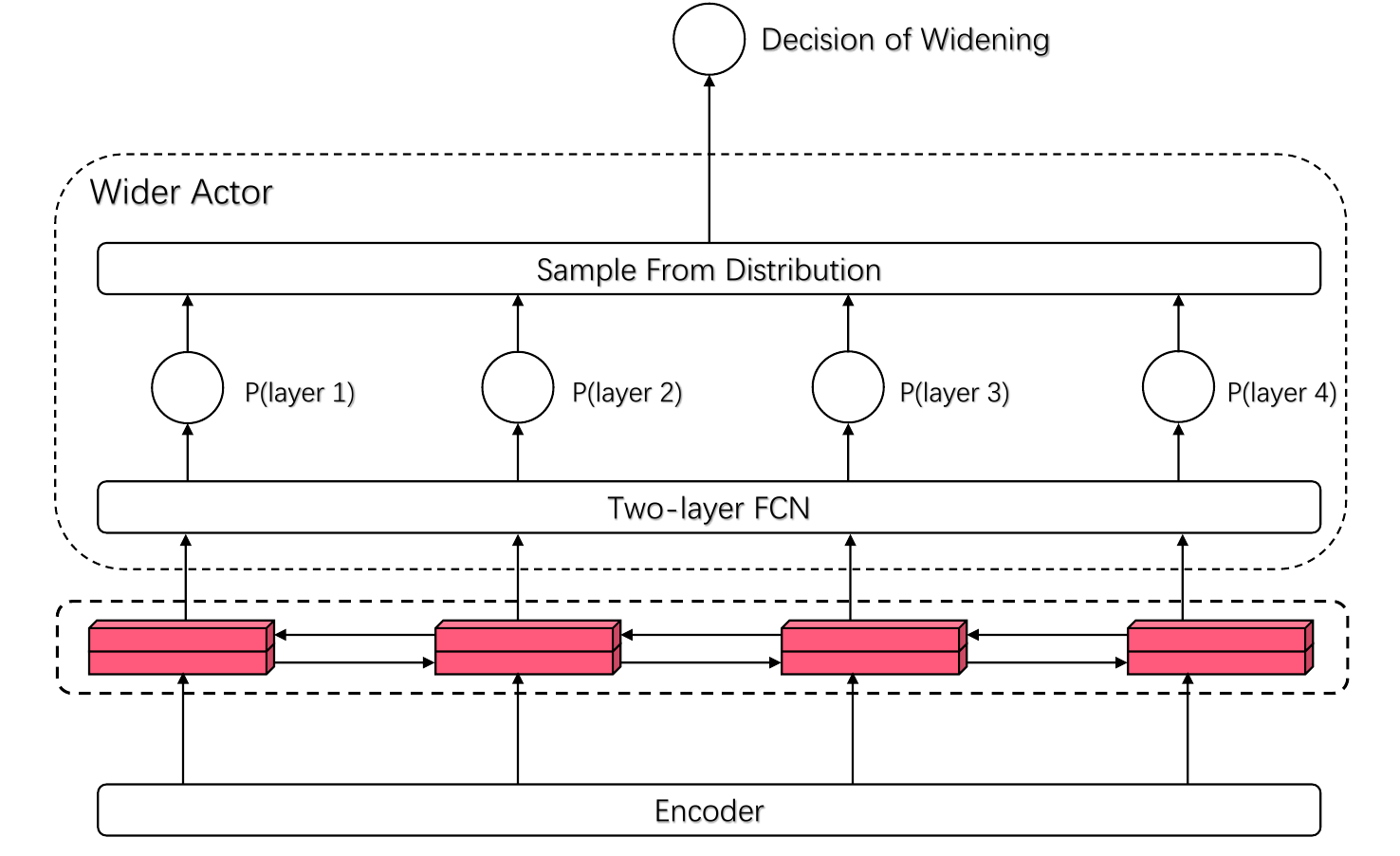}
    \caption{Illustration of the wider actor net with an input neural architecture with four layers \\
    {\small \textit{Note}. all the four layers have the same type of neural structure in this example; P indicates the probability.}
    \label{fig: wider}}
    
\end{figure}

This actor is designed to widen a network layer with more units. Fig. \ref{fig: wider} shows the structure of the wider actor net. Similar to the selector actor, we also use an FCN to learn the features from the encoder network. Different from the selector actor, the wider actor aims to evaluate all layers and increase the units for a specific layer. Thus, the layer-wise features are fed into the FCN to determine actions (We mark these features as red in Fig. \ref{fig: wider}.). It is worth noting that a shared FCN is used to generate the probability of widening actions for all layers. Thus, the FCN still remains the same after changing the number of layers (i.e., after the deeper transformations). The wider actor net will output four probabilities for a four-layer neural architecture as depicted in Fig. \ref{fig: wider}. Instead of directly widening the layer with the highest probability, we normalize these probabilities and make the normalized probability as a discrete probability distribution, which is the same as that in the selector actor net. With such a distribution, we can sample an action to evaluate which layer should be widened eventually.  

After selecting the layer to be widened, the wider actor needs to specify the number of units after the widening operation. Note the units for different network structures represent different items. For example, in FCN, the units represent the neurons, while in CNN, the units represent the channels. We set a widening rule with a self-defined sequence $e.g.,\ [4,\ 8,\ 16,\ 32,\ \dots,\ K,\ K+16].$ It denotes that if the current layer has four units, we then widen it to eight units; if the current layer has 8 units, we then widen it to 16 units; if the current layer has \textit{K }units with $K\ge 16$, we then widen it to \textit{K}+16 units. Hence, from a theoretical perspective, \textit{K }could be infinitely large. If the number of units in the widening layer is not the exact number in the sequence, we will first widen the number to the nearest one. For instance, we widen a CNN layer with three output channels; then, we should first widen it to a CNN layer with four output channels. Therefore, this setting is quite accommodating since we do not need to manually set the maximum widening units.

\subsection*{3.3.3 Deeper Actor}\label{subsubsection:deeper-actor}
\begin{figure}
\centering
    \includegraphics[scale=0.4]{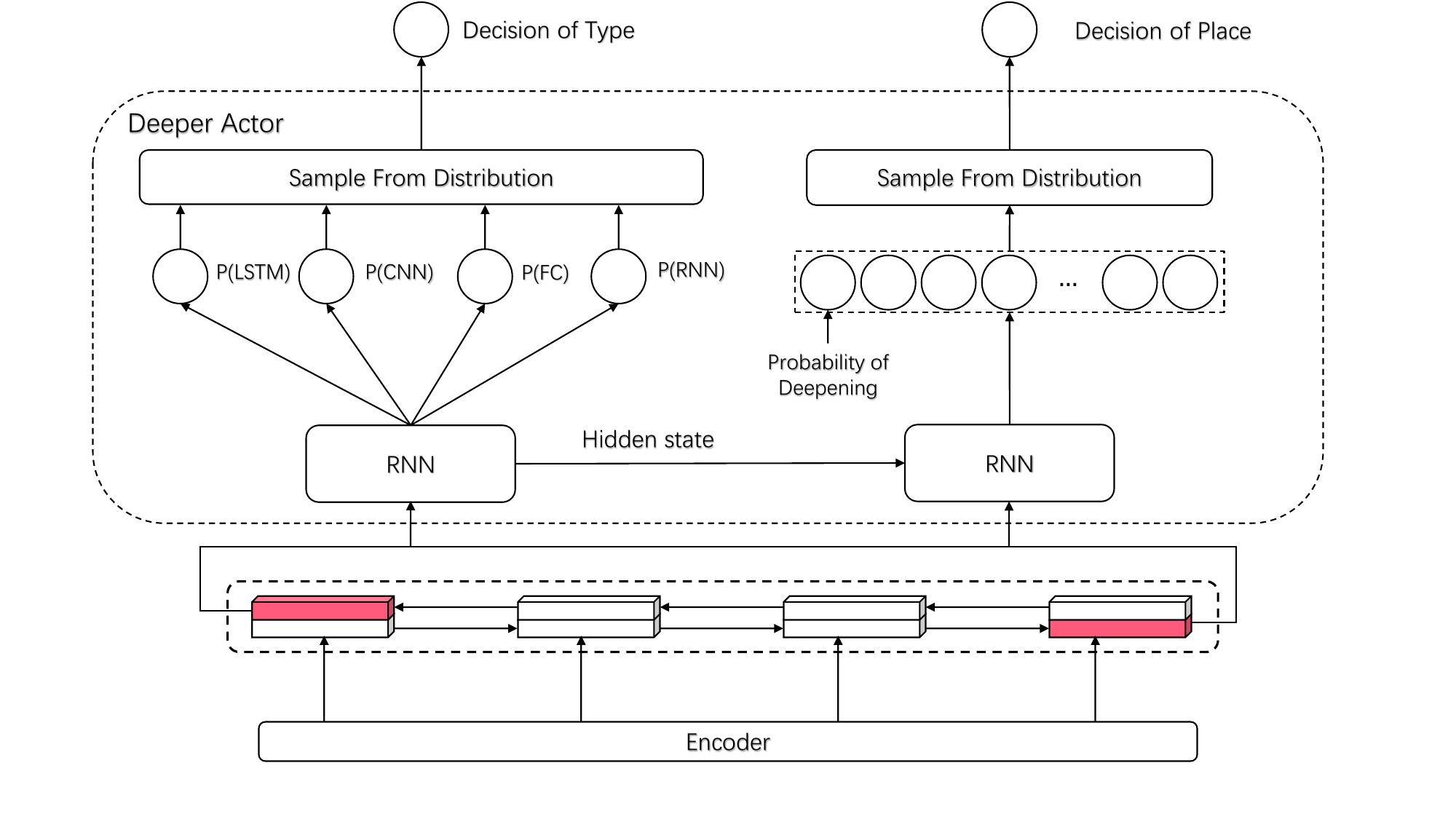}
    \caption{ Illustration of deeper actor net\\
    {\small \textit{Note.} P indicates the probability of using CNN, FCN or RNN layers.}
    \label{fig: deeper}}
    
\end{figure}

This section proposes a powerful deeper actor to deepen the given network. Specifically, we have 4 layer candidates, namely, FCN layer, CNN layer, RNN layer, and LSTM layer. In the process of actor training, we believe any layer of the given network architecture can be deepened. For comparison, the Net2Deeper actor in the EAS framework set a limitation that only an FCN layer can be added on top of all CNN and pooling layers for a given network. Therefore, to make the deepening process more flexible, the deeper actor is designed with the structure presented in Fig. \ref{fig: deeper}. Similar to the selector actor, the learned last two features (we mark them as red in Fig. \ref{fig: deeper}) from the encoder network are selected because they represent the characteristics of the current model architecture. These features are fed into an RNN structure with a softmax activation function. There is a two-step action for the deeper actor. The first step is to determine what type of layer (\textit{e.g., }FCN, CNN, RNN, or LSTM layer) will be added, which is also determined by the distribution sampling method introduced above. The second step is to select the position of the newly added layer using the information from the first step and learned features from the encoder network. Briefly, we index each layer in the given network and calculate the probability of inserting the newly added layer. Then, the position to insert can be selected using the distribution sampling method, and the model can be accordingly deepened. As to the output unit amount of the inserted student network layer, we have the following rule. For an inserted student CNN layer, the output channel amount should equal that of its teacher network layer. Note that  we set the kernel size in CNN to three since the functionality of a kernel with any size can be equally represented by such a kernel size \citep{simonyan2014very}. Meanwhile, we set the stride to one when the CNN layer is selected in the first step. Moreover, for an inserted student RNN or LSTM layer, the output unit amount should also be equal to that of its teacher network layer. However, for an inserted student FCN layer, the output unit amount should be equal to the time-series length of the output data for its teacher network layer. For instance, if inserting an FCN layer to the network and its teacher network layer is a CNN layer with an output data size of $168\times 3$, the output unit amount for this student FCN layer is 168. Conversely, if inserting an RNN layer to the network and its teacher network is this same CNN layer, the output unit amount for this student RNN layer is 3.

\subsection*{3.3.4. Pruning Implementation}\label{subsubsection:prune-implementatioin}
This section describes how the pruning action is implemented. Unlike the wider and deeper actions determined by RL, the pruning action is deterministic, derived from the movement pruning method \citep{sanh2020movement}. As introduced in Section \ref{subsection:shrinkage-transformation}, we update the importance scores in each layer along with the corresponding weights. In FCN, RNN, and LSTM, one output neuron is a unit, while in CNN, one output channel is a unit. We create an importance score $\boldsymbol{\lambda}^{(i+1)}$ for each output unit in layer $i$ so that the importance scores of each layer can be represented as a vector that has the same size as the output units. When the selector actor takes the pruning action, we first concatenate the vectors of importance scores from each layer to generate a unified vector $\boldsymbol{\lambda}=\{\boldsymbol{\lambda}^{(2)},\cdots,\boldsymbol{\lambda}^{(n+1)}\}$ (assuming that we have $n$ layers in total). After that, we use Eq. (\ref{eq:top-v}) to obtain the pruning mask $\boldsymbol{m}$, which selects the top $v\%$ important units by setting the importance mask value to 1 and 0 for the rest. In our study, this value is set to $v\% = 90\%$. Once the mask $\boldsymbol{m}$ is determined, we break it into $n$ vectors and each vector represents the mask values of a layer. Hence, we will obtain the mask values for all units in the network. Subsequently, for each layer, we perform the pruning action described in Section \ref{subsection:shrinkage-transformation} to discard the units whose mask values are 0.

\subsection{Net Pool} \label{subsection:net-pool}

A net pool is developed in our proposed IAAS framework to iteratively remove the transformed neural architectures with bad performance so as to improve the quality of the searched architectures. This net pool has a heuristic screening algorithm that is similar to the standard genetic algorithm (GA). In the following, we present the process of how the net pool works. Initially, suppose we have $S_Q$ network structures in the net pool and transform each of them with the proposed framework. After that, all the transformed networks are added to the net pool, and there are 2$S_Q$ network structures in the pool. Next, other $M$ networks are randomly generated and added to the pool. This random generation aims to reduce the possibility of premature phenomena in the results. All the 2$S_Q+M$ networks are evaluated based on their forecasting performance. Moreover, the capacity of our net pool is denoted as $C_Q$ (the maximum number of network architectures in the net pool), which means $C_Q \geq S_Q$ or $M$. Note that we require the initial settings for $S_Q$ and $M$ as $2S_Q+M \geq C_Q$. Therefore, after ranking their forecasting performance from the best to the worst, we only keep the top $C_Q$ network architectures in the net pool. Starting from this search episode, $S_Q$ is equal to $C_Q$. The search process usually needs to take hundreds of episodes. After the search process is finished, we find the best network in the pool as our searched result. It should be noted that the proposed IAAS framework can be regarded as a two-stage operation in each search episode. Briefly, at the first stage, the RL-based network transformation will be implemented to transform each network structure in the net pool. At the second stage, the transformed networks will be re-entered into the net pool and a heuristic algorithm will be implemented to kick off the network structures with bad performance. In the EAS framework \citep{cai2018efficient}, it does not have such a net pool component, and the transformed network will be automatically used for the next-round transformation. This is like using one-fixed length trajectory for RL-based meta-controllers to collect samples. For a large search space, one trajectory usually is not enough. As a comparison, this net pool provides multiple candidates for RL-based meta-controllers to explore the search space. To better describe this heuristic screening algorithm, we present it in Algorithm \ref{alg:pool-management} in Appendix \ref{appx-algorithms} of the online supplement.

The elimination of networks in this heuristic approach happens under two scenarios. In the first one, if the performance of the transformed network $\mathcal{A'}$ is worse than that of the others (all the transformed networks and their parent networks) in the net pool, this network will be eliminated. In the second one, if the performance of both the network $\mathcal{A}$ and its transformed network $\mathcal{A'}$ have worse performance than the others, these two will be eliminated from the net pool, resulting in the termination of this network transformation trajectory. For the first scenario, the network $\mathcal{A}$ will be used for the next-round transformation, and its transformation trajectory will continue. For the second scenario, the transformation trajectory of the network $\mathcal{A}$ will be truncated. As noted, if the computational budget is fixed, the trajectory truncation has been demonstrated in \cite{poiani2023truncating,poiani2024truncating} as an effective method to improve the performance of the RL agent, as this will allocate more computational budget to collect samples for the RL agent at the early stage of interaction with the environment. Note that the computational budget indicates the total number of action times in \cite{poiani2023truncating}. For our IAAS, the computational budget is fixed (e.g., 1000 action times), while the number of trajectories and the length of each trajectory are not fixed. The computational budget is equal to the sum of the lengths of all the trajectories. Moreover, randomly generating one network structure at each iteration is also like providing more early-stage samples to the RL agent. This is because each randomly generated network is the start of a new MDP trajectory.

Notably, the initial net pool is constructed from all network types such as FCN, CNN, RNN, and LSTM with one layer that has four units. In the screening process, we first identify the maximum layer number $(\widetilde{A})$ and the maximum unit number of one layer $(\widetilde{B})$ in the existing networks. And then, we use $\widetilde{A}$ and $\widetilde{B}$ as the upper bounds to randomly generate $M$ networks as mentioned above. In Algorithm \ref{alg:pool-management}, the stopping criterion is used to stop the architecture search process. Hence, if the stopping criterion is not satisfied, the net pool will be dynamically updated. This stopping criterion is problem-dependent. In this work, this criterion is set to a fixed number of search episodes (e.g., 200) to validate the performance of our search scheme.

\section{Numerical Results}\label{section:Numerical-Results}

In this section, we present our results by conducting numerical experiments using the proposed IAAS framework to develop electricity forecasting models in power systems. Our experiments include two streams of forecasting: load forecasting from the electricity consumption side and wind power forecasting from the electricity production side. In the following, we first describe the data information of these two main experiments. Subsequently, we present the experimental settings for the proposed IAAS framework and baseline methods. Next, we demonstrate and discuss the experimental results to validate the advantages of the proposed IAAS framework. Finally, we conduct an ablation study to demonstrate the importance of incorporating the selector actor and net pool components in this framework.

\subsection{Data Description}
We perform our experiments based on the time-series data to build forecasting models of electricity load demand and wind power generation in power systems. For the electricity load forecasting experiment, two publicly-available datasets of Maine (ME) and New Hampshire (NH) for the year 2020 are collected from ISO New England Inc. Similarly, for the wind power forecasting experiment, we utilize two wind farm (WF) datasets from a Chinese wind power company. Due to the fact that the electricity time-series data are typically influenced by various seasons (\citealt{qin2021two,jalali2021novel}), we classify each dataset depending on the season, i.e., spring, summer, autumn, and winter, respectively. Consequently, there are sixteen independent subsets for our experimental evaluations. The detailed data description can be found in   Appendix \ref{appx:Data details} of the online supplement.

\subsection{Experimental Settings and Performance Metrics}\label{subsection:Experimental-settings-and-performance-metrics}

Comparison experiments are imperative for the demonstration of forecasting model performance. In this study, we exploit three classical ML methods, i.e., support vector regression (SVR), random forest (RF), and ridge regression (RR), as baselines. Moreover, we use the following seven DL models to showcase the advantages of our proposed IAAS framework. Firstly, we make use of three traditional DL methods, which involve CNN, LSTM network, and CNN+LSTM network. Secondly, we utilize three proposed networks in the literature, where ResNet and ResNetPlus were derived from \cite{chen2018short} and deep adaptive input normalization (DAIN) was introduced by \cite{passalis2019deep}. Briefly, \cite{chen2018short} proposed two residual neural networks, where ResNetPlus was developed based on ResNet by utilizing a side block technique; \cite{passalis2019deep} introduced the DAIN method for the time-series data in order to build a better forecasting model. Thirdly, we employ the latest SNAS framework \citep{chen2021scale}, which targets the MTF model development with the NAS technique. We put the details of experiment settings in   Appendix \ref{appx:experimental-setting} of the online supplement. Lastly, after conducting the trial and error simulation, for IAAS, we determine to use 256 as the batch size, 200 as the maximum search episodes, and 50 epochs for each search episode. The number of search episodes directly has an effect on the search quality of network structures.
Therefore, before the following experiments, we conduct a sensitivity analysis to investigate whether the usage of 200 episodes is reasonable. The results are presented in   Appendix \ref{sensitivity} of the online supplement.

Based on the existing literature in both load forecasting and wind power forecasting (e.g., \citealt{ccevik2019new,chen2018short,chen2023novel}), we use two evaluation metrics, namely RMSE and MAE, to quantify the performance of forecasting results:
\begin{equation} 
RMSE=\sqrt{\frac1{N}\sum^N_{i=1}{{(Y_i-{\hat{Y}}_i)}^{2}}}, 
\label{eq:rmse} 
\end{equation} 
\begin{equation}
MAE=\frac1{N}\sum^N_{i=1}{{|}Y_i-{\hat{Y}}_i{|}}, 
\label{eq:mae} 
\end{equation} 
where $Y$ is the actual value and $\hat{Y}$ is the predicted value, and $N$ is the number of forecasting time points.

\subsection{Results and Analysis}\label{subsection:Results-and-analysis}

\subsection*{4.3.1. Electricity Load Demand Forecasting Experiments}\label{subsubsection:load-experiments}
Prior to the discussion of electricity load forecasting performance, we analyze the experimental results derived from our proposed IAAS framework regarding the network architecture. Given that each dataset is split by four seasons, we thus obtain eight network structures of load forecasting models as presented in Table \ref{tab:load-structure}. To describe the architecture of each model, we leverage an example of the ME-autumn case for illustration. ME-autumn model is a three-layer network model: the first layer is a CNN layer with 15 output channels, and the second and third layers are both FCN layers with 14 and 24 output units, respectively. As observed, all these eight network structures are different from each other (the largest one has seven layers (see ME-winter case) and the smallest ones have two layers (see NH-spring case and NH-autumn case)). Since all the training data of these eight cases are different, the model structure differences indicate that the proposed IAAS framework is likely to automatically adapt to the given dataset by generating an appropriate structure. It is worth noting that different network structures could generate similar performance. Observing Fig. \ref{fig:data-description}, the ME-spring data pattern is similar to the NH-spring data pattern; however, the network structures of the ME-spring case and NH-spring case are quite different. The reason could be that the internal mechanism of the IAAS framework cannot guarantee the globally optimal search due to the limitations of RL and heuristic screening algorithm. 

\begin{table}[H]
\caption{Network structures of load demand forecasting experiments using IAAS framework
\label{tab:load-structure}}

\resizebox{\textwidth}{!}{\small
\begin{tabular}{p{0.18\columnwidth}p{0.8\columnwidth}} 
\hline
Cases & Network structures \\ \hline
ME-spring & rnn-3${\to}$rnn-3${\to}$rnn-3${\to}$rnn-3${\to}$rnn-3${\to}$rnn-3${\to}$conv-24${\to}$fc-16 \\ \hline
ME-summer & fc-3${\to}$fc-3${\to}$fc-3${\to}$fc-3${\to}$fc-3${\to}$conv-6${\to}$fc-5${\to}$fc-8${\to}$fc-8 \\ \hline
ME-autumn & conv-15${\to}$fc-14${\to}$fc-24 \\ \hline
ME-winter & conv-3${\to}$rnn-3${\to}$rnn-3${\to}$rnn-3${\to}$conv-3${\to}$rnn-3${\to}$rnn-3${\to}$rnn-3${\to}$rnn-1${\to}$rnn-1${\to}$rnn-1${\to}$fc-144 \\ \hline
NH-spring & conv-3${\to}$fc-12 \\ \hline
NH-summer & rnn-4${\to}$rnn-4${\to}$fc-16${\to}$rnn-16${\to}$rnn-16${\to}$rnn-16${\to}$rnn-16${\to}$fc-16${\to}$rnn-16 \\ \hline
NH-autumn & conv-4${\to}$fc-16 \\ \hline
NH-winter & fc-17${\to}$conv-64${\to}$fc-2${\to}$fc-7${\to}$conv-67${\to}$conv-5${\to}$fc-18${\to}$rnn-36${\to}$fc-64 \\ \hline
\end{tabular}
}
\end{table}

\begin{figure}
    \centering
    \includegraphics[width=\textwidth]{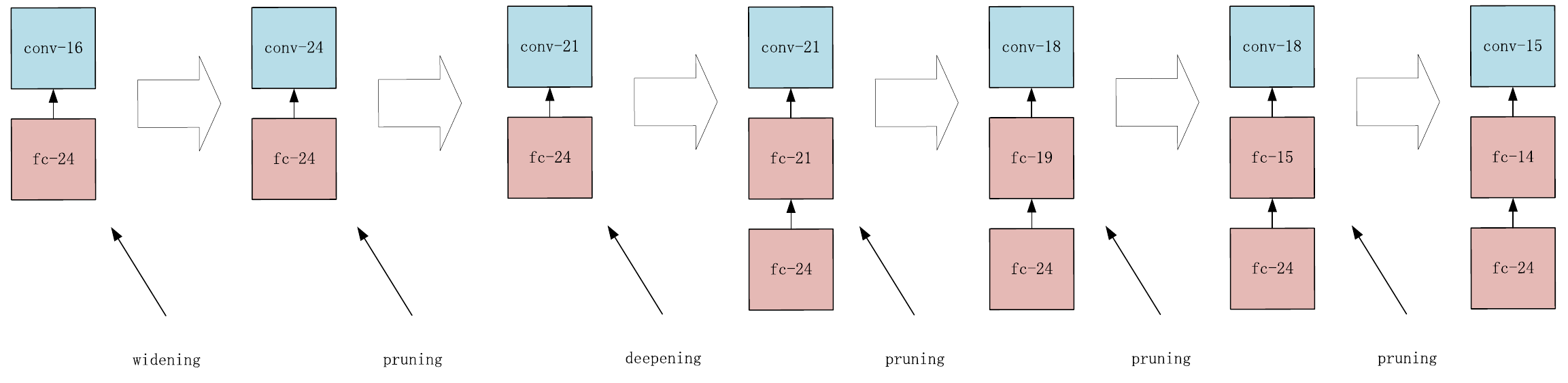}
    \caption{
    Network transformation illustration (ME-autumn case)\label{fig:transformation illustration}}
\end{figure}

Fig. \ref{fig:transformation illustration} depicts the network transformation procedures of the ME-autumn case using the IAAS framework. As shown, the final network structure contains one CNN layer and two FCN layers. As introduced in Section \ref{subsection:net-pool}, we initialize the networks in net pool with one LSTM layer, one CNN layer, one RNN layer, and one FCN layer. Each layer has four units. Since we have a random network generation step as described in Section \ref{subsection:net-pool}, this example indicates that, at a specific search episode, the randomly generated network (conv-16 ${\to}$ fc-24) performs better than all existing networks. Therefore, this randomly generated network is kept in the net pool. After six transformations with widening, deepening and pruning, this network is transformed to the one with the structure of ``conv-15${\to}$ fc-14 ${\to}$ fc-24", achieving the best forecasting performance in the searching process.

The forecasting accuracy results of our proposed IAAS framework and ten baseline models or methods with two evaluation metrics are presented in Table \ref{tab:load-performance}. Moreover, for the load forecasting accuracy evaluation, we also make use of root mean squared logarithmic error (RMSLE) as another evaluation metric since an electricity company usually prefers to overshoot the electricity, and the undershooting may not be able to serve the customers well.
\begin{equation}
RMSLE=\sqrt{\frac1{N}\sum^N_{i=1}{{(\log(Y_i+1)-\log({\hat{Y}}_i+1))}^{2}}} 
\label{eq:rmsle}
\end{equation}

\begin{table}[!htpb]
\caption{Forecasting performance of electricity load forecasting experiments\label{tab:load-performance}}
{
\resizebox{\linewidth}{!}{\small
\begin{tabular}{ccccccccccccc}
\hline
Case & Metrices & CNN-LSTM & CNN & LSTM & SVR & RF & RR & ResNet & ResNetPlus & DAIN & SNAS & IAAS\_LoadNet \\ \hline
\multirow{3}{*}{ME-spring} & RMSE & 70.559 & 74.583 & 116.771 & 115.912 & 69.665 & 68.557 & 64.038 & 64.585 & 65.686 & 62.866 & \textbf{61.807} \\ \cline{2-13} 
 & MAE & 54.903 & 59.998 & 98.872 & 97.220 & 58.396 & 54.187 & 51.105 & 51.228 & 53.023 & 50.918 & \textbf{48.879} \\ \cline{2-13} 
 & RMSLE & 0.055 & 0.059 & 0.092 & 0.099 & 0.055 & 0.054 & 0.050 & 0.051 & 0.052 & 0.050 & \textbf{0.049} \\ \hline
\multirow{3}{*}{ME-summer} & RMSE & 152.562 & 153.304 & 168.969 & 203.196 & 152.306 & 120.076 & 75.901 & 75.110 & 83.615 & 85.378 & \textbf{69.154} \\ \cline{2-13} 
 & MAE & 120.538 & 122.502 & 133.090 & 172.666 & 116.565 & 99.416 & \textbf{56.900} & 57.567 & 66.371 & 64.145 & 58.187 \\ \cline{2-13} 
 & RMSLE & 0.112 & 0.115 & 0.124 & 0.163 & 0.110 & 0.092 & 0.055 & 0.056 & 0.063 & 0.062 & \textbf{0.052} \\ \hline
\multirow{3}{*}{ME-autumn} & RMSE & 64.949 & 90.953 & 111.229 & 156.521 & 76.274 & 64.901 & 49.906 & 48.775 & 55.913 & 59.123 & \textbf{38.483} \\ \cline{2-13} 
 & MAE & 49.159 & 76.833 & 91.189 & 124.190 & 59.291 & 49.470 & 39.064 & 37.165 & 43.572 & 49.721 & \textbf{30.526} \\ \cline{2-13} 
 & RMSLE & 0.051 & 0.076 & 0.087 & 0.139 & 0.061 & 0.051 & 0.040 & 0.038 & 0.045 & 0.050 & \textbf{0.031} \\ \hline
\multirow{3}{*}{ME-winter} & RMSE & 91.504 & 107.586 & 129.708 & 150.629 & 111.414 & 117.403 & 89.133 & 98.878 & 91.950 & 106.098 & \textbf{85.386} \\ \cline{2-13} 
 & MAE & 75.127 & 83.435 & 102.449 & 125.602 & 92.020 & 94.374 & 74.332 & 80.917 & 70.789 & 88.392 & \textbf{68.773} \\ \cline{2-13} 
 & RMSLE & 0.067 & 0.078 & 0.094 & 0.110 & 0.081 & 0.087 & 0.065 & 0.073 & 0.067 & 0.078 & \textbf{0.062} \\ \hline
\multirow{3}{*}{NH-spring} & RMSE & 80.928 & 94.490 & 110.839 & 133.012 & 82.394 & 81.885 & 75.690 & 76.451 & 77.602 & 86.761 & \textbf{68.749} \\ \cline{2-13} 
 & MAE & 65.067 & 76.260 & 85.261 & 110.609 & 65.379 & 63.703 & 61.054 & 60.529 & 63.256 & 69.859 & \textbf{53.759} \\ \cline{2-13} 
 & RMSLE & 0.067 & 0.079 & 0.089 & 0.119 & 0.066 & 0.067 & 0.062 & 0.064 & 0.063 & 0.074 & \textbf{0.058} \\ \hline
\multirow{3}{*}{NH-summer} & RMSE & 170.803 & 191.243 & 198.392 & 263.721 & 196.841 & \textbf{139.446} & 154.906 & 150.601 & 161.118 & 167.340 & 144.016 \\ \cline{2-13} 
 & MAE & 134.767 & 155.533 & 149.884 & 209.686 & 143.986 & \textbf{105.739} & 117.775 & 114.732 & 131.704 & 133.334 & 114.45 \\ \cline{2-13} 
 & RMSLE & 0.120 & 0.134 & 0.136 & 0.189 & 0.134 & \textbf{0.093} & 0.106 & 0.102 & 0.113 & 0.111 & 0.096 \\ \hline
\multirow{3}{*}{NH-autumn} & RMSE & 106.226 & 133.442 & 130.408 & 207.988 & 110.012 & 98.570 & 94.535 & 86.676 & 84.303 & 78.102 & \textbf{62.468} \\ \cline{2-13} 
 & MAE & 87.791 & 106.388 & 107.390 & 165.998 & 78.383 & 75.753 & 69.169 & 65.660 & 64.958 & 62.730 & \textbf{45.985} \\ \cline{2-13} 
 & RMSLE & 0.088 & 0.113 & 0.104 & 0.186 & 0.086 & 0.079 & 0.076 & 0.071 & 0.069 & 0.068 & \textbf{0.052} \\ \hline
\multirow{3}{*}{NH-winter} & RMSE & 84.863 & 209.698 & 112.714 & 135.381 & 92.193 & 97.517 & 95.369 & 96.150 & 88.595 & 103.085 & \textbf{82.911} \\ \cline{2-13} 
 & MAE & 69.722 & 184.262 & 91.061 & 108.405 & 74.449 & 78.587 & 78.075 & 80.440 & 72.134 & 83.888 & \textbf{65.506} \\ \cline{2-13} 
 & RMSLE & 0.064 & 0.157 & 0.085 & 0.097 & 0.068 & 0.075 & 0.073 & 0.073 & 0.066 & 0.080 & \textbf{0.062} \\ \hline
\multirow{3}{*}{Average} & RMSE & 102.799 & 131.912 & 134.879 & 170.795 & 111.387 & 98.544 & 87.435 & 87.153 & 88.598 & 93.594 & \textbf{76.622} \\ \cline{2-13} 
 & MAE & 82.134 & 108.151 & 107.400 & 139.297 & 86.059 & 77.654 & 68.434 & 68.530 & 70.726 & 75.373 & \textbf{60.758} \\ \cline{2-13} 
 & RMSLE & 0.078 & 0.101 & 0.101 & 0.138 & 0.083 & 0.075 & 0.066 & 0.066 & 0.067 & 0.072 & \textbf{0.058} \\ \hline
\end{tabular}
}
}

\end{table}

\begin{figure}[!htpb]
    \begin{subfigure}[b]{0.5\textwidth}
        \centering
	    \includegraphics[width=\textwidth]{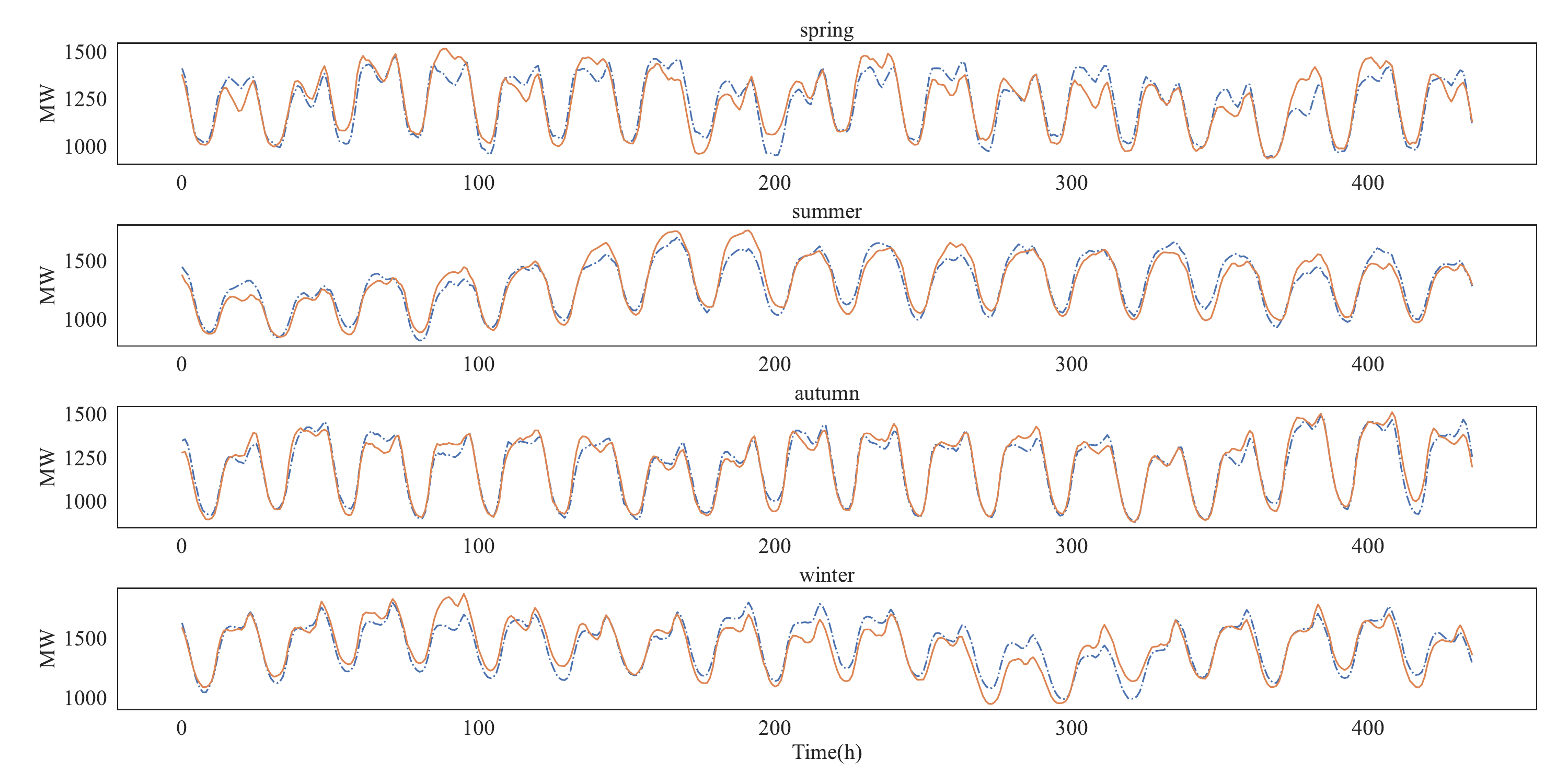}
        \caption*{\footnotesize{(a) ME case}}
    \end{subfigure}
    \hfill
    \begin{subfigure}[b]{0.5\textwidth}
        \centering
		\includegraphics[width=\textwidth]{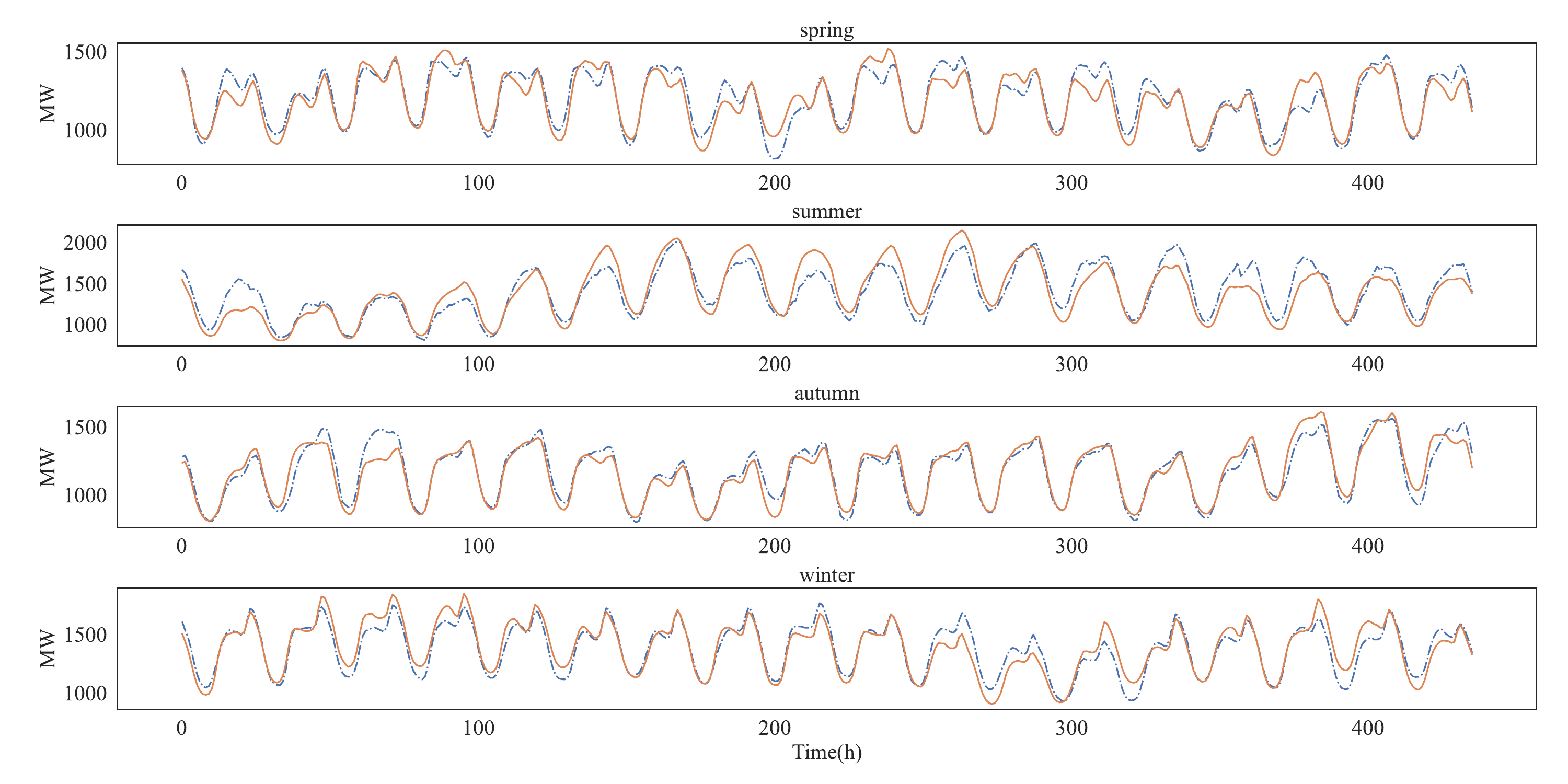}
        \caption*{\footnotesize{(b) NH case}}
    \end{subfigure}
\caption{ Comparison results between predicted electricity load and actual electricity load in eight testing sets\\
{\small \textit{Note.} the red solid line represents the actual load values, and the blue dashed line represents the predicted values.}
    \label{fig:performance-load}}
    
\end{figure}

We name the acquired load forecasting models using the IAAS framework as IAAS\_LoadNet. We mark the best of these models as boldface in each case. As we can observe, the obtained IAAS\_LoadNet models outperform the baseline models in most cases. Specifically, in terms of RMSE and RMSLE, IAAS\_LoadNet models perform the best in almost all cases, except in the case of NH-summer; as for MAE, IAAS\_LoadNet models exhibit the best performance in six out of eight cases. However, IAAS\_LoadNet models have close forecasting accuracies compared to the best models in the cases of ME-summer and NH-summer. Furthermore, we present the average forecasting accuracy of eight cases at the bottom row of Table  \ref{tab:load-performance} and determine that the IAAS\_LoadNet achieves much better performance than all the baseline models on average. In detail, the RMSE, MAE and RMSLE for the IAAS\_loadNet model are 76.622, 60.758, and 0.058, respectively, and they are higher than the second-best model with 12.1\% (RMSE), 11.2\% (MAE), and 12.1\% (RMSLE), individually. In addition, we plot the predicted and actual loads for the test datasets of both ME and NH cases in Fig. \ref{fig:performance-load}. As witnessed, our proposed IAAS framework can match the actual loads in 24h-ahead points very well. 

\begin{figure}[!htpb]
    \centering
    \includegraphics[scale=0.4]{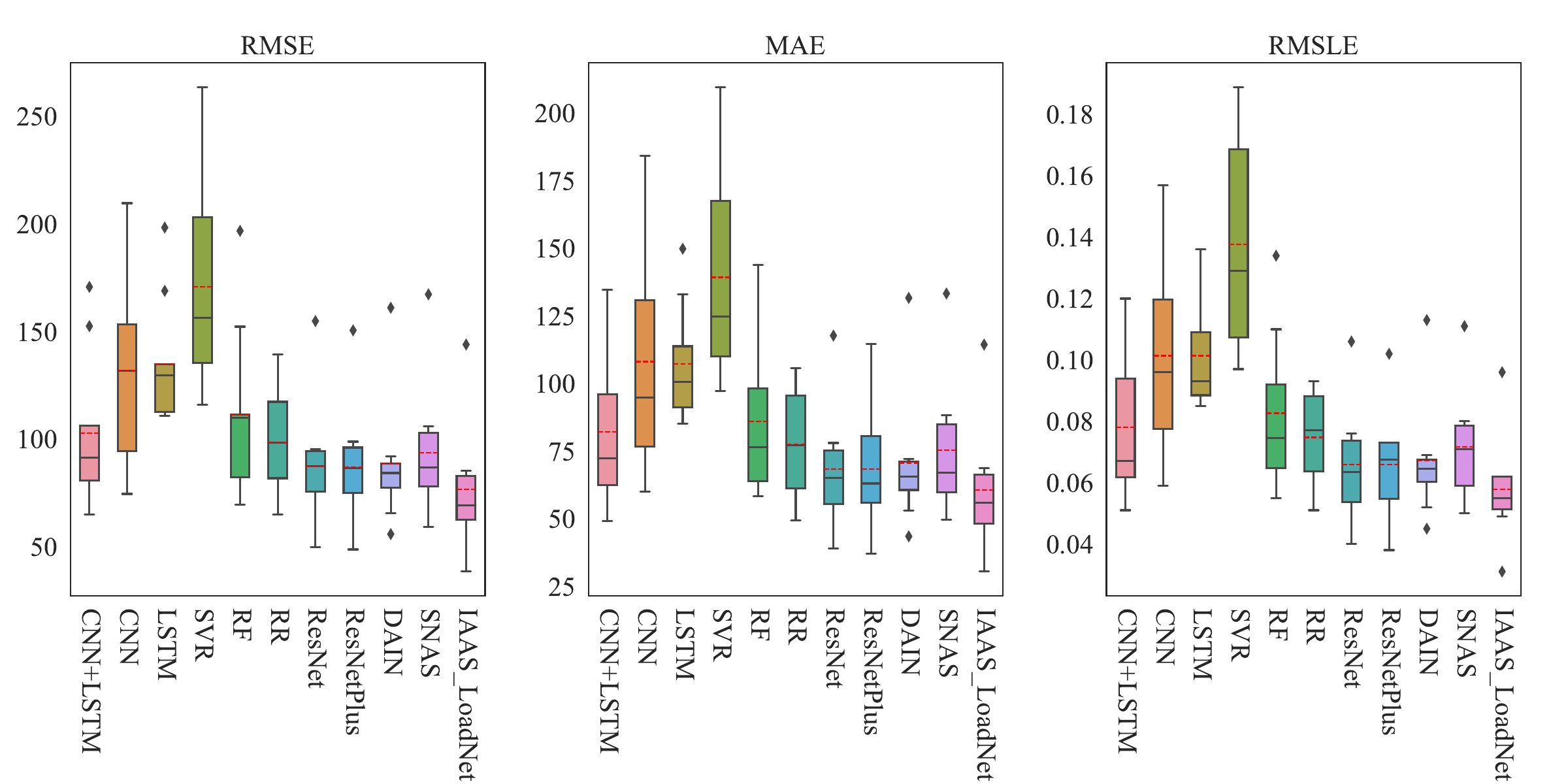}
    \caption{Boxplot of load demand forecasting accuracies
    \label{fig:box-load}}
\end{figure}

To further explore the stability of forecasting performance in the load forecasting experiments, we use the boxplot to demonstrate the variations in the forecasting accuracies, which are shown in Fig. \ref{fig:box-load}. As we can observe, all eleven models lead to various stabilities in terms of the three metrics; however, our proposed IAAS\_LoadNet achieves competitively stable accuracies in terms of RMSE, MAE, and RMSLE (see the box body size). Such results also denote the necessity of using our IAAS framework to design a network architecture for various datasets. Moreover, we use the red dashed line to indicate the average forecasting accuracy of each model in each box in terms of each metric. It should be noted that we expect the forecasting models to have the best average forecasting accuracy and low variance of forecasting accuracies. As seen, the IAAS framework can achieve this goal better than the baseline models. Based on this result and the forecasting accuracy results, we conclude that the proposed IAAS framework is able to adapt to the training data automatically for the development of load forecasting models.

\subsection*{4.3.2. Wind Power Forecasting Experiment}\label{subsubsection:wind-power-experiment}
As described above, we use two WF datasets considering the seasonal effects of conducting the wind power forecasting experiments, which constitutes eight cases in total. Table \ref{tab:windpower-structure} presents the obtained network structures for each case after implementing our proposed IAAS framework. We name the obtained network models as IAAS\_WindNet. As shown in the table, all the cases have different model structures. We present the forecasting accuracies of wind power with the two evaluation metrics in Table \ref{tab:windpower-performance}. It can be observed that the developed IAAS\_WindNet models have shown the best performance in all cases in terms of RMSE and MAE. Averagely, the RMSE and MAE for the IAAS\_WindNet model are 4.107 and 3.132, respectively, while those for the second-best model are 4.669 and 3.585, respectively. As presented in the table, our IAAS\_WindNet models show better performance than the second-best models, with 12.0\% (RMSE) and 12.6\% (MAE), separately. Note that we do not use RMSLE as the accuracy evaluation metric for wind power forecasting since RMSLE penalizes underprediction more than overprediction. In other words, when using RMSLE, generating more wind power than prediction will receive less penalty. On the electricity supply side, it pays more attention to the power system stability, which means producing more wind power than expected creates more challenges to the stability of the power system than producing less wind power than expected. Therefore, it is inappropriate to use RMSLE as the error measurement for wind power forecasting accuracy.

\begin{figure}[!htpb]
    \begin{subfigure}[b]{0.5\textwidth}
        \centering
	    \includegraphics[width=\textwidth]{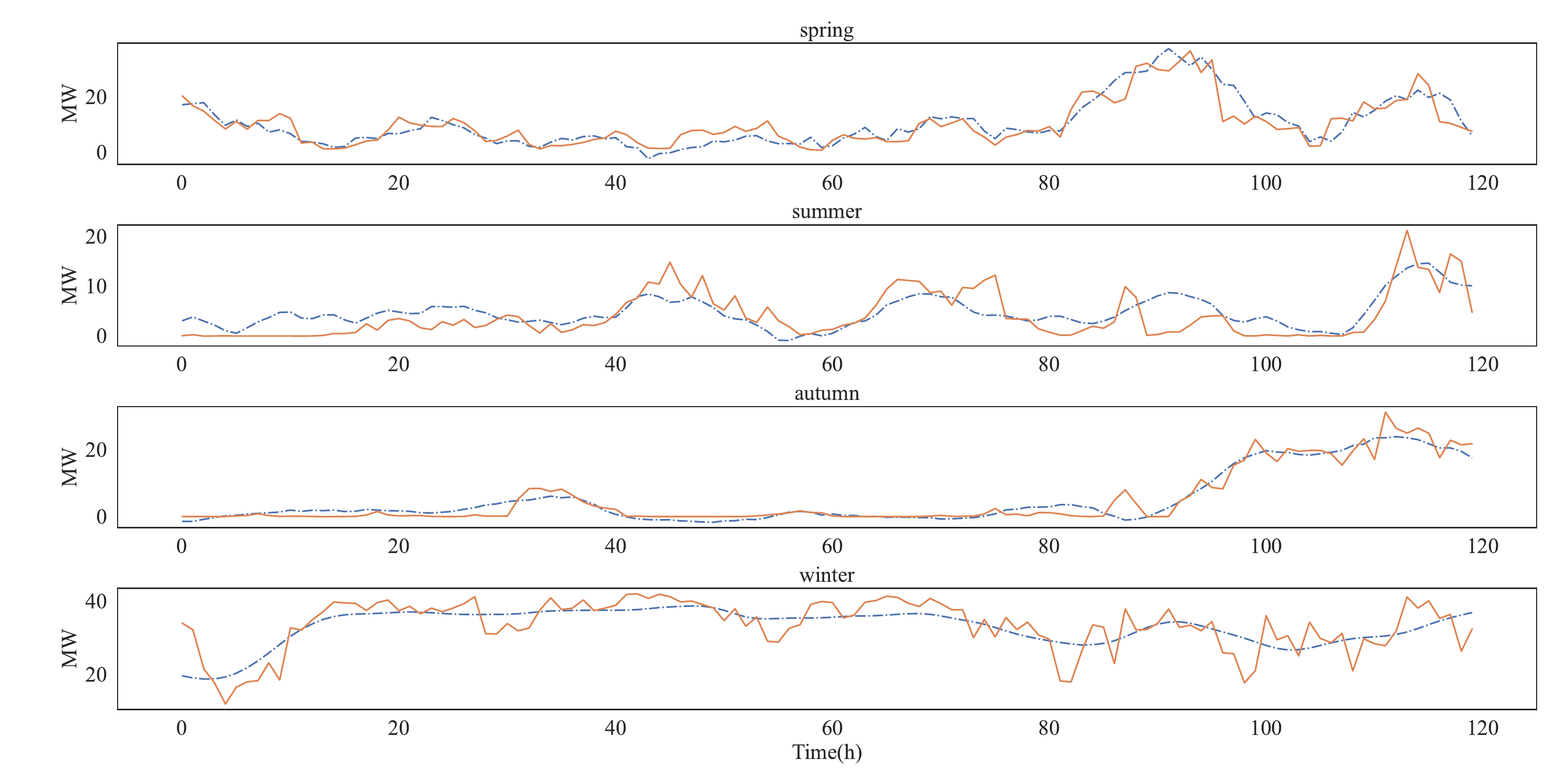}
        \caption*{\footnotesize{(a) WF1 case}}
    \end{subfigure}
    \hfill
    \begin{subfigure}[b]{0.5\textwidth}
        \centering
		\includegraphics[width=\textwidth]{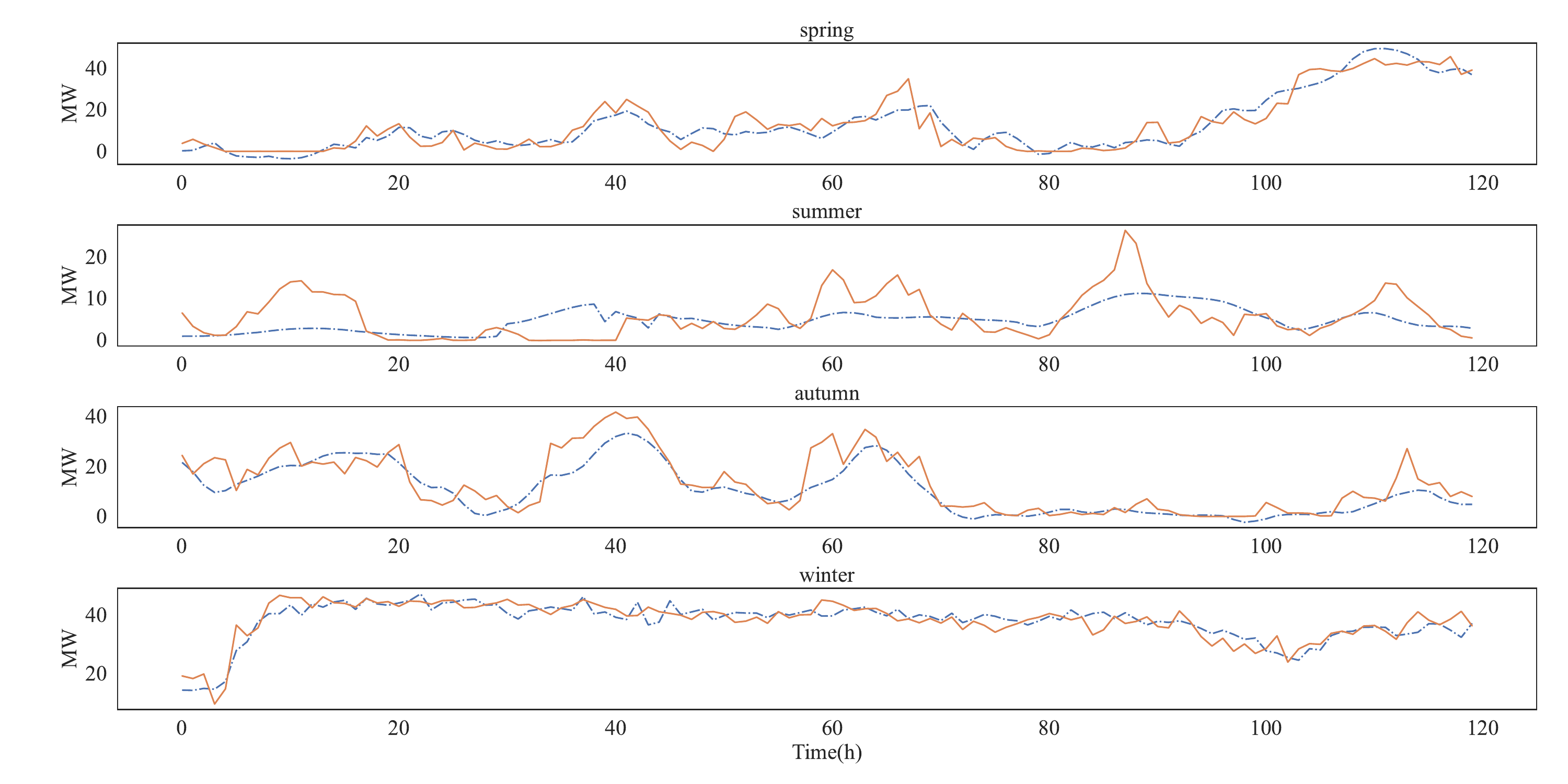}
        \caption*{\footnotesize{(b) WF2 case}}
    \end{subfigure}
\caption{ Comparison results between predicted wind power and actual wind power in eight testing sets\\
{\small \textit{Note.} the red solid line represents the actual wind power, and the dashed blue line represents the predicted wind power.}
    \label{fig:performance-wind}}
    
\end{figure}

\begin{figure}[!htpb]
    \centering
    \includegraphics[scale=0.4]{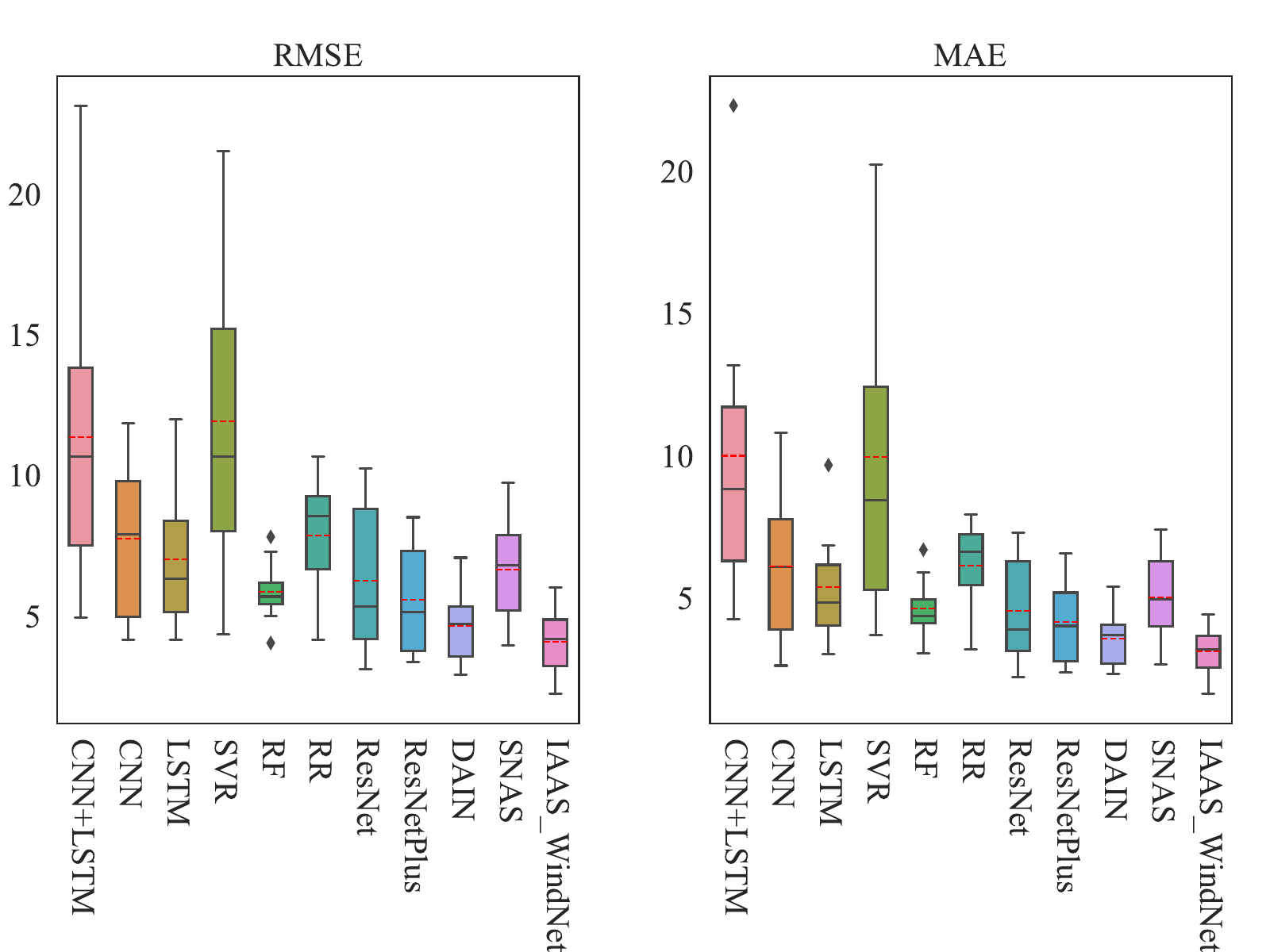}
    
    \caption{Boxplot of wind power forecasting accuracies
    \label{fig:box-wind}}
\end{figure}

Consistent with the load forecasting experiments, we compare the predicted and actual wind power for both WF datasets in Fig. \ref{fig:performance-wind} and find out IAAS\_WindNet is likely to generate an accurate 24h-ahead forecasting wind power. However, it should be noted that due to the more uncertain and intermittent nature of wind power compared to that of electricity load, sometimes, it is challenging to accurately forecast the 24h-ahead wind power generation as shown in Fig. \ref{fig:performance-wind}. We also draw the boxplot to evaluate the stability of the forecasting results in these two evaluation metrics in Fig. \ref{fig:box-wind}. As shown in the figure, the IAAS\_WindNet also achieves a competitively stable performance (see the box body size) compared to the others. The red dashed line in the box indicates the IAAS models have the best performance on average than the others in terms of these two metrics. Hence, we conclude that the proposed IAAS framework can obtain better forecasting models for the 24h-ahead wind power forecasting scenario as compared to the existing models or methods. 
\begin{table}
\caption{Network structures of wind power forecasting experiments using IAAS framework\label{tab:windpower-structure}}
{
    \resizebox{\textwidth}{!}{\small
\begin{tabular}{p{0.2\columnwidth}p{0.8\columnwidth}}
\hline
Cases & Network structure \\ \hline
WF1-spring & rnn-102${\to}$fc-102${\to}$rnn-102${\to}$fc-102${\to}$fc-102${\to}$fc-102${\to}$rnn-102${\to}$fc-102${\to}$fc-110${\to}$fc-108${\to}$fc-107${\to}$rnn-144${\to}$rnn-144${\to}$rnn-144${\to}$rnn-73${\to}$fc-4 \\ \hline
WF1-summer & conv-64${\to}$fc-144 \\ \hline
WF1-autumn & conv-8${\to}$fc-80 \\ \hline
WF1-winter & fc-2${\to}$conv-8${\to}$rnn-13${\to}$fc-32 \\ \hline
WF2-spring & conv-144${\to}$fc-108 \\ \hline
WF2-summer & rnn-9${\to}$lstm-63${\to}$fc-48 \\ \hline
WF2-autumn & lstm-80${\to}$lstm-80${\to}$conv-80${\to}$fc-144${\to}$conv-144${\to}$conv-144${\to}$conv-144${\to}$rnn-144${\to}$fc-144 \\ \hline
WF2-winter & rnn-30${\to}$fc-1${\to}$conv-76${\to}$fc-2${\to}$conv-3${\to}$fc-11${\to}$lstm-118${\to}$rnn-29${\to}$fc-32 \\ \hline
\end{tabular}
    }
}
\end{table}

\begin{table}[!htpb] 
\centering
\caption{Forecasting accuracy results of wind power forecasting experiments\label{tab:windpower-performance}}
{
\resizebox{\linewidth}{!}{\small
\begin{tabular}{ccccccccccccc}
\hline
Case                        & Metrices & CNN-LSTM & CNN    & LSTM   & SVR    & RF    & RR     & ResNet & ResNetPlus     & DAIN  & SNAS  & IAAS\_WindNet \\ \hline
\multirow{2}{*}{WF1-spring} & RMSE     & 9.096    & 6.677  & 8.819  & 8.744  & 5.818 & 9.924  & 4.311  & 4.341          & 4.407 & 5.506 & \textbf{3.901}         \\ \cline{2-13} 
                            & MAE      & 7.967    & 5.239  & 5.949  & 7.539  & 4.478 & 7.945  & 3.417  & 3.366          & 3.412 & 4.472 & \textbf{2.996}         \\ \hline
\multirow{2}{*}{WF1-summer} & RMSE     & 4.951    & 4.154  & 4.168  & 4.358  & 5.646 & 4.807  & 3.816  & 3.842          & 3.615 & 3.956 & \textbf{3.290}         \\ \cline{2-13} 
                            & MAE      & 4.458    & 3.084  & 3.027  & 3.718  & 4.264 & 4.051  & 2.837  & 2.860          & 2.756 & 2.803 & \textbf{2.614}         \\ \hline
\multirow{2}{*}{WF1-autumn} & RMSE     & 8.263    & 4.301  & 8.247  & 8.603  & 4.051 & 4.161  & 3.113  & 3.539          & 2.918 & 4.296 & \textbf{2.237}         \\ \cline{2-13} 
                            & MAE      & 6.932    & 2.634  & 6.853  & 5.589  & 3.063 & 3.210  & 2.219  & 2.396          & 2.353 & 2.686 & \textbf{1.633}         \\ \hline
\multirow{2}{*}{WF1-winter} & RMSE     & 14.327   & 9.132  & 5.836  & 19.597 & 7.821 & 8.554  & 8.635  & 8.088          & 5.271 & 7.384 & \textbf{4.500}         \\ \cline{2-13} 
                            & MAE      & 13.189   & 7.509  & 4.862  & 18.450 & 6.707 & 7.063  & 6.050  & 5.974          & 3.995 & 6.094 & \textbf{3.427}         \\ \hline
\multirow{2}{*}{WF2-spring} & RMSE     & 13.656   & 11.862 & 6.836  & 13.758 & 5.561 & 7.283  & 6.146  & 5.948          & 5.600 & 7.661 & \textbf{4.922}         \\ \cline{2-13} 
                            & MAE      & 11.234   & 8.542  & 4.843  & 10.441 & 4.660 & 5.947  & 4.367  & 4.693          & 4.170 & 5.429 & \textbf{3.953}         \\ \hline
\multirow{2}{*}{WF2-summer} & RMSE     & 5.237    & 5.202  & 5.195  & 6.307  & 5.753 & 10.672 & 10.251 & 7.057          & 5.034 & 6.235 & \textbf{4.870}         \\ \cline{2-13} 
                            & MAE      & 4.275    & 4.171  & 4.145  & 4.460  & 4.251 & 6.630  & 7.093  & 4.941          & 4.043 & 4.402 & \textbf{3.591}         \\ \hline
\multirow{2}{*}{WF2-autumn} & RMSE     & 12.238   & 9.202  & 12.003 & 12.595 & 7.288 & 8.550  & 9.357  & 8.518          & 7.082 & 9.754 & \textbf{6.031}         \\ \cline{2-13} 
                            & MAE      & 9.706    & 6.988  & 9.681  & 9.364  & 5.917 & 6.628  & 7.314  & 6.585          & 5.405 & 7.419 & \textbf{4.439}         \\ \hline
\multirow{2}{*}{WF2-winter} & RMSE     & 23.142   & 11.559 & 5.038  & 21.512 & 5.005 & 9.044  & 4.551  & 3.374          & 3.424 & 8.570 & \textbf{3.104}         \\ \cline{2-13} 
                            & MAE      & 22.316   & 10.811 & 3.716  & 20.240 & 3.772 & 7.779  & 3.242  & 2.589          & 2.543 & 6.939 & \textbf{2.402}         \\ \hline
\multirow{2}{*}{Average}    & RMSE     & 11.364   & 7.761  & 7.018  & 11.934 & 5.868 & 7.874  & 6.273  & 5.588          & 4.669 & 6.670 & \textbf{4.107}         \\ \cline{2-13} 
                            & MAE      & 10.010   & 6.122  & 5.385  & 9.975  & 4.639 & 6.157  & 4.567  & 4.176          & 3.585 & 5.031 & \textbf{3.132}         \\ \hline
\end{tabular}
}
}

\end{table}

\subsection*{4.3.3. Further Analysis}
As demonstrated above, RL has achieved outstanding performance in network transformation control. However, the appropriateness of using RL in the IAAS framework requires further analysis. Therefore, we follow \cite{cai2018efficient} to make use of the random search algorithm to demonstrate the advantages of the selected RL algorithm from three perspectives, namely the forecasting accuracy in the search process, the average training time per epoch, and the number of average parameters of each obtained network. We put such a comparison analysis in   Appendix \ref{time analysis} of the online supplement. The results indicate the selected RL algorithm is more efficient in achieving better RMSE forecasting accuracy. Moreover, the average training time per epoch of the selected RL algorithm is much smaller than that of the random search algorithm. Lastly, using the selected RL algorithm can result in a smaller number of the average parameters for each obtained model than using the random search algorithm. All in all, we believe it is appropriate to use the selected RL algorithm in the proposed IAAS framework.

In addition, the load forecasting and wind power forecasting tasks in practice usually are based on the rolling-window framework but not set as above. To better showcase the performance of the proposed IAAS framework practically, we conduct another experiment to compare the sequence-2-sequence (seq2seq) forecasting results in  Appendix \ref{TransformerCompare} of the online supplement between the IAAS framework and two existing transformer models from \cite{vaswani2017attention} and \cite{zhou2021informer}. The results demonstrate that under the seq2seq setting, the IAAS models still have better forecasting performance than the existing transformer models in terms of evaluation metrics used above. Such results also indicate the broader impact of the proposed IAAS framework in real applications.

Lastly, we investigate how the heuristic mechanism works in the net pool. We develop three variants for demonstration. The first one is without the heuristic approach. Specifically, we initialize five networks in the net pool, transform each of them, and replace the networks in the net pool with these five transformed networks. Such a variant follows the scheme of the fixed-length transformation trajectory used in EAS \citep{cai2018efficient}. The second one is without randomly generating one network at each episode. The third one considers the trajectory truncation manner. Particularly, if there are five networks in net pool, there will be ten networks after transformation. At this time, we randomly eliminate five networks and use the rest for the further transformation. We place the results in   Appendix \ref{appendix I} of the online supplement. As seen, our IAAS method generally achieves the better performance after comparing with each of these three variants. This result indicates that 1) this heuristic mechanism is effective in improving the quality of searching a neural architecture; 2) randomly generating one network at each iteration is important for RL agent; 3) the network elimination strategy in the heuristic algorithm is likely to truncate the network transformation trajectory appropriately.

\subsection{Ablation Studies}\label{subsection:Ablation-studies}

As discussed in Section \ref{section:IAAS-framework}, the primary differences between the IAAS and EAS frameworks are that the IAAS firstly utilizes a selector actor to intelligently determine how to transform the network instead of any manual process and secondly employs a network pool to improve the quality of the searched neural architectures. To demonstrate the advantages of these two novel features, we conduct ablation experiments in this subsection with three scenarios, which are based on three variants of the IAAS framework. The first variant is named IAAS\_s, which indicates that the IAAS framework does not have the selector actor component. The second variant is named IAAS\_n, which denotes the IAAS framework does not have the net pool component. The third variant is named IAAS\_sn, which connotes that the IAAS framework does not have both selector actor and net pool components. It should be noted that for IAAS\_s and IAAS\_sn, we do not consider the manual setting that RL actors take four wider actions and five deeper actions as conducted in EAS \citep{cai2018efficient} for each search episode. Instead, we consider to randomly select an action of ``widening", ``deepening", ``pruning and ``unchanged" for these two variants in each search episode. A random selection strategy might avoid a large network structure when the number of search episodes is large. Keeping the other hyperparameter settings the same, we obtain the forecasting accuracy results, which are presented in Tables \ref{tab:ablation-experiments} and \ref{tab:wind-power-ablation-experiments} in Appendix \ref{appx:Ablation study details} of the online supplement.

Based on Tables \ref{tab:ablation-experiments} and \ref{tab:wind-power-ablation-experiments}, we can conclude that (i) comparing IAAS to IAAS\_s, selector agent generally can improve the model performance, which indicates the advantage of RL-based control in the selector agent; (ii) comparing IAAS to IAAS\_n, net pool component can significantly improve the model performance, which demonstrates the benefits of enlarging the search space with different network candidates; (iii) comparing IAAS to IAAS\_sn, both of the selector and net pool component can jointly improve the model performance more than any of the other two variants. Such results denote the necessity of designing the selector and net pool components in IAAS framework.

\section{Conclusion}\label{section:conclusion}

Modern power systems require a real-time balance between electricity consumption and electricity production for their secure and smooth operation. To reduce the uncertainties and intermittencies from both the production and consumption sides, electricity forecasting has become one of the most effective methods. Thus far, most of the current electricity forecasting research studies have focused on applying DL techniques in constructing forecasting models. Even though these forecasting models have demonstrated outstanding performance, they are developed primarily based on the designer's inherent knowledge and experience without explaining whether the proposed neural architectures are optimized or not. Moreover, these models cannot self-adjust to various datasets automatically. Even though the NAS technique can automate the architecture search process by using optimization algorithms, most of the current techniques have computational issues, and only a few NAS techniques have been applied to the energy sector. Considering these, we propose an IAAS framework to search the high-quality neural architecture for the electricity forecasting model development. 

The IAAS framework builds on and significantly extends the EAS framework \citep{cai2018efficient}. Firstly, we develop the function-preversing transformation methods for RNN and LSTM considering their advantages in extracting features in time-series data. Secondly, we incorporate pruning action into the IAAS framework so that our framework cannot only enlarge the network structures but also shrink them, which increases the search flexibility. Thirdly, we consider a selector actor in IAAS so that we can intelligently and flexibly determine when to widen, deepen, prune the network or keep the network unchanged. Furthermore, we modify the wider and deeper actor networks more intelligently to transform the networks. Finally, we create a net pool component with a heuristic screening algorithm to expand the search space using multiple network structures as candidates in order to improve the probability of obtaining a high-quality network structure.

The experiments have been conducted on two types of electricity data, i.e., load and wind power data. The experimental results demonstrate that the proposed IAAS framework predominately outperforms the ten existing baseline models (or methods) according to the evaluation metrics of forecasting accuracy. Moreover, the proposed IAAS framework has competitively stable performance. Hence, we conclude that the proposed IAAS framework outperforms the existing methods. Moreover, we examine the appropriateness of the selected RL in the application of the IAAS framework from three perspectives by benchmarking with the random search algorithm. The results indicate the selected RL algorithm performs better than the random search algorithm in the forecasting accuracy, average training time per epoch, and average parameter number of each searched model. Furthermore, we make use of the seq2seq setting to compare the IAAS model with two existing transformer models. The results denote that the IAAS framework has a better performance, which means the IAAS framework is more suitable for practical applications in power systems. In addition, we implement an ablation study to showcase the significance of the selector actor and net pool components in the proposed IAAS framework after generating three variants of the framework. The ablation experiment results display that both the selector actor and the net pool component are likely to improve the forecasting model accuracy.

At the end of writing, we further discuss some extensions of the IAAS framework in   Appendix \ref{appx:Further Discussion} of the online supplement. We hope that our research is merely the first step toward more advanced analysis that focuses on electricity forecasting using the NAS technique, which could further improve the secured operation of the power systems. On the other hand, our proposed framework can also be applied to other sequential or time-series cases. Some examples include sales forecasting, travel demand forecasting, and speech recognition data to help improve their model accuracies. In addition, an extension of this research could focus on the theoretical analysis of the heuristic approach proposed in this study to explore why the integration of RL and this heuristic algorithm would improve the performance of the RL agent. And some computational experiments could be conducted to investigate whether this heuristic mechanism has a better performance than the other trajectory truncation methods.

\section*{Data Set and Software}

In accordance with the Journal on Computing's data policy and software policy, we are sharing the data set used and the software developed in \cite{code_sm}. As noted, we do not expose the name of the wind power company as this company requires. 

\section*{Endnotes}

${}^{1}$\textbf{ }Available at https://www.iso-ne.com/isoexpress/web/reports/pricing/-/tree/zone-info{.}

\section*{Declaration of Competing Interest}

All authors declare that they have no conflict of interest in this work.

\section*{Acknowledgment}
Jin Yang, Dr. Guangxin Jiang, and Dr. Ying Chen were supported by the National Natural Science Foundation of China (Grant No. 72293562, 72121001, 72101066, 72131005, 71801148, 72171060). Dr. Ying Chen was also supported by Heilongjiang Natural Science Excellent Youth Fund (YQ2022G004).

\bibliographystyle{informs2014}
\bibliography{main} 

\begin{thebibliography}{69}
\providecommand{\natexlab}[1]{#1}
\providecommand{\url}[1]{\texttt{#1}}
\providecommand{\urlprefix}{URL }

\bibitem[{Agga et~al.(2021)Agga, Abbou, Labbadi, \protect\BIBand{} Houm}]{agga2021short}
Agga A, Abbou A, Labbadi M, Houm Y (2021) Short-term self consumption \text{PV} plant power production forecasts based on hybrid \text{CNN-LSTM, ConvLSTM} models. \emph{Renewable Energy} 177:101--112.

\bibitem[{Ashok et~al.(2017)Ashok, Rhinehart, Beainy, \protect\BIBand{} Kitani}]{ashok2017n2n}
Ashok A, Rhinehart N, Beainy F, Kitani KM (2017) {N2N} learning: Network to network compression via policy gradient reinforcement learning. \emph{arXiv preprint} arXiv:1709.06030.

\bibitem[{Bahdanau et~al.(2015)Bahdanau, Cho, \protect\BIBand{} Bengio}]{Bahdanau2015}
Bahdanau D, Cho K, Bengio Y (2015) Neural machine translation by jointly learning to align and translate. Bengio Y, LeCun Y, eds., \emph{Proceedings of the 3rd International Conference on Learning Representations}.

\bibitem[{Baker et~al.(2017)Baker, Gupta, Naik, \protect\BIBand{} Raskar}]{baker2016designing}
Baker B, Gupta O, Naik N, Raskar R (2017) Designing neural network architectures using reinforcement learning. \emph{Proceedings of the 5th International Conference on Learning Representations}.

\bibitem[{Baydin et~al.(2018)Baydin, Pearlmutter, Radul, \protect\BIBand{} Siskind}]{baydin2018automatic}
Baydin AG, Pearlmutter BA, Radul AA, Siskind JM (2018) Automatic differentiation in machine learning: a survey. \emph{Journal of Marchine Learning Research} 18:1--43.

\bibitem[{Baymurzina et~al.(2022)Baymurzina, Golikov, \protect\BIBand{} Burtsev}]{BAYMURZINA202282}
Baymurzina D, Golikov E, Burtsev M (2022) A review of neural architecture search. \emph{Neurocomputing} 474:82--93.

\bibitem[{Bi et~al.(2022)Bi, Adomavicius, Li, \protect\BIBand{} Qu}]{bi2022improving}
Bi X, Adomavicius G, Li W, Qu A (2022) Improving sales forecasting accuracy: A tensor factorization approach with demand awareness. \emph{INFORMS Journal on Computing} 34(3):1644--1660.

\bibitem[{Billings et~al.(2022)Billings, Smith, Smith, \protect\BIBand{} Powell}]{billings2022industrial}
Billings BW, Smith PJ, Smith ST, Powell KM (2022) Industrial battery operation and utilization in the presence of electrical load uncertainty using bayesian decision theory. \emph{Journal of Energy Storage} 53:105054.

\bibitem[{Cai et~al.(2018)Cai, Chen, Zhang, Yu, \protect\BIBand{} Wang}]{cai2018efficient}
Cai H, Chen T, Zhang W, Yu Y, Wang J (2018) Efficient architecture search by network transformation. \emph{Proceedings of the Thirty-Second {AAAI} Conference on Artificial Intelligence}, volume~32, 2787--2794.

\bibitem[{{\c{C}}evik et~al.(2019){\c{C}}evik, {\c{C}}unka{\c{s}}, \protect\BIBand{} Polat}]{ccevik2019new}
{\c{C}}evik HH, {\c{C}}unka{\c{s}} M, Polat K (2019) A new multistage short-term wind power forecast model using decomposition and artificial intelligence methods. \emph{Physica A: Statistical Mechanics and its Applications} 534:122177.

\bibitem[{Chen et~al.(2021)Chen, Chen, Shang, Zhang, Wen, \protect\BIBand{} Yang}]{chen2021scale}
Chen D, Chen L, Shang Z, Zhang Y, Wen B, Yang C (2021) Scale-aware neural architecture search for multivariate time series forecasting. \emph{arXiv preprint} arXiv:2112.07459.

\bibitem[{Chen et~al.(2018)Chen, Chen, Wang, He, Hu, \protect\BIBand{} He}]{chen2018short}
Chen K, Chen K, Wang Q, He Z, Hu J, He J (2018) Short-term load forecasting with deep residual networks. \emph{IEEE Transactions on Smart Grid} 10(4):3943--3952.

\bibitem[{Chen et~al.(2013)Chen, Qian, Nabney, \protect\BIBand{} Meng}]{chen2013wind}
Chen N, Qian Z, Nabney IT, Meng X (2013) Wind power forecasts using gaussian processes and numerical weather prediction. \emph{IEEE Transactions on Power Systems} 29(2):656--665.

\bibitem[{Chen et~al.(2016)Chen, Goodfellow, \protect\BIBand{} Shlens}]{chen2015net2net}
Chen T, Goodfellow IJ, Shlens J (2016) {Net2Net}: Accelerating learning via knowledge transfer. \emph{Proceedings of the 4th International Conference on Learning Representations}.

\bibitem[{Chen et~al.(2023)Chen, Zhao, Qin, Li, \protect\BIBand{} Zhang}]{chen2023novel}
Chen Y, Zhao J, Qin J, Li H, Zhang Z (2023) A novel pure data-selection framework for day-ahead wind power forecasting. \emph{Fundamental Research} .

\bibitem[{Cheng et~al.(2020)Cheng, Lin, Juan, Wei, \protect\BIBand{} Sun}]{cheng2020instanas}
Cheng AC, Lin CH, Juan DC, Wei W, Sun M (2020) Instanas: Instance-aware neural architecture search. \emph{Proceedings of the AAAI Conference on Artificial Intelligence}, 3577--3584.

\bibitem[{Elsken et~al.(2019)Elsken, Metzen, \protect\BIBand{} Hutter}]{elsken2019neural}
Elsken T, Metzen JH, Hutter F (2019) Neural architecture search: A survey. \emph{The Journal of Machine Learning Research} 20(1):1997--2017.

\bibitem[{Gan et~al.(2020)Gan, Jiang, Lev, \protect\BIBand{} Zhou}]{gan2020balancing}
Gan L, Jiang P, Lev B, Zhou X (2020) Balancing of supply and demand of renewable energy power system: A review and bibliometric analysis. \emph{Sustainable Futures} 2:No.100013.

\bibitem[{Graves \protect\BIBand{} Schmidhuber(2005)}]{graves2005framewise}
Graves A, Schmidhuber J (2005) Framewise phoneme classification with bidirectional lstm and other neural network architectures. \emph{Neural networks} 18(5-6):602--610.

\bibitem[{Greff et~al.(2016)Greff, Srivastava, Koutn{\'\i}k, Steunebrink, \protect\BIBand{} Schmidhuber}]{greff2016lstm}
Greff K, Srivastava RK, Koutn{\'\i}k J, Steunebrink BR, Schmidhuber J (2016) {LSTM}: A search space odyssey. \emph{IEEE transactions on neural networks and learning systems} 28(10):2222--2232.

\bibitem[{Heo et~al.(2021)Heo, Song, Han, \protect\BIBand{} Lee}]{heo2021multi}
Heo J, Song K, Han S, Lee DE (2021) Multi-channel convolutional neural network for integration of meteorological and geographical features in solar power forecasting. \emph{Applied Energy} 295:No.117083.

\bibitem[{Hossain et~al.(2021)Hossain, Gray, Islam, Chakrabortty, \protect\BIBand{} Pota}]{hossain2021forecasting}
Hossain MA, Gray EM, Islam MR, Chakrabortty RK, Pota HR (2021) Forecasting very short-term wind power generation using deep learning, optimization and data decomposition techniques. \emph{24th International Conference on Electrical Machines and Systems (ICEMS)}, 323--327 (IEEE).

\bibitem[{Hu et~al.(2015)Hu, Zhang, Yu, \protect\BIBand{} Xie}]{hu2015short}
Hu Q, Zhang S, Yu M, Xie Z (2015) Short-term wind speed or power forecasting with heteroscedastic support vector regression. \emph{IEEE Transactions on Sustainable Energy} 7(1):241--249.

\bibitem[{Hu \protect\BIBand{} Hong(2022)}]{hu2022shedr}
Hu Y, Hong Y (2022) Shedr: An end-to-end deep neural event detection and recommendation framework for hyperlocal news using social media. \emph{INFORMS Journal on Computing} 34(2):790--806.

\bibitem[{Huang et~al.(2021)Huang, Pan, \protect\BIBand{} Guan}]{huang2021multistage}
Huang J, Pan K, Guan Y (2021) Multistage stochastic power generation scheduling co-optimizing energy and ancillary services. \emph{INFORMS Journal on Computing} 33(1):352--369.

\bibitem[{Islam et~al.(2014)Islam, Baharudin, Raza, \protect\BIBand{} Nallagownden}]{ul2014optimization}
Islam B, Baharudin Z, Raza MQ, Nallagownden P (2014) Optimization of neural network architecture using genetic algorithm for load forecasting. \emph{5th International Conference on Intelligent and Advanced Systems (ICIAS)}, 1--6 (IEEE).

\bibitem[{Jalali et~al.(2021{\natexlab{a}})Jalali, Ahmadian, Khosravi, Shafie-khah, Nahavandi, \protect\BIBand{} Catal{\~a}o}]{jalali2021novel}
Jalali SMJ, Ahmadian S, Khosravi A, Shafie-khah M, Nahavandi S, Catal{\~a}o JP (2021{\natexlab{a}}) A novel evolutionary-based deep convolutional neural network model for intelligent load forecasting. \emph{IEEE Transactions on Industrial Informatics} 17(12):8243--8253.

\bibitem[{Jalali et~al.(2021{\natexlab{b}})Jalali, Os{\'o}rio, Ahmadian, Lotfi, Campos, Shafie-khah, Khosravi, \protect\BIBand{} Catal{\~a}o}]{jalali2021new}
Jalali SMJ, Os{\'o}rio GJ, Ahmadian S, Lotfi M, Campos VM, Shafie-khah M, Khosravi A, Catal{\~a}o JP (2021{\natexlab{b}}) New hybrid deep neural architectural search-based ensemble reinforcement learning strategy for wind power forecasting. \emph{IEEE Transactions on Industry Applications} 58(1):15--27.

\bibitem[{Khodayar et~al.(2017)Khodayar, Kaynak, \protect\BIBand{} Khodayar}]{khodayar2017rough}
Khodayar M, Kaynak O, Khodayar ME (2017) Rough deep neural architecture for short-term wind speed forecasting. \emph{IEEE Transactions on Industrial Informatics} 13(6):2770--2779.

\bibitem[{Lahouar \protect\BIBand{} Slama(2015)}]{lahouar2015day}
Lahouar A, Slama JBH (2015) Day-ahead load forecast using random forest and expert input selection. \emph{Energy Conversion and Management} 103:1040--1051.

\bibitem[{Li et~al.(2022)Li, Huang, Liu, Zhao, Zhou, Zhao, \protect\BIBand{} Yuen}]{li2022day}
Li Y, Huang J, Liu Y, Zhao T, Zhou Y, Zhao Y, Yuen C (2022) Day-ahead risk averse market clearing considering demand response with data-driven load uncertainty representation: A singapore electricity market study. \emph{Energy} 254:123923.

\bibitem[{Liu \protect\BIBand{} Sun(2019)}]{liu2019random}
Liu D, Sun K (2019) Random forest solar power forecast based on classification optimization. \emph{Energy} 187:No.115940.

\bibitem[{Liu et~al.(2022)Liu, Hou, Liu, \protect\BIBand{} Cheng}]{liu2022poolnet+}
Liu JJ, Hou Q, Liu ZA, Cheng MM (2022) Poolnet+: Exploring the potential of pooling for salient object detection. \emph{IEEE Transactions on Pattern Analysis and Machine Intelligence} 45(1):887--904.

\bibitem[{Liu et~al.(2018)Liu, Sun, Zhou, Huang, \protect\BIBand{} Darrell}]{liu2018rethinking}
Liu Z, Sun M, Zhou T, Huang G, Darrell T (2018) Rethinking the value of network pruning. \emph{arXiv preprint} arXiv:1810.05270.

\bibitem[{Mendis et~al.(2021)Mendis, Kang, \protect\BIBand{} Hsiu}]{mendis2021intermittent}
Mendis HR, Kang CK, Hsiu PC (2021) Intermittent-aware neural architecture search. \emph{ACM Transactions on Embedded Computing Systems} 20(5s):1--27.

\bibitem[{Munawer(2018)}]{munawer2018human}
Munawer ME (2018) Human health and environmental impacts of coal combustion and post-combustion wastes. \emph{Journal of Sustainable Mining} 17(2):87--96.

\bibitem[{Munos et~al.(2016)Munos, Stepleton, Harutyunyan, \protect\BIBand{} Bellemare}]{Munos2016_Retrace}
Munos R, Stepleton T, Harutyunyan A, Bellemare M (2016) Safe and efficient off-policy reinforcement learning. Lee D, Sugiyama M, Luxburg U, Guyon I, Garnett R, eds., \emph{Advances in Neural Information Processing Systems}, volume~29 (Curran Associates, Inc.).

\bibitem[{Ng et~al.(2015)Ng, Hausknecht, Vijayanarasimhan, Vinyals, Monga, \protect\BIBand{} Toderici}]{yue2015beyond}
Ng JYH, Hausknecht M, Vijayanarasimhan S, Vinyals O, Monga R, Toderici G (2015) Beyond short snippets: Deep networks for video classification. \emph{Proceedings of the IEEE Conference on Computer Vision and Pattern Recognition}, 4694--4702.

\bibitem[{Nyashina et~al.(2020)Nyashina, Kuznetsov, \protect\BIBand{} Strizhak}]{nyashina2020effects}
Nyashina G, Kuznetsov G, Strizhak P (2020) Effects of plant additives on the concentration of sulfur and nitrogen oxides in the combustion products of coal-water slurries containing petrochemicals. \emph{Environmental Pollution} 258:No.113682.

\bibitem[{Pan et~al.(2021)Pan, Ke, Yang, Liang, Yu, Zhang, \protect\BIBand{} Zheng}]{pan2021autostg}
Pan Z, Ke S, Yang X, Liang Y, Yu Y, Zhang J, Zheng Y (2021) Auto{STG}: Neural architecture search for predictions of spatio-temporal graph. Leskovec J, Grobelnik M, Najork M, Tang J, Zia L, eds., \emph{Proceedings of the Web Conference 2021}, 1846--1855 (ACM).

\bibitem[{Passalis et~al.(2019)Passalis, Tefas, Kanniainen, Gabbouj, \protect\BIBand{} Iosifidis}]{passalis2019deep}
Passalis N, Tefas A, Kanniainen J, Gabbouj M, Iosifidis A (2019) Deep adaptive input normalization for time series forecasting. \emph{IEEE Transactions on Neural Networks and Learning Systems} 31(9):3760--3765.

\bibitem[{Pawlak(1982)}]{pawlak1982rough}
Pawlak Z (1982) Rough sets. \emph{International Journal of Computing and Information Sciences} 11(5):341--356.

\bibitem[{Poiani et~al.(2023)Poiani, Metelli, \protect\BIBand{} Restelli}]{poiani2023truncating}
Poiani R, Metelli AM, Restelli M (2023) Truncating trajectories in monte carlo reinforcement learning. \emph{Proceedings of 40th International Conference of Machine Learning} .

\bibitem[{Poiani et~al.(2024)Poiani, Nobili, Metelli, \protect\BIBand{} Restelli}]{poiani2024truncating}
Poiani R, Nobili N, Metelli AM, Restelli M (2024) Truncating trajectories in monte carlo policy evaluation: an adaptive approach. \emph{Advances in Neural Information Processing Systems} 36.

\bibitem[{Pryor et~al.(2020)Pryor, Barthelmie, Bukovsky, Leung, \protect\BIBand{} Sakaguchi}]{pryor2020climate}
Pryor SC, Barthelmie RJ, Bukovsky MS, Leung LR, Sakaguchi K (2020) Climate change impacts on wind power generation. \emph{Nature Reviews Earth \& Environment} 1(12):627--643.

\bibitem[{Qin et~al.(2021)Qin, Yang, Chen, Ye, \protect\BIBand{} Li}]{qin2021two}
Qin J, Yang J, Chen Y, Ye Q, Li H (2021) Two-stage short-term wind power forecasting algorithm using different feature-learning models. \emph{Fundamental Research} 1(4):472--481.

\bibitem[{Rabier(2005)}]{rabier2005overview}
Rabier F (2005) Overview of global data assimilation developments in numerical weather-prediction centres. \emph{Quarterly Journal of the Royal Meteorological Society} 131(613):3215--3233.

\bibitem[{Real et~al.(2017)Real, Moore, Selle, Saxena, Suematsu, Tan, Le, \protect\BIBand{} Kurakin}]{real2017large}
Real E, Moore S, Selle A, Saxena S, Suematsu YL, Tan J, Le QV, Kurakin A (2017) Large-scale evolution of image classifiers. \emph{Proceedings of the 34th International Conference on Machine Learning}, 2902--2911.

\bibitem[{Ren et~al.(2021)Ren, Xiao, Chang, Huang, Li, Chen, \protect\BIBand{} Wang}]{ren2021comprehensive}
Ren P, Xiao Y, Chang X, Huang PY, Li Z, Chen X, Wang X (2021) A comprehensive survey of neural architecture search: Challenges and solutions. \emph{ACM Computing Surveys} 54(4):1--34.

\bibitem[{Sanh et~al.(2020)Sanh, Wolf, \protect\BIBand{} Rush}]{sanh2020movement}
Sanh V, Wolf T, Rush A (2020) Movement pruning: Adaptive sparsity by fine-tuning. \emph{Advances in Neural Information Processing Systems} 33:20378--20389.

\bibitem[{Shahid et~al.(2021)Shahid, Zameer, \protect\BIBand{} Muneeb}]{shahid2021novel}
Shahid F, Zameer A, Muneeb M (2021) A novel genetic \text{LSTM} model for wind power forecast. \emph{Energy} 223:No.120069.

\bibitem[{Shepero et~al.(2018)Shepero, Van Der~Meer, Munkhammar, \protect\BIBand{} Wid{\'e}n}]{shepero2018residential}
Shepero M, Van Der~Meer D, Munkhammar J, Wid{\'e}n J (2018) Residential probabilistic load forecasting: A method using gaussian process designed for electric load data. \emph{Applied Energy} 218:159--172.

\bibitem[{Simonyan \protect\BIBand{} Zisserman(2015)}]{simonyan2014very}
Simonyan K, Zisserman A (2015) Very deep convolutional networks for large-scale image recognition. Bengio Y, LeCun Y, eds., \emph{Proceedings of the 3rd International Conference on Learning Representations}.

\bibitem[{Solaun \protect\BIBand{} Cerd{\'a}(2019)}]{solaun2019climate}
Solaun K, Cerd{\'a} E (2019) Climate change impacts on renewable energy generation. a review of quantitative projections. \emph{Renewable and Sustainable Energy Reviews} 116:No.109415.

\bibitem[{Srivastava et~al.(2019)Srivastava, Tiwari, \protect\BIBand{} Giri}]{srivastava2019solar}
Srivastava R, Tiwari A, Giri V (2019) Solar radiation forecasting using \text{MARS, CART, M5,} and random forest model: A case study for {I}ndia. \emph{Heliyon} 5(10):No.e02692.

\bibitem[{Sunar \protect\BIBand{} Birge(2019)}]{sunar2019strategic}
Sunar N, Birge JR (2019) Strategic commitment to a production schedule with uncertain supply and demand: Renewable energy in day-ahead electricity markets. \emph{Management Science} 65(2):714--734.

\bibitem[{Tan et~al.(2019)Tan, Chen, Pang, Vasudevan, Sandler, Howard, \protect\BIBand{} Le}]{tan2019mnasnet}
Tan M, Chen B, Pang R, Vasudevan V, Sandler M, Howard A, Le QV (2019) {MnasNet}: Platform-aware neural architecture search for mobile. \emph{Proceedings of the IEEE/CVF Conference on Computer Vision and Pattern Recognition}, 2820--2828.

\bibitem[{Torres et~al.(2019)Torres, Guti{\'e}rrez-Avil{\'e}s, Troncoso, \protect\BIBand{} Mart{\'i}nez-{\'A}lvarez}]{torres2019random}
Torres JF, Guti{\'e}rrez-Avil{\'e}s D, Troncoso A, Mart{\'i}nez-{\'A}lvarez F (2019) Random hyper-parameter search-based deep neural network for power consumption forecasting. Rojas I, Joya G, Catala A, eds., \emph{International Work-Conference on Artificial Neural Networks: Advances in Computational Intelligence}, 259--269.

\bibitem[{Vaswani et~al.(2017)Vaswani, Shazeer, Parmar, Uszkoreit, Jones, Gomez, Kaiser, \protect\BIBand{} Polosukhin}]{vaswani2017attention}
Vaswani A, Shazeer N, Parmar N, Uszkoreit J, Jones L, Gomez AN, Kaiser {\L}, Polosukhin I (2017) Attention is all you need. \emph{Advances in neural information processing systems} 30.

\bibitem[{Wang et~al.(2017)Wang, Bapst, Heess, Mnih, Munos, Kavukcuoglu, \protect\BIBand{} de~Freitas}]{wang2017sample}
Wang Z, Bapst V, Heess N, Mnih V, Munos R, Kavukcuoglu K, de~Freitas N (2017) Sample efficient actor-critic with experience replay. \emph{International Conference on Learning Representations}.

\bibitem[{Weron(2014)}]{weron2014electricity}
Weron R (2014) Electricity price forecasting: A review of the state-of-the-art with a look into the future. \emph{International journal of forecasting} 30(4):1030--1081.

\bibitem[{Xie \protect\BIBand{} Yuille(2017)}]{xie2017genetic}
Xie L, Yuille A (2017) Genetic \text{CNN}. \emph{2017 IEEE International Conference on Computer Vision}, 1379--1388.

\bibitem[{Xiong et~al.(2022)Xiong, Lou, Meng, Wang, Ma, \protect\BIBand{} Wang}]{xiong2022short}
Xiong B, Lou L, Meng X, Wang X, Ma H, Wang Z (2022) Short-term wind power forecasting based on attention mechanism and deep learning. \emph{Electric Power Systems Research} 206:No.107776.

\bibitem[{Yang et~al.(2023)Yang, Jiang, Wang, \protect\BIBand{} Chen}]{code_sm}
Yang J, Jiang G, Wang Y, Chen Y (2023) {An Intelligent End-to-End NAS Framework for Electricity Forecasting}. \urlprefix\url{http://dx.doi.org/10.1287/ijoc.2023.0034.cd}, available for download at https://github.com/INFORMSJoC/2023.0034.

\bibitem[{Zhang et~al.(2020)Zhang, Li, \protect\BIBand{} Zhang}]{zhang2020short}
Zhang Y, Li Y, Zhang G (2020) Short-term wind power forecasting approach based on {Seq2Seq} model using {NWP} data. \emph{Energy} 213:No.118371.

\bibitem[{Zhang et~al.(2022)Zhang, Wei, Zheng, Li, \protect\BIBand{} Zeng}]{zhang2021detecting}
Zhang Z, Wei X, Zheng X, Li Q, Zeng DD (2022) Detecting product adoption intentions via multiview deep learning. \emph{INFORMS Journal on Computing} 34(1):541--556, \urlprefix\url{http://dx.doi.org/10.1287/ijoc.2021.1083}.

\bibitem[{Zhou et~al.(2021{\natexlab{a}})Zhou, Zhang, Peng, Zhang, Li, Xiong, \protect\BIBand{} Zhang}]{zhou2021informer}
Zhou H, Zhang S, Peng J, Zhang S, Li J, Xiong H, Zhang W (2021{\natexlab{a}}) Informer: Beyond efficient transformer for long sequence time-series forecasting. \emph{Proceedings of the AAAI conference on artificial intelligence}, volume~35, 11106--11115.

\bibitem[{Zhou et~al.(2021{\natexlab{b}})Zhou, Li, Chen, Chandrasekaran, \protect\BIBand{} Sanyal}]{zhou2021temporal}
Zhou S, Li X, Chen Y, Chandrasekaran ST, Sanyal A (2021{\natexlab{b}}) Temporal-coded deep spiking neural network with easy training and robust performance. \emph{Proceedings of the 35th AAAI Conference on Artificial Intelligence}, volume~35, 11143--11151.

\bibitem[{Zoph \protect\BIBand{} Le(2017)}]{zoph2016neural}
Zoph B, Le QV (2017) Neural architecture search with reinforcement learning. \emph{Proceedings of the 5th International Conference on Learning Representations}.

\end{thebibliography}

\newpage
\appendix

\setcounter{table}{0}
\renewcommand{\thetable}{A.\arabic{table}}
\setcounter{figure}{0}
\renewcommand{\thefigure}{A.\arabic{figure}}

\section{RNN and LSTM Reformulation}\label{section:RNN and LSTM}
Function-preserving transformation is a key step in architecture design, which ensures that the architecture after transformation has the same performance as its predecessor. To realize this in RNN, we first propose a general operation scheme in this section to unify the feature learning operation in an ordinary neural network (e.g., FCN). And then, we reformulate the feature learning operation of RNN and LSTM based on this unified scheme.

\subsection{Feature learning operation}\label{subsection:feature-learning}

RNN is able to pass information through time. In addition to receiving information from the input features at the current time step, the output features from the previous time steps are also incorporated to generate the RNN outputs. However, the function-preserving transformation method from \cite{chen2015net2net} only considers the temporal independent situation, which cannot change the hidden state at the current time step when the outputs from the previous time steps are changed. Therefore, the EAS framework \citep{cai2018efficient} built on the Net2Net framework \citep{chen2015net2net} that targets image data, cannot be intuitively used in time-series data. Regarding the excellence of RNN in preprocessing time-series data and extracting features from them, it is necessary to develop the RNN function-preserving transformation so that the EAS framework can be adapted to such data and improve the search quality of network structures for power systems. To prepare for deriving the RNN function-preserving transformation, we first propose a general two-phase formulation of the feature learning process in a deep learning model.

Suppose that for layer $i$ in an FCN, there are $p_i$ input features ${\boldsymbol{x}}^{(i)}=(x^{(i)}_1,\ x^{(i)}_2,\dots,x^{(i)}_{p_i})^{\top} \in \mathbb{R}^{{p_i}\times 1}$ and $p_{i+1}$ output features ${\boldsymbol{x}}^{(i+1)}=(x^{(i+1)}_1,\ x^{(i+1)}_2,\dots,x^{(i+1)}_{p_{i+1}})^{\top}  \in \mathbb{R}^{p_{i+1}\times 1}$. ${\boldsymbol{x}}^{(i+1)}$ will be fed into layer $i+1$ as the input. The weight matrix for layer $i$ is ${\boldsymbol{W}}^{(i)}\in {\mathbb{R}}^{p_i \times p_{i+1}}$ and its transpose is represented by ${{\boldsymbol{W}}^{(i)}}^{\top}\in {\mathbb{R}}^{p_{i+1} \times p_i}$. The forward calculation of FCN can be represented by Eq. (\ref{eq: Forward-FCN}), where $f_a$ is the activation function. Note that we ignore the bias term in this equation for notation convenience, and the bias can be assumed to be integrated into the weight matrix.
\begin{equation}
    \boldsymbol{x}^{(i+1)} = f_a({\boldsymbol{W}^{(i)}}^{\top} \boldsymbol{x}^{(i)})
    \label{eq: Forward-FCN}
\end{equation}
\quad The Phase-1 learning is specified as a function $f_1(x^{(i)}_j,W^{(i)}_{j,k})= x^{(i)}_j\cdot W^{(i)}_{j,k}$, for $j=1,2,\dots,p_i$ and $k=1,2,\dots,p_{i+1}$, which is a scalar multiplication. Subsequently, the Phase-1 learning results ${\boldsymbol{o}}_k$ for layer $i$ are obtained as follows:
\begin{equation} {\boldsymbol{o}}^{(i)}_k=\left(f_1(x^{(i)}_1,W^{(i)}_{1,k}),f_1(x^{(i)}_2,W^{(i)}_{2,k}),\dots,f_1(x^{(i)}_{p_i},W^{(i)}_{p_i,k})\right), k=1,\dots ,p_{i+1}
    \label{eq:feature-extrat-O}
\end{equation}

The Phase-2 learning is denoted by an aggregating function that summarizes information from the Phase-1 learning process, $f_2({\boldsymbol{o}}^{(i)}_k)=\sum^{p_i}_{j=1}{f_1(x^{(i)}_j, W^{(i)}_{j,k})}$, for $k=1,\dots,p_{i+1}$. Then, we generate the output features ${\boldsymbol{x}}^{(i+1)}$, which are given by
\begin{equation}
    {\boldsymbol{x}}^{(i+1)}=f_a\left(f_2({\boldsymbol{o}}^{(i)}_1),f_2({\boldsymbol{o}}^{(i)}_2),\dots ,f_2({\boldsymbol{o}}^{(i)}_{p_{i+1}})\right)
    \label{eq:generate-outfeature}
\end{equation}

\begin{figure}[H]

    \includegraphics[scale=0.5]{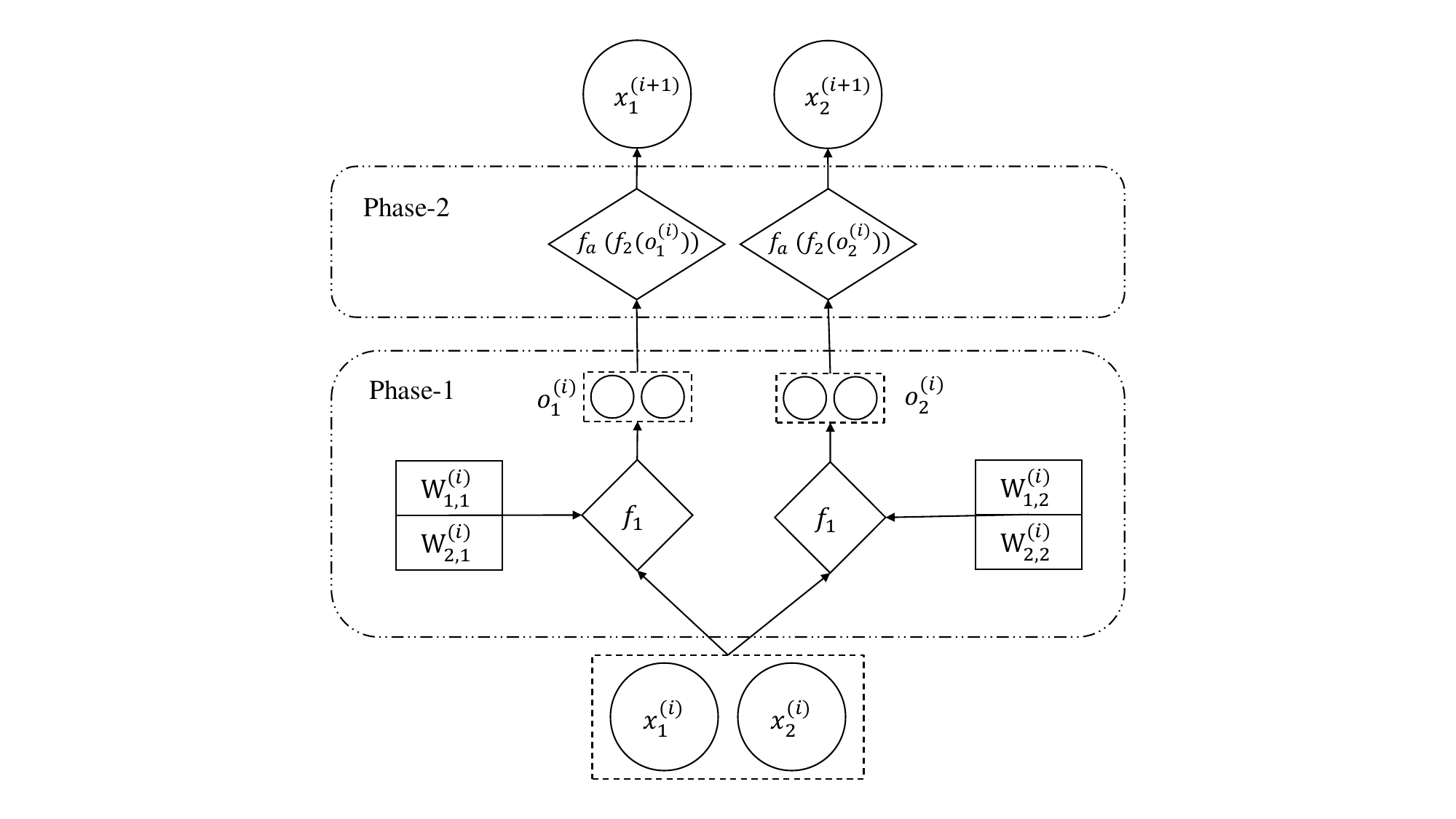}
    \caption{An example of two-phase feature learning operation in FCN\label{fig:spatial-feature-learning}}
\end{figure}

Fig. \ref{fig:spatial-feature-learning} depicts a simple FCN example of such a feature learning process. As shown in the figure, layer $i$ has two input and two output features. Hence, the meta operation $f_1$ is utilized to do the mathematical operation (scalar multiplication here) with the input features (i.e., Phase-1 learning), and the aggregation operation $f_2$ is employed to accumulate the learned knowledge to produce new features (i.e., Phase-2 learning). Finally, we acquire input features of the next layer through an activation function $f_a$.

As noted, such a two-phase description of the feature learning process can be easily extended to CNN. Briefly, for the Phase-1 learning of a CNN layer, a convolution operation is performed on each input channel with a kernel, generating the same number of intermediate results as that of the input features. For the Phase-2 learning, all the intermediate results are summed up to generate one new input feature for the next layer, and this process is repeated multiple times to obtain all the output features. For instance, there are $p_i$ channels of inputs and $p_{i+1}$ channels of outputs for layer $i$. Hence, we employ $p_i$ kernels to get the intermediate results from inputs ${\boldsymbol{x}}^{(i)}$. We then have intermediate results:
\begin{equation}
    o\boldsymbol{K}^{(i)}_{j,k} =f_1(\boldsymbol{x}^{(i)}_j, \boldsymbol{k}^{(i)}_{j,k}),j=1,\dots,p_i,k=1,\dots,p_{i+1}, 
    \label{eq: intermediate-CNN}
\end{equation}
where $f_1$ is a convolution function, $\boldsymbol{x}^{(i)}_j$ is the $j^{th}$ channel of input tensor $\boldsymbol{x}^{(i)}$ and $\boldsymbol{k}^{(i)}_{j,k}$ is a convolution kernel. Thus, we have the Phase-1 learning results $\boldsymbol{o}^{(i)}_k=(o\boldsymbol{K}^{(i)}_{1,k},\dots,o\boldsymbol{K}^{(i)}_{p_i,k})$.
After that, all the intermediate results are summed up to form one output feature in Phase-2 learning:
\begin{equation}
    \boldsymbol{x}^{(i+1)}_k = f_2(\boldsymbol{o}^{(i)}_k)=\sum_{j=1}^{p_i}o\boldsymbol{K}^{(i)}_{j,k},k=1,\dots,p_{i+1}. 
    \label{eq: intermediate-CNN2}
\end{equation}
This process is repeated $p_{i+1}$ times to achieve $p_{i+1}$ output features. In total, we use $p_i \times p_{i+1}$ kernels to extract the features from the inputs ${\boldsymbol{x}}^{(i)}$. With these two-phase feature learning process described above, we can obtain the outputs ${\boldsymbol{x}}^{(i+1)}$ as new input features for layer $i+1$.
\subsection{RNN reformulation}\label{subsection:rnn-reformulation}
\vspace{0.2cm}
Following the settings of the feature learning process in spatial networks (e.g., FCN), we assume that at time $t$, an RNN layer $i$ has $p_i$ input features ${\boldsymbol{x}}^{(i)}_t=(x^{(i)}_{1;t},\ x^{(i)}_{2;t},\dots,x^{(i)}_{p_i;t}) ^{\top} \in \mathbb{R}^{{p_i}\times 1}$ and $p_{i+1}$ output features ${\boldsymbol{x}}^{(i+1)}_t=(x^{(i+1)}_{1;t},\ x^{(i+1)}_{2;t},\dots,x^{(i+1)}_{{p_{i+1}};t}) ^{\top} \in \mathbb{R}^{p_{i+1}\times 1}$. As seen, we have used a semicolon in subscript notation to distinguish the time information from other information. We assume the input weight matrix as ${\boldsymbol{W}}^{(i)}\in {\mathbb{R}}^{p_i \times p_{i+1}}$ and the output hidden state weight matrix as ${\boldsymbol{H}}^{(i)}\in {\mathbb{R}}^{p_{i+1}\times p_{i+1}}$. Then, the ordinary RNN feature learning process is defined as:
\begin{equation}
    {\boldsymbol{x}}^{(i+1)}_t=f_a({{\boldsymbol{W}}^{(i)}}^{\top}  {\boldsymbol{x}}^{(i)}_t+{{\boldsymbol{H}}^{(i)}}^{\top}  {\boldsymbol{x}}^{(i+1)}_{t-1}) \label{eq:rnn_process}.
\end{equation}

In the following, we present how to reformulate the RNN feature learning process with the proposed two-phase operation. The meta operation $f_1$ for RNN is a scalar multiplication and can be used to extract the information from ${\boldsymbol{x}}^{(i)}_t$ and ${\boldsymbol{x}}^{(i+1)}_{t-1}$. Therefore, we specify the Phase-1 learning results of layer $i$ at time $t$ related to ${\boldsymbol{W}}^{(i)}$ and ${\boldsymbol{H}}^{(i)}\ $ as follows:
\begin{equation}
    {oW}^{(i)}_{j,k;t}=f_1(x^{(i)}_{j;t},W^{(i)}_{j,k}),j=1,2,\dots ,p_i,k=1,\dots,p_{i+1} 
    \label{eq:extraction-W},
\end{equation}
\begin{equation}
    {oH}^{(i)}_{l,k;t}=f_1(x^{(i+1)}_{l;t-1},H^{(i)}_{l,k}),l=1,2,\dots ,p_{i+1},k=1,\dots,p_{i+1}
    \label{eq:extraction-H}.
\end{equation}

After that, we obtain the Phase-1 learning results ${\boldsymbol{o}}^{(i)}_{k;t}$ of layer $i\ $at time $t$:
\begin{equation}
    {\boldsymbol{o}}^{(i)}_{k;t}=({oW}^{(i)}_{1,k;t},\dots{oW}^{(i)}_{p_i,k;t},{oH}^{(i)}_{1,k;t},\dots ,{oH}^{(i)}_{p_{i+1},k;t}),k=1,\dots,p_{i+1}.
    \label{eq:extraction-result}
\end{equation}

In Phase-2 learning, we sum up the information from the Phase-1 learning results with the use of the aggregating function $f_2({\boldsymbol{o}}^{(i)}_{k;t})=\sum^{p_i}_{j=1}{oW}^{(i)}_{j,k;t}+\sum^{p_{i+1}}_{l=1}{oH}^{(i)}_{l,k;t}$, for $k=1,\dots,p_{i+1}$. Hence, we generate the new outputs ${\boldsymbol{x}}^{(i+1)}_t$ at time $t$ through the following equation:
\begin{equation}
    {\boldsymbol{x}}^{(i+1)}_t=(f_a(f_2({\boldsymbol{o}}^{(i)}_{1;t})),f_a(f_2({\boldsymbol{o}}^{(i)}_{2;t})),\dots ,f_a(f_2({\boldsymbol{o}}^{(i)}_{{p_{i+1}};t}))).
    \label{eq:new-output-RNN}
\end{equation}

\begin{figure}
    \centering
    \includegraphics[scale=0.5]{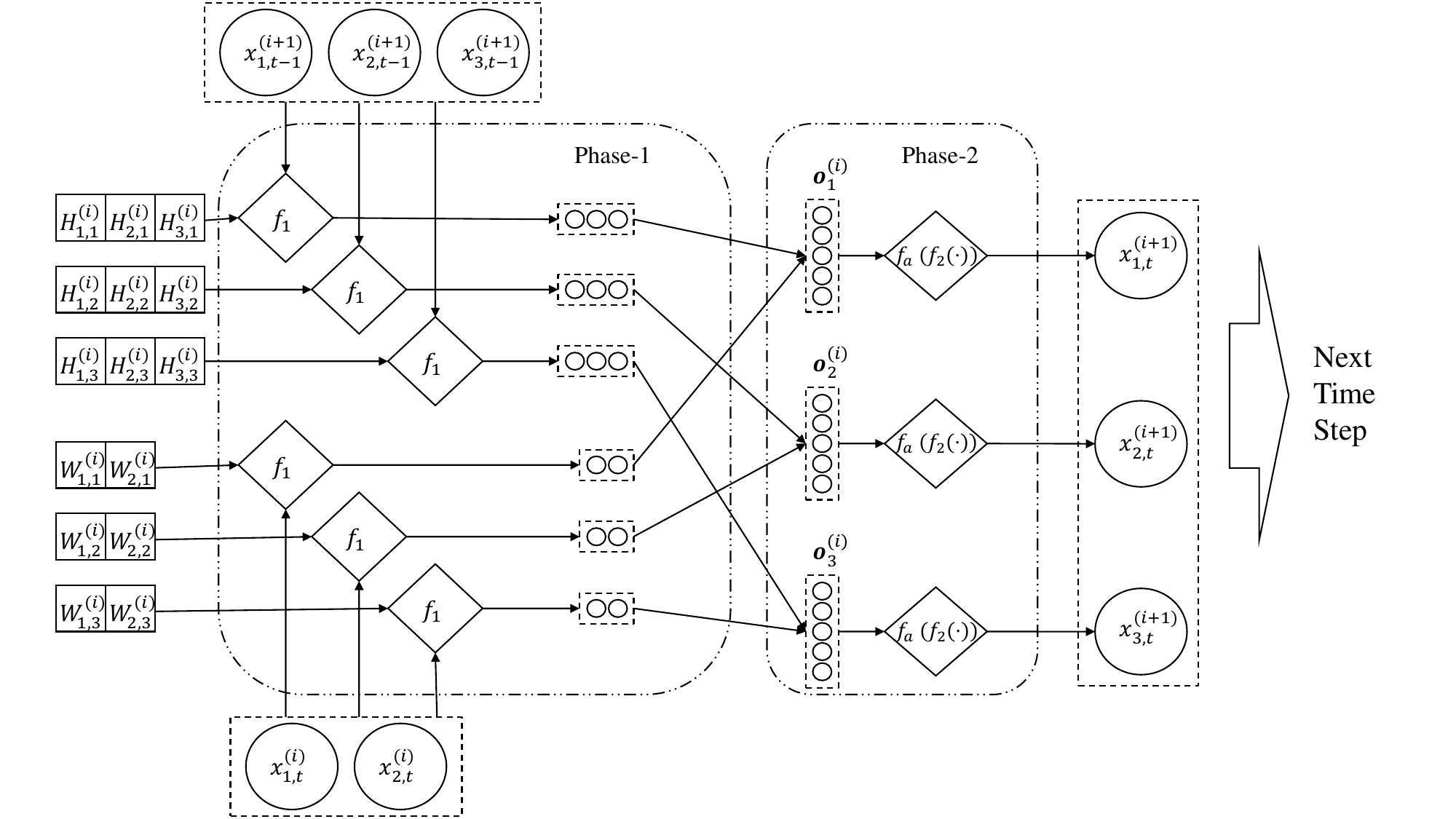}
    \caption{An example of a two-phase feature learning operation in RNN \label{fig:rnn-feature-learning}}
    
\end{figure}

An example of such a two-phase feature learning in RNN is depicted in Fig. \ref{fig:rnn-feature-learning}. As shown, we have two input features at time $t$ and three output features at time $t-1$. Both input and output features perform the element-wise meta operation $f_1$ with its corresponding weight matrices, and then we acquire the intermediate results. Subsequently, we utilize the aggregating function $f_2$ to generate new output features of RNN at time $t$. In the following, we present how to reformulate the LSTM with the two-phase feature learning process.

\subsection{LSTM reformulation}\label{subsection:lstm-reformulation}
Different from the general RNN architecture, LSTM is able to explore the long-term dependencies in the data by employing a series of memory cells, leading to a much more complicated function-preserving transformation than that of RNN in the Net2Net framework \citep{chen2015net2net}. Similar to RNN,  we also assume that at time $t$, an LSTM layer $i$ has $p_i$ input features ${\boldsymbol{x}}^{(i)}_t=(x^{(i)}_{1;t},\ x^{(i)}_{2;t},\dots ,x^{(i)}_{p_i;t}) ^{\top} \in \mathbb{R}^{{p_i}\times 1}$ and $p_{i+1}$ output features ${\boldsymbol{x}}^{(i+1)}_t=(x^{(i+1)}_{1;t},\ x^{(i+1)}_{2;t},\dots ,x^{(i+1)}_{{p_{i+1}};t}) ^{\top} \in \mathbb{R}^{{p_{i+1}}\times 1}$. The calculation of the LSTM process is shown in Eq. (\ref{eq: compact lstm}).
\begin{equation}
    \label{eq: compact lstm}
    \begin{aligned}
    \boldsymbol{f}^{(i)}_t &= \sigma_g({\boldsymbol{W}^{(i)}_f}^{\top}\boldsymbol{x}^{(i)}_t+{\boldsymbol{H}^{(i)}_f}^{\top}\boldsymbol{x}^{(i+1)}_{t-1})\\
    \boldsymbol{i}^{(i)}_t &= \sigma_g({\boldsymbol{W}^{(i)}_i}^{\top}\boldsymbol{x}^{(i)}_t+{\boldsymbol{H}^{(i)}_i}^{\top}\boldsymbol{x}^{(i+1)}_{t-1})\\
    \boldsymbol{o}^{(i)}_t &= \sigma_g({\boldsymbol{W}^{(i)}_o}^{\top}\boldsymbol{x}^{(i)}_t+{\boldsymbol{H}^{(i)}_o}^{\top}\boldsymbol{x}^{(i+1)}_{t-1})\\
    \Tilde{\boldsymbol{c}_t}^{(i)} &= \sigma_c({\boldsymbol{W}^{(i)}_c}^{\top}\boldsymbol{x}^{(i)}_t+{\boldsymbol{H}^{(i)}_c}^{\top}\boldsymbol{x}^{(i+1)}_{t-1})\\
    \boldsymbol{c}^{(i)}_t &= \boldsymbol{f}^{(i)}_t \circ \boldsymbol{c}^{(i)}_{t-1} + \boldsymbol{i}^{(i)}_t \circ \Tilde{\boldsymbol{c}_t}^{(i)}\\
    \boldsymbol{x}^{(i+1)}_t &= \boldsymbol{o}^{(i)}_t \circ \sigma_c(\boldsymbol{c}^{(i)}_t)
    \end{aligned}
\end{equation}
In the LSTM formulations,  $\boldsymbol{f}^{(i)}_t, \boldsymbol{i}^{(i)}_t, \boldsymbol{o}^{(i)}_t$ and $\boldsymbol{c}^{(i)}_t$ denote the forget gate, input gate, output gate, and memory cell vectors at time $t$, respectively; $\Tilde{\boldsymbol{c}_t}^{(i)}$ denotes the cell input activation vector; $\boldsymbol{x}^{(i+1)}_t$ also denotes the hidden state for layer $i$; operator $\circ$ represents the Hadamard product, which calculates the element-wise products of two same dimensional vectors or matrices; $\sigma_g$ and $\sigma_c$ represent the Sigmoid activation function and the hyperbolic tangent function, respectively; $\boldsymbol{W}^{(i)}_f\in \mathbb{R}^{p_i \times p_{i+1}}, \boldsymbol{W}^{(i)}_i\in \mathbb{R}^{p_i \times p_{i+1}}$, $\boldsymbol{W}^{(i)}_o\in \mathbb{R}^{p_i \times p_{i+1}}$, $\boldsymbol{W}^{(i)}_c\in \mathbb{R}^{p_i \times p_{i+1}}$, $\boldsymbol{H}^{(i)}_f\in \mathbb{R}^{p_{i+1}\times p_{i+1}}$, $\boldsymbol{H}^{(i)}_i\in \mathbb{R}^{p_{i+1}\times p_{i+1}}$, $\boldsymbol{H}^{(i)}_o\in \mathbb{R}^{p_{i+1}\times p_{i+1}}$ and $\boldsymbol{H}^{(i)}_c\in \mathbb{R}^{p_{i+1}\times p_{i+1}}$ are weight parameters. 

As introduced in \cite{chen2015net2net}, the identity mapping function is paramount to the function-preserving transformation of deepening the networks. However, the LSTM calculation process in Eq. (\ref{eq: compact lstm}) contains a lot of activation functions to calculate $\boldsymbol{x}^{(i+1)}_t$, which directly prevent the application of identity mapping. Hence, we restructure the forward propagation of LSTM in Eq. (\ref{eq: compact lstm}) by altering the formulations of both $\boldsymbol{o}^{(i)}_t$ and $\boldsymbol{x}^{(i+1)}_t$ as

\begin{equation}
    \label{eq: custom lstm}
    \begin{aligned}
   \boldsymbol{o}^{(i)}_t  &=  \min\{\max\{{\boldsymbol{W}^{(i)}_o}^{\top}\boldsymbol{x}^{(i)}_t+{\boldsymbol{H}^{(i)}_o}^{\top} \boldsymbol{x}^{(i+1)}_{t-1},1\},0\},\\
     \boldsymbol{x}^{(i+1)}_t&=\min\{\max\{\boldsymbol{o}^{(i)}_t \circ \sigma_c(\boldsymbol{c}^{(i)}_t) + \boldsymbol{o}^{(i)}_t,1\},-1\},
    \end{aligned}
\end{equation}
where $\min{(\boldsymbol{a},a)}$ ($\max{(\boldsymbol{a},a)}$) of vector $\boldsymbol{a}$ and constant $a$ gives the element-wise minimum (maximum) value.
Note that the output ranges of $o^{(i)}_t$ and $x^{(i+1)}_t$ in Eq. (\ref{eq: custom lstm}) are the same as the original ranges [0, 1] and [-1, 1] by the truncate operation, respectively. Moreover, we add an additional $\boldsymbol{o}^{(i)}$ in the second term for the following reasons. The original output $\boldsymbol{o}^{(i)}_t \circ \sigma_c(\boldsymbol{c}^{(i)}_t)$ always modifies the input $\boldsymbol{o}^{(i)}$ because the multiplication term $\sigma_c(\boldsymbol{c}^{(i)}_t)$ cannot be set to 1. To preserve the capability of passing the input unchanged to the next layer, we add a new term $\boldsymbol{o}^{(i)}$. In this case, when the multiplication factor $\sigma_c(\boldsymbol{c}^{(i)}_t)$ is set to 0, the modified LSTM is an identity mapping to preserve and pass the input information, which is especially important in building deeper models. For LSTM, the meta operation $f_1$ is also a scalar multiplication and can be used to extract the information from $\boldsymbol{x}^{(i)}_t$ and $\boldsymbol{x}^{(i+1)}_{t-1}$ (i.e., $f_1(a,b)=a\cdot b$). We specify the intermediate results as follows:
\begin{equation}
    \begin{aligned}
{oW}^{(i)}_{f;j,k;t}&=f_1(x^{(i)}_{j;t},W^{(i)}_{f;j,k}),j=1,2,\dots ,p_i,k=1,\dots,p_{i+1}; \\
{oW}^{(i)}_{i;j,k;t}&=f_1(x^{(i)}_{j;t},W^{(i)}_{i;j,k}),j=1,2,\dots ,p_i,k=1,\dots,p_{i+1}; \\
{oW}^{(i)}_{o;j,k;t}&=f_1(x^{(i)}_{j;t},W^{(i)}_{o;j,k}),j=1,2,\dots ,p_i,k=1,\dots,p_{i+1}; \\
{oW}^{(i)}_{c;j,k;t}&=f_1(x^{(i)}_{j;t},W^{(i)}_{c;j,k}),j=1,2,\dots ,p_i,k=1,\dots,p_{i+1}; \\
{oH}^{(i)}_{f;l,k;t}&=f_1(x^{(i+1)}_{l;t-1},H^{(i)}_{f;l,k}),l=1,2,\dots ,p_{i+1},k=1,\dots,p_{i+1};\\
{oH}^{(i)}_{i;l,k;t}&=f_1(x^{(i+1)}_{l;t-1},H^{(i)}_{i;l,k}),l=1,2,\dots ,p_{i+1},k=1,\dots,p_{i+1};\\
{oH}^{(i)}_{c;l,k;t}&=f_1(x^{(i+1)}_{l;t-1},H^{(i)}_{c;l,k}),l=1,2,\dots ,p_{i+1},k=1,\dots,p_{i+1};\\
{oH}^{(i)}_{o;l,k;t}&=f_1(x^{(i+1)}_{l;t-1},H^{(i)}_{o;l,k}),l=1,2,\dots ,p_{i+1},k=1,\dots,p_{i+1}.
    \end{aligned}
    \label{eq:extraction-LSTM}
\end{equation}

After that, we obtain the Phase-1 learning results ${\boldsymbol{o}}^{(i)}_{k;t}$ of LSTM layer $i\ $at time $t$:

\begin{equation}
\begin{aligned}
     {\boldsymbol{o}}^{(i)}_{k;t}=(
    &{oW}^{(i)}_{f;1,k;t},\dots{oW}^{(i)}_{f;p_i,k;t},
    {oW}^{(i)}_{i;1,k;t},\dots{oW}^{(i)}_{i;p_i,k;t},\\
    &{oW}^{(i)}_{c;1,k;t},\dots{oW}^{(i)}_{c;p_i,k;t},
    {oW}^{(i)}_{o;1,k;t},\dots{oW}^{(i)}_{o;p_i,k;t},\\
    &{oH}^{(i)}_{f;1,k;t},\dots ,{oH}^{(i)}_{f;p_{i+1}k;t},
    {oH}^{(i)}_{i;1,k;t},\dots ,{oH}^{(i)}_{i;p_{i+1}k;t},\\
    &{oH}^{(i)}_{c;1,k;t},\dots ,{oH}^{(i)}_{c;p_{i+1}k;t},
    {oH}^{(i)}_{o;1,k;t},\dots ,{oH}^{(i)}_{o;p_{i+1}k;t}
    ),k=1,\dots,p_{i+1}.
\end{aligned}
    \label{eq:extraction-result-LSTM}
\end{equation}

According to LSTM formulation in Eq. (\ref{eq: compact lstm}) and (\ref{eq: custom lstm}), we can aggregate all these intermediate results with the aggregating function $f_{2;c}$ for the cell state and $f_{2;x}$ for the hidden state as

\begin{equation}
\begin{aligned}
    c^{(i)} _{k;t}& = f_{2;c}({\boldsymbol{o}}^{(i)}_{k;t})=
    \sigma_g(\sum^{p_i}_{j=1}{oW}^{(i)}_{f;j,k;t}+\sum^{p_{i+1}}_{l=1}{oH}^{(i)}_{f;l,k;t}) \cdot c^{(i)} _{k;t-1}\\ + &\sigma_g(\sum^{p_i}_{j=1}{oW}^{(i)}_{i;j,k;t}+\sum^{p_{i+1}}_{l=1}{oH}^{(i)}_{i;l,k;t}) \cdot \sigma_c(\sum^{p_i}_{j=1}{oW}^{(i)}_{c;j,k;t}+\sum^{p_{i+1}}_{l=1}{oH}^{(i)}_{c;l,k;t})
    ,k=1,\dots,p_{i+1},
\end{aligned}\label{eq:aggretation-cell-state-LSTM}
\end{equation}
\begin{equation}
\begin{aligned}
    x^{(i+1)}_{k;t} =f_{2;x}({\boldsymbol{o}}^{(i)}_{k;t})=(\sum^{p_i}_{j=1}{oW}^{(i)}_{o;j,k;t}+\sum^{p_{i+1}}_{l=1}{oH}^{(i)}_{o;l,k;t}) \cdot  (f_{2,c}({\boldsymbol{o}}^{(i)}_{k;t})+1)
    ,k=1,\dots,p_{i+1}.
\end{aligned}\label{eq:aggretation-hidden-state-LSTM}
\end{equation}
With these two aggregating functions in Phase-2 learning, we generate the new output $\boldsymbol{x}^{(i+1)}_t$ and cell state $\boldsymbol{c}^{(i)}_t$ at time $t$ through the following equations:
\begin{equation}
    \boldsymbol{c}^{(i)}_t = (
    f_{2;c}({\boldsymbol{o}}^{(i)}_{1;t}),
    f_{2;c}({\boldsymbol{o}}^{(i)}_{2;t}),
    \dots,
    f_{2;c}({\boldsymbol{o}}^{(i)}_{{p_{i+1}};t})
    ),
    \label{eq:new-output-cell-state-LSTM}
\end{equation}
\begin{equation}
    \boldsymbol{x}^{(i+1)}_t = (
    f_{2;x}({\boldsymbol{o}}^{(i)}_{1;t}),
    f_{2;x}({\boldsymbol{o}}^{(i)}_{2;t}),
    \dots,
    f_{2;x}({\boldsymbol{o}}^{(i)}_{{p_{i+1}};t})
    ).
    \label{eq:new-output-hidden-state-LSTM}
\end{equation}

\newpage
\section{Proof of Propositions}\label{sec:appx-proof-of-propositions}

\begin{proof}
First, we consider the feature extraction for the \textit{wider} layer $i'$. We apply the new student network parameters of ${\boldsymbol{W}}^{(i')}$\textit{ }and\textit{ }${\boldsymbol{H}}^{(i')}$ in Proposition \ref{proposition: RNN-wider} to Eq. \eqref{eq:extraction-W} and \eqref{eq:extraction-H}, and the extracted features at time $t$ are given by:
\begin{equation}
    {oW}^{(i')}_{j,k;t}={oW}^{(i)}_{j,g(k);t},j=1,2,\dots ,p_i,k=1,\dots,{p'_{i+1}},
    \label{proof-p1:oW}
\end{equation}
\begin{equation}
    {oH}^{(i')}_{l,k;t}=f_w(l){oH}^{(i)}_{g(l),g(k);t},l=1,2,\dots ,{p'_{i+1}},k=1,\dots,{p'_{i+1}}.
    \label{proof-p1:oH}
\end{equation}
Following Eq. \eqref{eq:extraction-result}, the feature extraction results of layer $i'\ $at time $t$, ${\boldsymbol{o}}^{(i')}_{k;t}$, are given by:
\begin{equation}
    {\boldsymbol{o}}^{(i')}_{k;t}=({oW}^{(i')}_{1,k;t},\dots{oW}^{(i')}_{p_i,k;t},{oH}^{(i')}_{1,k;t},\dots         ,{oH}^{(i')}_{{p'_{i+1}},k;t}),k=1,\dots,{p'_{i+1}}.
    \label{proof-p1:oi}
\end{equation}
Then, by Eq. \eqref{eq:new-output-RNN} and the aggregating function $f_2({\boldsymbol{o}}^{(i')}_{k;t})=\sum^{p_i}_{j=1}{{oW}^{(i')}_{j,g(k);t}}+\sum^{p'_{i+1}}_{l=1}{{oH}^{(i')}_{l,k;t}}$, $k=1,\dots,{p'_{i+1}}$, the new output $x^{(i'+1)}_{k;t}$ are given by:
\begin{equation}
\begin{aligned}
x^{(i'+1)}_{k;t}&=f_a(f_2({\boldsymbol{o}}^{(i')}_{k;t})) \\ 
&=f_a(\sum^{p_i}_{j=1}{{oW}^{(i')}_{j,g(k);t}}+\sum^{p'_{i+1}}_{l=1}{{oH}^{(i')}_{l,k;t}}) \\ 
&=f_a(\sum^{p_i}_{j=1}{{oW}^{(i)}_{j,g(k);t}}+\sum^{p'_{i+1}}_{l=1}{f_w(l){oH}^{(i)}_{g(l),g(k);t}}) \\
&=f_a(\sum^{p_i}_{j=1}{{oW}^{(i)}_{j,g(k);t}}+\sum^{p_{i+1}}_{l=1}{{oH}^{(i)}_{l,g(k);t}}) \\
&=x^{(i+1)}_{g(k);t}\ \ {,\ \ }k=1,\dots,{p'_{i+1}},
\end{aligned}
\label{proof-p1:wider-new-output-i}
\end{equation}
where the third equality holds by Eq. \eqref{proof-p1:oW} and \eqref{proof-p1:oH}, the fourth equality holds by Eq. \eqref{eq:replication-factor}, and the last equality holds by the definition of the aggregating function and Eq. \eqref{eq:new-output-RNN}. Hence, the output feature vector ${\boldsymbol{x}}^{(i'+1)}_{\ t}$ of layer $i'$ also serves as the input features of layer $i'+1$. Following Eq. \eqref{eq:new-output-RNN}, and ${\boldsymbol{x}}^{(i'+1)}_{\ t}$ can be identified as
\begin{equation}
  {\boldsymbol{x}}^{(i'+1)}_t=(x^{(i+1)}_{g(1);t},x^{(i+1)}_{g(2);t},\dots ,x^{(i+1)}_{g({p'_{i+1}});t}).  
  \label{proof-p1:input-of-i+1}
\end{equation}
Note that ${\boldsymbol{x}}^{(i'+1)}_t$ is essentially a random mapping defined by $g$. Similar to the RNN reformulation in wider layer $i'$ as shown above, we leverage the new student network parameters of ${\boldsymbol{W}}^{(i'+1)}$\textit{ }and\textit{ }${\boldsymbol{H}}^{(i'+1)}$ in Proposition \ref{proposition: RNN-wider} to Eq. \eqref{eq:extraction-W} and \eqref{eq:extraction-H}, and acquire the extracted features for the next layer $i'+1$ as follows:
\begin{equation}
    {oW}^{(i'+1)}_{k,h;t}=f_w(k){oW}^{(i+1)}_{g(k),h;t},k=1,2,\dots ,{p'_{i+1}},h=1,\dots,{p_{i+2}},
    \label{proof-p1:oWi+1}
\end{equation}
\begin{equation}
    {oH}^{(i'+1)}_{r,h;t}={oH}^{(i+1)}_{r,h;t},r=1,2,\dots ,{p_{i+2}},h=1,\dots,{p_{i+2}}.
    \label{proof-p1:oHi+1}
\end{equation}
As observed in Eq. \eqref{proof-p1:oHi+1}, the feature extraction for outputs of layer $i'+1$ is the same as that of layer $i+1$ since the dimension is unchanged. Following Eq. \eqref{eq:extraction-result}, the feature extraction results of layer $i'+1\ $at time $t$, ${\boldsymbol{o}}^{(i'+1)}_{h;t}$, are given by
\begin{equation}
    {\boldsymbol{o}}^{(i'+1)}_{h;t}=({oW}^{(i'+1)}_{1,h;t},\dots{oW}^{(i'+1)}_{{p'_{i+1}},h;t},{oH}^{(i'+1)}_{1,h;t},\dots ,{oH}^{(i'+1)}_{{p_{i+2}},h;t}),h=1,\dots,{p_{i+2}}.
    \label{proof-p1:extraction-result-i+1}
\end{equation}
Next, we sum up the extracted features with the aggregating function $f_2({\boldsymbol{o}}^{(i'+1)}_{h;t})=\sum^{p'_{i+1}}_{k=1}{{oW}^{(i'+1)}_{k,h;t}}+\sum^{p_{i+2}}_{r=1}{{oH}^{(i'+1)}_{r,h;t}}$, for $h=1,\dots,{p_{i+2}}$, to generate the new output $x^{(i'+2)}_{h;t}$, $h=1,\dots,{p_{i+2}}.$ Similar to the analysis of Eq. \eqref{proof-p1:wider-new-output-i}, by incorporating Eq. (9-10) and Eq. (16-18), we have the following output:
\begin{equation}
    \begin{aligned}
    x^{(i'+2)}_{h;t}
    &=\ f_{{a}}(f_2({\boldsymbol{o}}^{(i'+1)}_{h;t})) \\ 
    &=f_a(\sum^{p'_{i+1}}_{k=1}{{oW}^{(i'+1)}_{k,h;t}}+\sum^{p_{i+2}}_{r=1}{{oH}^{(i'+1)}_{r,h;t}}) \\ 
    &=f_a(\sum^{p'_{i+1}}_{k=1}{f_w(k){oW}^{(i+1)}_{g(k),h;t}}+\sum^{p_{i+2}}_{r=1}{{oH}^{(i+1)}_{r,h;t}}) \\ 
    &=f_a(\sum^{p_{i+1}}_{k=1}{{oW}^{(i+1)}_{k,h;t}}+\sum^{p_{i+2}}_{r=1}{{oH}^{(i+1)}_{r,h;t}}) \\ 
    &=x^{(i+2)}_{h;t},h=1,\dots,{p_{i+2}}.
    \end{aligned}
    \label{proof-p1:wider-new-output-i+1}
\end{equation}
Therefore, we have the output features of layer $i'+1$
\begin{equation}
    {\boldsymbol{x}}^{(i'+2)}_t=(x^{(i+2)}_{1,t},x^{(i+2)}_{2,t},\dots ,x^{(i+2)}_{{p_{i+2}},t}).
    \label{proof-p1:wider-same-output1}
\end{equation}
From Eq. \eqref{proof-p1:wider-same-output1}, the output feature vector ${\boldsymbol{x}}^{(i'+2)}_t$ of layer $i'+1$ is identical to that of layer $i+1$, i.e., ${\boldsymbol{x}}^{(i'+2)}_t=$ ${\boldsymbol{x}}^{(i+2)}_t$, in line with the function-preserving transformation in Eq. \eqref{eq:function-preserving}, which indicates that the wider transformation for layer $i$ is a function-preserving transformation. Hence, the proof of Proposition \ref{proposition: RNN-wider} is completed. \hfill 
\end{proof}

\begin{proof}
{Proof of Proposition \ref{proposition: LSTM-wider}:}
First, we consider the first stage processing for the \textit{wider} layer $i'$. We apply the new student network parameters of $\boldsymbol{W}^{(i')}_f\in \mathbb{R}^{p_i \times p'_{i+1}}, \boldsymbol{W}^{(i')}_i\in \mathbb{R}^{p_i \times p'_{i+1}}$, $\boldsymbol{W}^{(i')}_o\in \mathbb{R}^{p_i \times p'_{i+1}}$, $\boldsymbol{W}^{(i')}_c\in \mathbb{R}^{p_i \times p'_{i+1}}$, $\boldsymbol{H}^{(i')}_f\in \mathbb{R}^{{p'_{i+1}}\times {p'_{i+1}}}$, $\boldsymbol{H}^{(i')}_i\in \mathbb{R}^{{p'_{i+1}}\times {p'_{i+1}}}$, $\boldsymbol{H}^{(i')}_o\in \mathbb{R}^{{p'_{i+1}}\times {p'_{i+1}}}$ and $\boldsymbol{H}^{(i')}_c\in \mathbb{R}^{{p'_{i+1}}\times {p'_{i+1}}}$ in Proposition \ref{proposition: LSTM-wider} to Eq. \eqref{eq:extraction-LSTM}, and the extracted features at time $t$ are given by:
\begin{equation}
\begin{aligned}
        {oW}^{(i')}_{f;j,k;t}&={oW}^{(i)}_{f;j,g(k);t},j=1,2,\dots ,p_i,k=1,\dots,{p'_{i+1}},\\
        {oW}^{(i')}_{i;j,k;t}&={oW}^{(i)}_{i;j,g(k);t},j=1,2,\dots ,p_i,k=1,\dots,{p'_{i+1}},\\
        {oW}^{(o')}_{f;j,k;t}&={oW}^{(i)}_{o;j,g(k);t},j=1,2,\dots ,p_i,k=1,\dots,{p'_{i+1}},\\
        {oW}^{(c')}_{c;j,k;t}&={oW}^{(i)}_{c;j,g(k);t},j=1,2,\dots ,p_i,k=1,\dots,{p'_{i+1}},\\
        {oH}^{(i')}_{f;l,k;t}&=f_w(l){oH}^{(i)}_{f;g(l),g(k);t},l=1,2,\dots ,{p'_{i+1}},k=1,\dots,{p'_{i+1}},\\
        {oH}^{(i')}_{i;l,k;t}&=f_w(l){oH}^{(i)}_{i;g(l),g(k);t},l=1,2,\dots ,{p'_{i+1}},k=1,\dots,{p'_{i+1}},\\
        {oH}^{(i')}_{o;l,k;t}&=f_w(l){oH}^{(i)}_{o;g(l),g(k);t},l=1,2,\dots ,{p'_{i+1}},k=1,\dots,{p'_{i+1}},\\
        {oH}^{(i')}_{c;l,k;t}&=f_w(l){oH}^{(i)}_{c;g(l),g(k);t},l=1,2,\dots ,{p'_{i+1}},k=1,\dots,{p'_{i+1}}.
\end{aligned}
    \label{proof-p1:oWoH-LSTM}
\end{equation}
Following Eq. \eqref{eq:extraction-result-LSTM}, the feature extraction results of layer $i'\ $at time $t$, ${\boldsymbol{o}}^{(i')}_{k;t}$, are given by:
\begin{equation}
\begin{aligned}
     {\boldsymbol{o}}^{(i')}_{k;t}=(
    &{oW}^{(i')}_{f;1,k;t},\dots{oW}^{(i')}_{f;p_i,k;t},
    {oW}^{(i')}_{i;1,k;t},\dots{oW}^{(i')}_{i;p_i,k;t},\\
    &{oW}^{(i')}_{c;1,k;t},\dots{oW}^{(i')}_{c;p_i,k;t},
    {oW}^{(i')}_{o;1,k;t},\dots{oW}^{(i')}_{o;p_i,k;t},\\
    &{oH}^{(i')}_{f;1,k;t},\dots ,{oH}^{(i')}_{f;p_{i+1}k;t},
    {oH}^{(i')}_{i;1,k;t},\dots ,{oH}^{(i')}_{i;p_{i+1},k;t},\\
    &{oH}^{(i')}_{c;1,k;t},\dots ,{oH}^{(i')}_{c;p_{i+1},k;t},
    {oH}^{(i')}_{o;1,k;t},\dots ,{oH}^{(i')}_{o;p_{i+1},k;t},
    ),k=1,\dots,{p'_{i+1}}.
\end{aligned}
    \label{proof-p1:oi-LSTM}
\end{equation}
Then, by Eq. \eqref{eq:aggretation-cell-state-LSTM} and Eq. \eqref{eq:aggretation-hidden-state-LSTM} the new cell state $c^{(i')}_{k;t}$ is given by:
\begin{equation}
\begin{aligned}
c^{(i')}_{k;t}&=f_a(f_{2;c}({\boldsymbol{o}}^{(i')}_{k;t})) \\ 
&=f_a(\sigma_g(\sum^{p_i}_{j=1}{oW}^{(i')}_{f;j,k;t}+\sum^{p'_{i+1}}_{l=1}{oH}^{(i')}_{f;l,k;t}) \cdot c^{(i')} _{k;t-1}\\&\ \ + \sigma_g(\sum^{p_i}_{j=1}{oW}^{(i')}_{i;j,k;t}+\sum^{p'_{i+1}}_{l=1}{oH}^{(i')}_{i;l,k;t}) \cdot \sigma_c(\sum^{p_i}_{j=1}{oW}^{(i')}_{c;j,k;t}+\sum^{p'_{i+1}}_{l=1}{oH}^{(i')}_{c;l,k;t})) \\ 
&=f_a(\sigma_g(\sum^{p_i}_{j=1}{oW}^{(i)}_{f;j,g(k);t}+\sum^{p'_{i+1}}_{l=1}f_w(l){oH}^{(i)}_{f;g(l),g(k);t}) \cdot c^{(i)} _{g(k);t-1}\\&\ \ + \sigma_g(\sum^{p_i}_{j=1}{oW}^{(i)}_{i;j,g(k);t}+\sum^{p'_{i+1}}_{l=1}f_w(l){oH}^{(i)}_{i;g(l),g(k);t}) \cdot \sigma_c(\sum^{p_i}_{j=1}{oW}^{(i)}_{c;j,g(k);t}+\sum^{p'_{i+1}}_{l=1}f_w(l){oH}^{(i)}_{c;g(l),g(k);t})) \\
&=f_a(\sigma_g(\sum^{p_i}_{j=1}{oW}^{(i)}_{f;j,g(k);t}+\sum^{p_{i+1}}_{l=1}{oH}^{(i)}_{f;l,g(k);t}) \cdot c^{(i)} _{g(k);t-1}\\&\ \ + \sigma_g(\sum^{p_i}_{j=1}{oW}^{(i)}_{i;j,g(k);t}+\sum^{p_{i+1}}_{l=1}{oH}^{(i)}_{i;l,g(k);t}) \cdot \sigma_c(\sum^{p_i}_{j=1}{oW}^{(i)}_{c;j,g(k);t}+\sum^{p_{i+1}}_{l=1}{oH}^{(i)}_{c;l,g(k);t})) \\
&=c^{(i)}_{g(k);t}\ \ {,\ \ }k=1,\dots,{p'_{i+1}},
\end{aligned}
\label{proof-p1:wider-new-cell-output-LSTM}
\end{equation}
where the third equality holds by Eq. \eqref{proof-p1:oWoH-LSTM}, the fourth equality holds by replication factor in Eq. \eqref{eq:replication-factor}, and the last equality holds by the definition of the aggregating function of the cell state in Eq. \eqref{eq:aggretation-cell-state-LSTM}. Moreover, the second and third equality implies $ c^{(i')} _{k;t-1} =  c^{(i)} _{g(k);t-1}$, which can be proved by the following mathematical induction. Initially, we have $c^{(i)}_{k,0}=0,k=1,\dots,p_{i+1}$ and $ c^{(i')} _{k;0}=0,k=1,\dots,{p'_{i+1}}$. When $t=1$, we have $ c^{(i')} _{k;t-1} =  c^{(i)} _{g(k);t-1} = 0$. Suppose we have $ c^{(i')} _{k;t-1} =  c^{(i)} _{g(k);t-1}$, then we have $ c^{(i')} _{k;t} =  c^{(i)} _{g(k);t}$ according to Eq. (\ref{proof-p1:wider-new-cell-output-LSTM}). Hence, we can obtain a conclusion that $ c^{(i')} _{k;t-1} =  c^{(i)} _{g(k);t-1}$ holds for any $t$. The new output $x^{(i'+1)}_{k;t}$ is subsequently given by:

\begin{equation}
\begin{aligned}
x^{(i'+1)}_{k;t}&=f_a(f_{2;x}({\boldsymbol{o}}^{(i')}_{k;t})) \\ 
&=f_a((\sum^{p_i}_{j=1}{oW}^{(i')}_{o;j,k;t}+\sum^{p'_{i+1}}_{l=1}{oH}^{(i')}_{o;l,k;t}) \cdot  (c^{(i')} _{k;t}+1)) \\ 
&=f_a((\sum^{p_i}_{j=1}{oW}^{(i)}_{o;j,g(k);t}+\sum^{p'_{i+1}}_{l=1}f_w(l){oH}^{(i)}_{o;g(l),g(k);t}) \cdot  (c^{(i)}_{g(k);t}+1)) \\ 
&=f_a((\sum^{p_i}_{j=1}{oW}^{(i)}_{o;j,g(k);t}+\sum^{p_{i+1}}_{l=1}{oH}^{(i)}_{o;l,g(k);t}) \cdot  (c^{(i)}_{g(k);t}+1)) \\ 
&=x^{(i+1)}_{g(k);t}\ \ {,\ \ }k=1,\dots,{p'_{i+1}},
\end{aligned}
\label{proof-p1:wider-new-state-output-LSTM}
\end{equation}
where the third equality holds by Eq. \eqref{proof-p1:oWoH-LSTM}, the fourth equality holds by replication factor in Eq. \eqref{eq:replication-factor}, and the last equality holds by the aggregating function of the hidden state in Eq. \eqref{eq:aggretation-hidden-state-LSTM}. Hence, the output feature vector ${\boldsymbol{x}}^{(i'+1)}_{\ t}$ of layer $i'$ are the input feature vector of layer $i'+1$. Following Eq. \eqref{eq:new-output-hidden-state-LSTM}, ${\boldsymbol{x}}^{(i'+1)}_{\ t}$ can be represented as
\begin{equation}
  {\boldsymbol{x}}^{(i'+1)}_t=(x^{(i+1)}_{g(1);t},x^{(i+1)}_{g(2);t},\dots ,x^{(i+1)}_{g({p'_{i+1}});t}).  
  \label{proof-p1:input-of-i+1-LSTM}
\end{equation}
As the same as that in the wider transformer of RNN, ${\boldsymbol{x}}^{(i'+1)}_t$ is also a random mapping defined by $g$. Similar to the LSTM reformulation in wider layer $i'$ as shown above, we apply the new student network parameters of $\boldsymbol{W}^{(i'+1)}_f\in \mathbb{R}^{p'_{i+1}\times p_{i+2}}, \boldsymbol{W}^{(i'+1)}_i\in \mathbb{R}^{p'_{i+1}\times p_{i+2}}$, $\boldsymbol{W}^{(i'+1)}_o\in \mathbb{R}^{p'_{i+1}\times p_{i+2}}$, $\boldsymbol{W}^{(i'+1)}_c\in \mathbb{R}^{p'_{i+1}\times p_{i+2}}$, $\boldsymbol{H}^{(i'+1)}_f\in \mathbb{R}^{{p_{i+2}}\times {p_{i+2}}}$, $\boldsymbol{H}^{(i'+1)}_i\in \mathbb{R}^{{p_{i+2}}\times {p_{i+2}}}$, $\boldsymbol{H}^{(i'+1)}_o\in \mathbb{R}^{{p_{i+2}}\times {p_{i+2}}}$, and $\boldsymbol{H}^{(i'+1)}_c\in \mathbb{R}^{{p_{i+2}}\times {p_{i+2}}}$ in Proposition \ref{proposition: LSTM-wider} to Eq. \eqref{eq:extraction-LSTM}, and acquire the extracted features for the next layer $i'+1$ as follows:
\begin{equation}
    {oW}^{(i'+1)}_{\alpha;k,h;t}=f_w(k){oW}^{(i+1)}_{\alpha;g(k),h;t},\alpha \in \{f,i,o,c\},k=1,2,\dots ,{p'_{i+1}},h=1,\dots,{p_{i+2}},
    \label{proof-p1:oWi+1-LSTM}
\end{equation}
\begin{equation}
    {oH}^{(i'+1)}_{\alpha;r,h;t}={oH}^{(i+1)}_{\alpha;r,h;t},\alpha \in \{f,i,o,c\},r=1,2,\dots ,{p_{i+2}},h=1,\dots,{p_{i+2}}.
    \label{proof-p1:oHi+1-LSTM}
\end{equation}
As observed in Eq. \eqref{proof-p1:oHi+1-LSTM}, the feature extraction for outputs of layer $i'+1$ is the same as that of layer $i+1$ since the dimension is unchanged. Following Eq. \eqref{eq:extraction-result-LSTM}, the feature extraction results of layer $i'+1\ $at time $t$, ${\boldsymbol{o}}^{(i'+1)}_{h;t}$, are given by
\begin{equation}
\begin{aligned}
     {\boldsymbol{o}}^{(i'+1)}_{h;t}=(
    &{oW}^{(i'+1)}_{f;1,h;t},\dots{oW}^{(i'+1)}_{f;{p'_{i+1}},h;t},
    {oW}^{(i'+1)}_{i;1,h;t},\dots{oW}^{(i'+1)}_{i;{p'_{i+1}},h;t},\\
    &{oW}^{(i'+1)}_{c;1,h;t},\dots{oW}^{(i'+1)}_{c;{p'_{i+1}},h;t},
    {oW}^{(i'+1)}_{o;1,h;t},\dots{oW}^{(i'+1)}_{o;{p'_{i+1}},h;t},\\
    &{oH}^{(i'+1)}_{f;1,h;t},\dots ,{oH}^{(i'+1)}_{f;{p_{i+2}},h;t},
    {oH}^{(i'+1)}_{i;1,h;t},\dots ,{oH}^{(i'+1)}_{i;{p_{i+2}},h;t},\\
    &{oH}^{(i'+1)}_{c;1,h;t},\dots ,{oH}^{(i'+1)}_{c;{p_{i+2}},h;t},
    {oH}^{(i'+1)}_{o;1,h;t},\dots ,{oH}^{(i'+1)}_{o;{p_{i+2}},h;t},
    ),h=1,\dots,{p_{i+2}}.
\end{aligned}
    \label{proof-p1:oi+1-LSTM}
\end{equation}

Next, we aggregate the extracted features with the aggregating functions, Eq. (\ref{eq:aggretation-cell-state-LSTM}) and (\ref{eq:aggretation-hidden-state-LSTM}), to generate the new cell state $c^{(i'+1)}_{h;t}$ and new output $x^{(i'+2)}_{h;t}$, $h=1,\dots,{p_{i+2}}.$ Similar to the analysis in Eq. \eqref{proof-p1:wider-new-cell-output-LSTM} and  Eq. \eqref{proof-p1:wider-new-state-output-LSTM}, by incorporating
Eq. (\ref{eq:aggretation-cell-state-LSTM}-\ref{eq:new-output-hidden-state-LSTM}) and Eq. (\ref{eq:random-mapping}-\ref{eq:replication-factor}), we have the following cell state:
\begin{equation}
    \begin{aligned}
    c^{(i'+1)}_{h;t}
    &=\ f_a(f_{2;c}({\boldsymbol{o}}^{(i'+1)}_{h;t})) \\ 
    &=\ f_a(\sigma_g(\sum^{p'_{i+1}}_{k=1}{oW}^{(i'+1)}_{f;k,h;t}+\sum^{p_{i+2}}_{l=1}{oH}^{(i'+1)}_{f;r,h;t}) \cdot c^{(i'+1)} _{h;t-1}\\&\ \ + \sigma_g(\sum^{p'_{i+1}}_{k=1}{oW}^{(i'+1)}_{i;k,h;t}+\sum^{p_{i+2}}_{r=1}{oH}^{(i'+1)}_{i;r,h;t}) \cdot \sigma_c(\sum^{p'_{i+1}}_{k=1}{oW}^{(i'+1)}_{c;k,h;t}+\sum^{p_{i+2}}_{r=1}{oH}^{(i'+1)}_{c;r,h;t})) \\ 
    &=\ f_a(\sigma_g(\sum^{p'_{i+1}}_{k=1}f_w(k){oW}^{(i+1)}_{f;g(k),h;t}+\sum^{p_{i+2}}_{l=1}{oH}^{(i+1)}_{f;r,h;t}) \cdot c^{(i+1)} _{h;t-1}\\&\ \ + \sigma_g(\sum^{p'_{i+1}}_{k=1}f_w(k){oW}^{(i+1)}_{i;g(k),h;t}+\sum^{p_{i+2}}_{r=1}{oH}^{(i+1)}_{i;r,h;t}) \cdot \sigma_c(\sum^{p'_{i+1}}_{k=1}f_w(k){oW}^{(i+1)}_{c;g(k),h;t}+\sum^{p_{i+2}}_{r=1}{oH}^{(i+1)}_{c;r,h;t})) \\ 
    &=\ f_a(\sigma_g(\sum^{p_{i+1}}_{k=1}{oW}^{(i+1)}_{f;k,h;t}+\sum^{p_{i+2}}_{l=1}{oH}^{(i+1)}_{f;r,h;t}) \cdot c^{(i+1)} _{h;t-1}\\&\ \ + \sigma_g(\sum^{p_{i+1}}_{k=1}{oW}^{(i+1)}_{i;k,h;t}+\sum^{p_{i+2}}_{r=1}{oH}^{(i+1)}_{i;r,h;t}) \cdot \sigma_c(\sum^{p_{i+1}}_{k=1}{oW}^{(i+1)}_{c;k,h;t}+\sum^{p_{i+2}}_{r=1}{oH}^{(i+1)}_{c;r,h;t})) \\ 
    &=c^{(i+1)}_{h;t},h=1,\dots,{p_{i+2}},
    \end{aligned}
    \label{proof-p1:wider-new-cell-output-i+1-LSTM}
\end{equation}
and  hidden state:
\begin{equation}
    \begin{aligned}
    x^{(i'+2)}_{h;t}
    &=\ f_{a}(f_{2;x}({\boldsymbol{o}}^{(i'+1)}_{h;t})) \\ 
    &=f_a((\sum^{p'_{i+1}}_{k=1}{oW}^{(i'+1)}_{o;k,h;t}+\sum^{p_{i+2}}_{r=1}{oH}^{(i'+1)}_{o;r,h;t}) \cdot  (c^{(i'+1)} _{h;t}+1)) \\ 
    &=f_a((\sum^{p'_{i+1}}_{k=1}f_w(k){oW}^{(i+1)}_{o;g(k),h;t}+\sum^{p_{i+2}}_{r=1}{oH}^{(i+1)}_{o;r,h;t}) \cdot  (c^{(i+1)}_{h;t}+1)) \\ 
    &=f_a((\sum^{p_{i+1}}_{k=1}{oW}^{(i)}_{o;k,h;t}+\sum^{p_{i+2}}_{r=1}{oH}^{(i)}_{o;r,h;t}) \cdot  (c^{(i+1)}_{h;t}+1)) \\ 
        &=x^{(i+2)}_{h;t},h=1,\dots,{p_{i+2}}.
    \end{aligned}
    \label{proof-p1:wider-new-hidden-output-i+1-LSTM}
\end{equation}
Therefore, we have the output feature vector of layer $i'+1$
\begin{equation}
    {\boldsymbol{x}}^{(i'+2)}_t=(x^{(i+2)}_{1,t},x^{(i+2)}_{2,t},\dots ,x^{(i+2)}_{{p_{i+2}},t}).
    \label{proof-p1:wider-same-output2}
\end{equation}
From Eq. \eqref{proof-p1:wider-same-output2}, the output feature vector ${\boldsymbol{x}}^{(i'+2)}_t$ of layer $i'+1$ is identical to that of layer $i+1$, i.e., ${\boldsymbol{x}}^{(i'+2)}_t=$ ${\boldsymbol{x}}^{(i+2)}_t$, which indicates that the wider transformation for layer $i$ is a function-preserving transformation. Hence, the proof of Proposition \ref{proposition: LSTM-wider} is completed. \hfill 
\end{proof}

\begin{proof}{Proof of Proposition \ref{proposition: RNN-deeper}:}
First, we consider the feature extraction for the deeper layer $i''$. We apply the new network parameters of ${\boldsymbol{W}}^{(i^{''})}$\textit{ }and\textit{ }${\boldsymbol{H}}^{(i^{''})}$ in Proposition \ref{proposition: RNN-deeper} to Eq. \eqref{eq:extraction-W} and \eqref{eq:extraction-H}, and we get the extracted features for inputs at time $t$ and outputs at time $t+1$ as follows:
\begin{align}
 \quad \quad {oW}^{(i^{''})}_{l,k;t}&=f_1(x^{(i^{''})}_{l;t},W^{(i^{''})}_{l,k})=
    \begin{cases}
    x^{(i^{''})}_{l;t}& l=k \\ 
    0& l\neq k 
    \end{cases}
,&l=1,2,\dots ,{p_{i+1}},k=1,\dots,{p_{i+1}}, \quad \quad \quad
    \label{proof-p2:oW} \\
{oH}^{(i^{''})}_{l,k;t}&=f_1(x^{(i^{''}+1)}_{l;t-1},H^{(i^{''})}_{l,k})=0,&l=1,2,\dots,{p_{i+1}},k=1,\dots,{p_{i+1}}. \quad \quad \quad
    \label{proof-p2:oH}
\end{align}
Based on Eq. \eqref{eq:extraction-result}, the feature extraction results (${\boldsymbol{o}}^{(i^{''})}_{k;t}$) of layer $i''$ at time $t$, are described as
\begin{equation}
    {\boldsymbol{o}}^{(i^{''})}_{k;t}=(0,\dots x^{(i^{''})}_{k;t},\dots ,0,0,\dots ,0){,\ \ \ \ \ \ \ }k=1,\dots,{p_{i+1}}.
    \label{proof-p2:o}
\end{equation}
By evaluating the aggregating function $f_2({\boldsymbol{o}}^{(i^{''})}_{k;t})=\sum^{p_{i+1}}_{l=1}{{oW}^{(i^{''})}_{l,k;t}}+\sum^{p_{i+1}}_{l=1}{{oH}^{(i^{''})}_{l,k;t}=x^{(i^{''})}_{k;t}}$, for $k=1,\dots,{p_{i+1}}$, the new output $x^{(i^{''}+1)}_{k;t}$ is then obtained:
\begin{equation}
    x^{(i^{''}+1)}_{k;t}=f_a(f_2({\boldsymbol{o}}^{(i^{''})}_{k;t}))=f_a(x^{(i^{''})}_{k;t})=x^{(i^{''})}_{k;t},k=1,\dots,{p_{i+1}},
    \label{proof-p2:new-output1}
\end{equation}
Therefore, we have:
\begin{equation}
	{\boldsymbol{x}}^{(i^{''}+1)}_t={\boldsymbol{x}}^{(i^{''})}_t.
	\label{proof-p2:same-output}
\end{equation}
As observed in Eq. \eqref{proof-p2:same-output}, the input and output features are identical, verifying the function-preserving transformation for layer $i\ $in Eq. \eqref{eq:function-preserving}. Hence, the proof of Proposition \ref{proposition: RNN-deeper} is completed. \hfill 
\end{proof}

\begin{proof}{Proof of Proposition \ref{proposition: LSTM-deeper}:}
First, we take the feature extraction for the deeper layer $i''$ into account. We leverage the new network parameters of $\boldsymbol{W}^{(i)}_f\in \mathbb{R}^{p_{i+1}\times p_{i+1}}, \boldsymbol{W}^{(i)}_i\in \mathbb{R}^{p_{i+1}\times p_{i+1}}$, $\boldsymbol{W}^{(i)}_o\in \mathbb{R}^{p_{i+1}\times p_{i+1}}$, $\boldsymbol{W}^{(i)}_c\in \mathbb{R}^{p_{i+1}\times p_{i+1}}$, $\boldsymbol{H}^{(i)}_f\in \mathbb{R}^{p_{i+1}\times p_{i+1}}$, $\boldsymbol{H}^{(i)}_i\in \mathbb{R}^{p_{i+1}\times p_{i+1}}$, $\boldsymbol{H}^{(i)}_o\in \mathbb{R}^{p_{i+1}\times p_{i+1}}$ and $\boldsymbol{H}^{(i)}_c\in \mathbb{R}^{p_{i+1}\times p_{i+1}}$ in Proposition \ref{proposition: LSTM-deeper} to Eq. \eqref{eq:extraction-LSTM} , and we obtain the extracted features for inputs at time $t$ and outputs at time $t+1$ as follows:
\begin{align}
   {oW}^{(i^{''})}_{o;l,k;t}&=
    \begin{cases}
    x^{(i^{''})}_{l;t}& l=k \\ 
    0& l\neq k 
    \end{cases}
,\quad l=1,2,\dots ,p_{i+1},k=1,\dots,p_{i+1},    \\
{oW}^{(i^{''})}_{\alpha;l,k;t}&=0,\quad\quad\alpha\in\{f,i,c\},l=1,2,\dots,p_{i+1},k=1,\dots,p_{i+1},   \\
{oH}^{(i^{''})}_{\alpha;l,k;t}&=0,\quad\quad\alpha\in\{f,i,o,c\},l=1,2,\dots,p_{i+1},k=1,\dots,p_{i+1}.
    \label{proof-p2:oWoH}
\end{align}
By evaluating the aggregating functions, $f_{2;c}$ and $f_{2;x}$, we can obtain the new cell state:
\begin{equation}
\begin{aligned}
    c^{(i'')} _{k;t}& = f_{2;c}({\boldsymbol{o}}^{(i)}_{k;t})=
    \sigma_g(\sum^{p_{i+1}}_{j=1}{oW}^{(i)}_{f;j,k;t}+\sum^{p_{i+1}}_{l=1}{oH}^{(i)}_{f;l,k;t}) \cdot c^{(i)} _{k;t-1}\\ + &\sigma_g(\sum^{p_{i+1}}_{j=1}{oW}^{(i)}_{i;j,k;t}+\sum^{p_{i+1}}_{l=1}{oH}^{(i)}_{i;l,k;t}) \cdot \sigma_c(\sum^{p_{i+1}}_{j=1}{oW}^{(i)}_{c;j,k;t}+\sum^{p_{i+1}}_{l=1}{oH}^{(i)}_{c;l,k;t})\\
    &=\boldsymbol{0}
    ,k=1,\dots,p_{i+1}.
\end{aligned}\label{proof-p2:new-cell-state-LSTM}
\end{equation}
And the new hidden state $x^{(i^{''}+1)}_{k;t}$ is:
\begin{equation}
\begin{aligned}
 x^{(i^{''}+1)}_{k;t}&=f_a(f_{2;x}({\boldsymbol{o}}^{(i^{''})}_{k;t}))\\
 &=(\sum^{p_{i+1}}_{j=1}{oW}^{(i'')}_{o;j,k;t}+\sum^{p_{i+1}}_{l=1}{oH}^{(i'')}_{o;l,k;t}) \cdot  (f_{2,c}({\boldsymbol{o}}^{(i'')}_{k;t})+1)\\
 &=x^{(i^{''})}_{k;t},k=1,\dots,p_{i+1}.
\end{aligned}
    \label{proof-p2:new-output2}
\end{equation}
Therefore, we have:
\begin{equation}
	{\boldsymbol{x}}^{(i^{''}+1)}_t={\boldsymbol{x}}^{(i^{''})}_t.
	\label{proof-p2:same-output-LSTM}
\end{equation}
As seen in Eq. \eqref{proof-p2:same-output-LSTM}, the input and output features are the same, which verifies the function-preserving transformation for layer $i\ $in Eq. \eqref{eq:function-preserving}. Hence, the proof of Proposition \ref{proposition: LSTM-deeper} is completed. \hfill 
\end{proof}

\clearpage
\section{Pruning Formulations of FCN, RNN, CNN and LSTM}\label{appendix:pruning-cnn-lstm}
Consider a specific case of pruning a single FCN layer $i$ with $p_i$ input features and $p_{i+1}$ output features. The FCN layer $i+1$ has ${p_{i+2}}$ output features, then the parameters of the layers $i$ and $i+1$ are ${\boldsymbol{W}}^{(i)}\in {\mathbb{R}}^{p_i \times p_{i+1}}$ and ${\boldsymbol{W}}^{(i{+1)}}\in {\mathbb{R}}^{p_{i+1}\times {p_{i+2}}}$,  respectively. If we prune layer $i$ to layer $i'$ with ${p'_{i+1}}$ output features (${p'_{i+1}}<p_{i+1}$) according to the importance score $\boldsymbol{\lambda}^{(i+1)}$ where ${\lambda}^{(i+1)}_k = \sum \Lambda ^{(i)}_{l,k}(l=1,\cdots,p_i)$ and $\boldsymbol{\Lambda}^{(i)}$ is the importance score of $\boldsymbol{W}^{(i)}$, the original network parameters of layers $i$ and $i+1$ are replaced by ${\boldsymbol{W}}^{(i')}\in {\mathbb{R}}^{p_i\times {p'_{i+1}}}$ and ${\boldsymbol{W}}^{(i'{+1)}}\in {\mathbb{R}}^{{p'_{i+1}}\times {p_{i+2}}}$, separately. The mask $\boldsymbol{m}^{(i+1)} = Top_v(\boldsymbol{\lambda}^{(i+1)})$ is a vector with $p_{i+1}$ dimensions, and we can obtain the mapping function $g$ satisfying the following conditions:
\begin{align}
    {m}^{(i+1)}_{g(k)} &= 1 ,\ k=1,2,\dots,{p'_{i+1}},\label{eq:prune-condition-topv}\\ 
    |\{l|g(k)=g(l)\}| &= 1,l=1,2,\dots,p'_{i+1},\ k=1,2,\dots,{p'_{i+1}}.\label{eq:prune-condition-no-duplicate}
\end{align}
Eq. (\ref{eq:prune-condition-topv}) requires that $g$ choose the neurons that are in the highest $v\%$ and Eq. (\ref{eq:prune-condition-no-duplicate}) guarantees that each chosen neuron is distinct, which means that no neuron is overlooked and there are no duplicates. After pruning, we can obtain the updated parameters for this FCN layer as:
\begin{align*}
    \quad \quad \quad \quad \quad W^{(i')}_{j,k}&=W^{(i)}_{j,g(k)},&j=1,2,\dots ,p_i,k=1,\dots,{p'_{i+1}}, \quad \quad \quad \quad \quad
    \\
    W^{(i'+1)}_{k,h}&=W^{(i+1)}_{g(k),h},&k=1,2,\dots ,{p'_{i+1}},h=1,\dots,{p_{i+2}}. \quad \quad \quad \quad \quad
\end{align*}

Moreover, we consider pruning a single RNN layer $i$ with $p_i$ input features and $p_{i+1}$ output features. The RNN layer $i+1$ has ${p_{i+2}}$ output features; then the parameters of the layers $i$ and $i+1$ are ${\boldsymbol{W}}^{(i)}\in {\mathbb{R}}^{p_i \times p_{i+1}}$, ${\boldsymbol{H}}^{(i)}\in {\mathbb{R}}^{p_{i+1}\times p_{i+1}}$ and ${\boldsymbol{W}}^{(i{+1)}}\in {\mathbb{R}}^{p_{i+1}\times {p_{i+2}}}$, ${\boldsymbol{H}}^{(i{+1)}}\in {\mathbb{R}}^{{p_{i+2}}\times {p_{i+2}}}$, respectively. If we prune layer $i$ to layer $i'$ with ${p'_{i+1}}$ output features (${p'_{i+1}}<p_{i+1}$) according to the importance score $\boldsymbol{\lambda}^{(i+1)}$, the original network parameters of layers $i$ and $i+1$ are replaced by ${\boldsymbol{W}}^{(i')}\in {\mathbb{R}}^{p_i\times {p'_{i+1}}}$, ${\boldsymbol{H}}^{(i')}\in {\mathbb{R}}^{{p'_{i+1}}\times {p'_{i+1}}}$ and ${\boldsymbol{W}}^{(i'{+1)}}\in {\mathbb{R}}^{{p'_{i+1}}\times {p_{i+2}}}$, ${\boldsymbol{H}}^{(i'{+1)}}\in {\mathbb{R}}^{{p_{i+2}}\times {p_{i+2}}}$, separately. Here,  ${\lambda}^{(i+1)}_k = \sum_l \Lambda w^{(i)}_{l,k} + \sum_h \Lambda h^{(i)}_{h,k}(l=1,\cdots,p_i; h=1,\cdots,p_{i+1})$ and $\boldsymbol{\Lambda w}^{(i)}$ is the importance score of $\boldsymbol{W}^{(i)}$, $\boldsymbol{\Lambda h}^{(i)}$ is the importance score of $\boldsymbol{H}^{(i)}$. The mask $\boldsymbol{m}^{(i+1)} = Top_v(\boldsymbol{\lambda}^{(i+1)})$ is a vector with $p_{i+1}$ dimensions. After pruning, we can achieve the parameters for this pruned RNN layer as: 
\begin{align*}
    \quad \quad \quad \quad \quad W^{(i')}_{j,k}&=W^{(i)}_{j,g(k)},&j=1,2,\dots ,p_i,k=1,\dots,{p'_{i+1}}, \quad \quad \quad \quad \quad
    \\
    H^{(i')}_{l,k}&=H^{(i)}_{g(l),g(k)},&l=1,2,\dots ,{p'_{i+1}},k=1,\dots,{p'_{i+1}}, \quad \quad \quad \quad \quad
        \\
    W^{(i'+1)}_{k,h}&=W^{(i+1)}_{g(k),h},&k=1,2,\dots ,{p'_{i+1}},h=1,\dots,{p_{i+2}}, \quad \quad \quad \quad \quad
        \\
    H^{(i'+1)}_{r,h}&=H^{(i+1)}_{r,h},&r=1,2,\dots ,{p_{i+2}},h=1,\dots,{p_{i+2}}. \quad \quad \quad \quad \quad
\end{align*}

Furthermore, we consider pruning a single one-dimensional CNN layer $i$ with $p_i$ input channels and $p_{i+1}$ output channels. The CNN layer $i+1$ has ${p_{i+2}}$ output channels. The parameters of the layers $i$ and $i+1$ are then ${\boldsymbol{W}}^{(i)}\in {\mathbb{R}}^{p_i \times p_{i+1} \times d_k}$ and ${\boldsymbol{W}}^{(i{+1)}}\in {\mathbb{R}}^{p_{i+1}\times {p_{i+2}}  \times d_k} $, respectively, where $d_k$ is the kernel size of the one-dimensional CNN layer. If we prune layer $i$ to layer $i'$ with ${p'_{i+1}}$ output channels (${p'_{i+1}}<p_{i+1}$) according to the importance score $\boldsymbol{\lambda}^{(i+1)}$, the original network parameters of layers $i$ and $i+1$ are replaced by ${\boldsymbol{W}}^{(i')}\in {\mathbb{R}}^{p_i\times {p'_{i+1}}\times d_k}$ and ${\boldsymbol{W}}^{(i'{+1)}}\in {\mathbb{R}}^{{p'_{i+1}}\times {p_{i+2}}\times d_k}$, separately. Therefore, we can obtain the prune mask  $\boldsymbol{m}^{(i+1)} = Top_v(\boldsymbol{\lambda}^{(i+1)})$ the mapping function $g$ that satisfies the conditions in Eq. (\ref{eq:prune-condition-topv}) and (\ref{eq:prune-condition-no-duplicate}). After pruning, we can obtain the updated parameters for this CNN layer as:
\begin{align*}
    \quad \quad \quad \quad \quad W^{(i')}_{j,k,l}&=W^{(i)}_{j,g(k),l},&j=1,2,\dots ,p_i,k=1,\dots,{p'_{i+1}},l=1,\dots,d_k, \quad \quad \quad \quad \quad
    \\
    W^{(i'+1)}_{k,h,l}&=W^{(i+1)}_{g(k),h,l},&k=1,2,\dots ,{p'_{i+1}},h=1,\dots,{p_{i+2}},l=1,\dots,d_k. \quad \quad \quad \quad \quad
\end{align*}

Lastly, we consider pruning LSTM neurons. Assume LSTM layer $i$ has $p_i$ input and $p_{i+1}$ output features, and LSTM layer $i + 1$ has $p_{i+1}$ input and ${p_{i+2}}$ output features. Hence, the parameters of layer $i$ are $\boldsymbol{W}^{(i)}_f\in \mathbb{R}^{p_i \times p_{i+1}}, \boldsymbol{W}^{(i)}_i\in \mathbb{R}^{p_i \times p_{i+1}}$, $\boldsymbol{W}^{(i)}_o\in \mathbb{R}^{p_i \times p_{i+1}}$, $\boldsymbol{W}^{(i)}_c\in \mathbb{R}^{p_i \times p_{i+1}}$, $\boldsymbol{H}^{(i)}_f\in \mathbb{R}^{p_{i+1}\times p_{i+1}}$, $\boldsymbol{H}^{(i)}_i\in \mathbb{R}^{p_{i+1}\times p_{i+1}}$, $\boldsymbol{H}^{(i)}_o\in \mathbb{R}^{p_{i+1}\times p_{i+1}}$, and $\boldsymbol{H}^{(i)}_c\in \mathbb{R}^{p_{i+1}\times p_{i+1}}$; moreover, the parameters of layer $i+1$ are $\boldsymbol{W}^{(i+1)}_f\in \mathbb{R}^{p_{i+1}\times p_{i+2}}, \boldsymbol{W}^{(i+1)}_i\in \mathbb{R}^{p_{i+1}\times p_{i+2}}$, $\boldsymbol{W}^{(i+1)}_o\in \mathbb{R}^{p_{i+1}\times p_{i+2}}$, $\boldsymbol{W}^{(i+1)}_c\in \mathbb{R}^{p_{i+1}\times p_{i+2}}$, $\boldsymbol{H}^{(i+1)}_f\in \mathbb{R}^{{p_{i+2}}\times {p_{i+2}}}$, $\boldsymbol{H}^{(i+1)}_i\in \mathbb{R}^{{p_{i+2}}\times {p_{i+2}}}$, $\boldsymbol{H}^{(i+1)}_o\in \mathbb{R}^{{p_{i+2}}\times {p_{i+2}}}$, and $\boldsymbol{H}^{(i+1)}_c\in \mathbb{R}^{{p_{i+2}}\times {p_{i+2}}}$. If we prune layer $i$ to layer $i'$ with ${p'_{i+1}}$ output features (${p'_{i+1}}<p_{i+1}$), the original network parameters of layers $i$ are replaced by $\boldsymbol{W}^{(i')}_f\in \mathbb{R}^{p_i \times p'_{i+1}}, \boldsymbol{W}^{(i')}_i\in \mathbb{R}^{p_i \times p'_{i+1}}$, $\boldsymbol{W}^{(i')}_o\in \mathbb{R}^{p_i \times p'_{i+1}}$, $\boldsymbol{W}^{(i')}_c\in \mathbb{R}^{p_i \times p'_{i+1}}$, $\boldsymbol{H}^{(i')}_f\in \mathbb{R}^{{p'_{i+1}}\times {p'_{i+1}}}$, $\boldsymbol{H}^{(i')}_i\in \mathbb{R}^{{p'_{i+1}}\times {p'_{i+1}}}$, $\boldsymbol{H}^{(i')}_o\in \mathbb{R}^{{p'_{i+1}}\times {p'_{i+1}}}$, and $\boldsymbol{H}^{(i')}_c\in \mathbb{R}^{{p'_{i+1}}\times {p'_{i+1}}}$. In addition, the parameters of layer $i+1$ are replaced by $\boldsymbol{W}^{(i'+1)}_f\in \mathbb{R}^{p'_{i+1}\times p_{i+2}}, \boldsymbol{W}^{(i'+1)}_i\in \mathbb{R}^{p'_{i+1}\times p_{i+2}}$, $\boldsymbol{W}^{(i'+1)}_o\in \mathbb{R}^{p'_{i+1}\times p_{i+2}}$, $\boldsymbol{W}^{(i'+1)}_c\in \mathbb{R}^{p'_{i+1}\times p_{i+2}}$, $\boldsymbol{H}^{(i'+1)}_f\in \mathbb{R}^{{p_{i+2}}\times {p_{i+2}}}$, $\boldsymbol{H}^{(i'+1)}_i\in \mathbb{R}^{{p_{i+2}}\times {p_{i+2}}}$, $\boldsymbol{H}^{(i'+1)}_o\in \mathbb{R}^{{p_{i+2}}\times {p_{i+2}}}$, and $\boldsymbol{H}^{(i'+1)}_c\in \mathbb{R}^{{p_{i+2}}\times {p_{i+2}}}$. After calculating the importance score $\boldsymbol{\lambda}^{(i+1)}$ by automatic differentiation technique \citep{baydin2018automatic}, we can obtain the prune mask $\boldsymbol{m}^{(i+1)} = Top_v(\boldsymbol{\lambda}^{(i+1)})$ and the mapping function $g$. Consequently, we can achieve the updated parameters for this LSTM layer as:
\begin{align*}
    \quad \quad \quad \quad \quad W^{(i')}_{f;j,k}&=W^{(i)}_{f;j,g(k)},&j=1,2,\dots ,p_i,k=1,\dots,{p'_{i+1}}, \quad \quad \quad \quad \quad
    \\
    \quad \quad \quad \quad \quad W^{(i')}_{i;j,k}&=W^{(i)}_{i;j,g(k)},&j=1,2,\dots ,p_i,k=1,\dots,{p'_{i+1}}, \quad \quad \quad \quad \quad
    \\
    \quad \quad \quad \quad \quad W^{(i')}_{o;j,k}&=W^{(i)}_{o;j,g(k)},&j=1,2,\dots ,p_i,k=1,\dots,{p'_{i+1}}, \quad \quad \quad \quad \quad
    \\
    \quad \quad \quad \quad \quad W^{(i')}_{c;j,k}&=W^{(i)}_{c;j,g(k)},&j=1,2,\dots ,p_i,k=1,\dots,{p'_{i+1}}, \quad \quad \quad \quad \quad
    \\
    H^{(i')}_{f;l,k}&=f_w(l)H^{(i)}_{f;g(l),g(k)},&l=1,2,\dots ,{p'_{i+1}},k=1,\dots,{p'_{i+1}}, \quad \quad \quad \quad \quad
        \\
    H^{(i')}_{i;l,k}&=f_w(l)H^{(i)}_{i;g(l),g(k)},&l=1,2,\dots ,{p'_{i+1}},k=1,\dots,{p'_{i+1}}, \quad \quad \quad \quad \quad
        \\
    H^{(i')}_{o;l,k}&=f_w(l)H^{(i)}_{o;g(l),g(k)},&l=1,2,\dots ,{p'_{i+1}},k=1,\dots,{p'_{i+1}}, \quad \quad \quad \quad \quad
        \\
    H^{(i')}_{c;l,k}&=f_w(l)H^{(i)}_{c;g(l),g(k)},&l=1,2,\dots ,{p'_{i+1}},k=1,\dots,{p'_{i+1}}, \quad \quad \quad \quad \quad
        \\
    W^{(i'+1)}_{f;k,h}&=f_w(k)W^{(i+1)}_{f;g(k),h},&k=1,2,\dots ,{p'_{i+1}},h=1,\dots,{p_{i+2}}, \quad \quad \quad \quad \quad
        \\
    W^{(i'+1)}_{i;k,h}&=f_w(k)W^{(i+1)}_{i;g(k),h},&k=1,2,\dots ,{p'_{i+1}},h=1,\dots,{p_{i+2}}, \quad \quad \quad \quad \quad
        \\
    W^{(i'+1)}_{o;k,h}&=f_w(k)W^{(i+1)}_{o;g(k),h},&k=1,2,\dots ,{p'_{i+1}},h=1,\dots,{p_{i+2}}, \quad \quad \quad \quad \quad
        \\
    W^{(i'+1)}_{c;k,h}&=f_w(k)W^{(i+1)}_{c;g(k),h},&k=1,2,\dots ,{p'_{i+1}},h=1,\dots,{p_{i+2}}, \quad \quad \quad \quad \quad
        \\
    H^{(i'+1)}_{f;r,h}&=H^{(i+1)}_{f;r,h},&r=1,2,\dots ,{p_{i+2}},h=1,\dots,{p_{i+2}}, \quad \quad \quad \quad \quad 
    \\
    H^{(i'+1)}_{i;r,h}&=H^{(i+1)}_{i;r,h},&r=1,2,\dots ,{p_{i+2}},h=1,\dots,{p_{i+2}}, \quad \quad \quad \quad \quad 
    \\
    H^{(i'+1)}_{o;r,h}&=H^{(i+1)}_{o;r,h},&r=1,2,\dots ,{p_{i+2}},h=1,\dots,{p_{i+2}}, \quad \quad \quad \quad \quad 
    \\
    H^{(i'+1)}_{c;r,h}&=H^{(i+1)}_{c;r,h},&r=1,2,\dots ,{p_{i+2}},h=1,\dots,{p_{i+2}}. \quad \quad \quad \quad \quad 
\end{align*}
\newpage
\section{Operation of ACER Algorithm}\label{ACER}

Compared to the REINFORCE algorithm used in \cite{cai2018efficient}, we make use of the ACER algorithm (\citealt{wang2017sample}) to train the three RL agents. As presented in Section 3.2, the goal of the RL agents in this study is to maximize the discounted reward over transformation steps. For an agent following policy $\pi$, we use the standard definitions of the state-action and state only value function:
\begin{equation}
    Q^{\pi}(s_u,a_u) = E_{s_{u+1:\infty},a_{u+1:\infty}}(R_u|s_u,a_u),
    \label{eq:state-action-value-function}
\end{equation}
\begin{equation}
    V^{\pi}(s_u) = E_{a_u}( Q^{\pi}(s_u,a_u)|s_u),
    \label{eq:state-action-only-function}
\end{equation}
where the expectations are with respect to $s_u$ and $a_u$ generated by the policy $\pi$, and $s_{u+1:\infty}$ indicates a state trajectory starting at time $u+1$. In ACER, policy $\pi$ is the target policy. ACER also uses a behavior policy $\mu$. Generally, the actions made by behavior policy is used to update the target policy. ACER estimates $Q^{\pi}(s_u,a_u)$ using Retrace \citep{Munos2016_Retrace}. Given a trajectory generated under the behavior policy $\mu$, the Retrace estimator can be written in a recursive manner:
\begin{equation}
    Q^{ret}(s_u,a_u) = r_u + \gamma \Bar{\rho}_{u+1}\left[ Q^{ret}(s_{u+1},a_{u+1})-Q(s_{u+1},a_{u+1})\right] + \gamma V(s_{u+1}),
    \label{eq:retrace-Q-estimation}
\end{equation}
where $\rho _u=\frac{\pi(a_u|s_u)}{\mu(a_u|s_u)}$ denotes the importance weight, $\Bar{\rho}_u=\min(c,\rho_u)$ is the truncated importance weight, $c$ is a constant, $Q$ is the current value estimate of $Q^{\pi}$, and $V(s)=E_{a\sim\pi}(Q(s,a))$. To compute $Q$, we adopt the proposed actor networks in Section 3.3 with ``two heads" that output the estimate $Q_{\theta}(s_u,a_u)$ as well as the policy $\pi_\theta(a_u|s_u)$. The estimate $V_{\theta}(s_u)$ can be obtained by taking the expectation of $Q_{\theta}(s_u,a_u)$ under $\pi_\theta$.

Given the sampling trajectories $\{s_0,a_0,r_0,\mu(\cdot|s_0),\dots,s_u,a_u,r_u,\mu(\cdot|s_u)\}$ generated from the behavior policy ($\mu(\cdot|s_i)$ are the policy vectors, $i\in {1,2,\dots,u}$), the actor network can be updated using the off-policy ACER gradient:
\begin{equation}
\begin{aligned}
    {\Hat{g_u}}^{acer} = &\Bar{\rho}_{u} \nabla _\theta \log \pi _\theta (a_u|s_u)[Q^{ret}(s_u,a_u)-V_{\theta}(s_u)] \\
    &+ E_{a\sim\pi}\left([\frac{\rho _u(a)-c}{\rho _u(a)}]_{+}\nabla _\theta \log \pi _\theta (a|s_u)[Q_\theta (s_u,a)-V_\theta(s_u)]\right), 
\end{aligned}
    \label{eq:ACER-gradient}
\end{equation}
where ${\Hat{g_u}}^{acer}$ is the ACER gradient, $[x]_{+}=x$ if $x>0$ and  $[x]_{+}=0$ otherwise.

The critic network uses a mean squared error loss and updates its parameters by the standard gradient:
\begin{equation}
    {\Hat{g_u}}^{critic} = \left(Q^{ret}(s_u,a_u)-Q_\theta(s_u,a_u)\right)\nabla_\theta Q_\theta(s_u,a_u).
    \label{eq:critic-gradient}
\end{equation}

For more details about ACER algorithm, please refer to \cite{wang2017sample}.

\clearpage

\section{Operation of RL in IAAS Framework}\label{appx-algorithms}

The heuristic screening algorithm in Algorithm \ref{alg:pool-management} involves three other algorithms related to RL. In the following, we first present the Algorithm \ref{alg:pool-management} to show how RL is involved in IAAS and then introduce the operation of RL in the other three algorithms. In particular, Algorithm \ref{alg:sample-new-network} details the process of sampling a new network architecture by selector actor, wider actor, and deeper actor. It also adds a step information to the trajectory. Note that the reward signal is initiated to zero since we can only evaluate the reward signal after some training in the new network structure. Subsequently, the succeeding Algorithm \ref{alg:record-reward} supplements the vacancy left in the Algorithm \ref{alg:sample-new-network}. After one episode of network search, we update the three RL agents using Algorithm \ref{alg:update-rl}. We construct a trajectory training set $T_{train}$ with maximum capacity $N_{Tmax}$ ($N_{Tmax}$ = 100 in our implementation). After that, we update the agents by the training set using the ACER algorithm \citep{wang2017sample} described in Appendix \ref{ACER}.

{
\begin{algorithm}[h]
\caption{Heuristic Screening Algorithm}
\label{alg:pool-management}
\begin{algorithmic}[1]
\footnotesize
\ENSURE{Network pool set $Q$; Net pool capacity $C_Q$; Number of networks in net pool $S_{Q}$; Number of random generated structures $M$; Maximum number of search episodes $N_{max}$; Historical RL trajectories $T$; RL trajectories of networks in pool $T_p$.} 
\STATE Initialize network pool set $Q$;
\STATE Train and test networks in $Q$;
\WHILE{stopping criteria are not met (total episodes$<N_{max}$ in our case)}
\FOR{each network $q_i \in Q$}
    \STATE  ${q_i}'$ = SAMPLE\_NEW\_NETWORK($q_i$,$T$,$T_p$);
    \STATE Add ${q_i}'$ to $Q$;
\ENDFOR
\STATE Collect the maximum layer number $\widetilde{A}$ and maximum unit number in one layer $\widetilde{B}$ from $Q$;
\STATE Generate $M$ network(s) constrained by $\widetilde{A}$ and $\widetilde{B}$, then add to $Q$;
\STATE Train networks in $Q$;
\FOR{each network $q_i \in Q$}
    \STATE $\widetilde{M_i}$ = RMSE evaluation result of $q_i$
    \STATE RECORD\_REWARD($T_p, \widetilde{M_i}, q_i$)
\ENDFOR
\STATE Sort networks in $Q$ by their performance;
\IF{$S_{Q} > C_Q$}
 \STATE Update $Q$ by dropping the last $S_{Q} - C_Q$ networks ;
 \STATE Move trajectories of dropped networks from $T_p$ to $T$
 \ENDIF
\STATE UPDATE\_RL($T$,$T_p$);
\ENDWHILE
\STATE Identify the best network $q^{*}$ in $Q$;
\lastcon{Search result best network $q^{*}$.}
\end{algorithmic}
\end{algorithm}

\begin{algorithm}[h]
\caption{SAMPLE\_NEW\_NETWORKS}
\label{alg:sample-new-network}
\begin{algorithmic}[1]
\footnotesize
\ENSURE{Original network structure $q_i$; RL trajectories of networks in the netpool $T_p$; Selector Actor $A_s$; Wider Actor $A_w$; Deeper Actor $A_d$.} 
\STATE Get structure state $s_{q_i}$ from $q_i$;
\STATE Get decision polices $\mu_{q_i,s}(\cdot|s_{q_i})=A_s(s_{q_i}), \mu_{q_i,w}(\cdot|s_{q_i})=A_w(s_{q_i}), \mu_{q_i,d}(\cdot|s_{q_i})=A_d(s_{q_i})$
\STATE Sample actions $a_{q_i,s}$ from polices $\mu_{q_i,s}(\cdot|s_{q_i})$
\IF{$a_{q_i,s}$ is unchanged}
\STATE $q_i \xrightarrow{ } {q_i}'$ ;
\ELSIF{$a_{q_i,s}$ is prune}
\STATE Pruning ${q_i}$ with the movement pruning algorithm $\xrightarrow{ } {q_i}'$;
\ELSIF{$a_{q_i,s}$ is wider}
\STATE Sample actions $a_{q_i,w}$ from polices $\mu_{q_i,w}(\cdot|s_{q_i})$

\STATE Widening ${q_i}$ according to $a_{q_i,w} \xrightarrow{ } {q_i}'$;
\ELSE
\STATE Sample actions $a_{q_i,d}$ from polices $\mu_{q_i,d}(\cdot|s_{q_i})$
\STATE Deepening ${q_i}$ according to $a_{q_i,d} \xrightarrow{ } {q_i}'$;
\ENDIF
\STATE Set reward $r=0$ as a placeholder for future update;
\STATE Record actions $a_{q_i} = (a_{q_i,s},a_{q_i,w},a_{q_i,d})$;
\STATE Record behavior polices $\mu_{q_i}(\cdot|s_{q_i}) = (\mu_{q_i,s}(\cdot|s_{q_i}), \mu_{q_i,w}(\cdot|s_{q_i}), \mu_{q_i,d}(\cdot|s_{q_i}))$;
\STATE Add new RL record$\{s_{q_i},a_{q_i},r,\mu_{q_i}(\cdot|s_{q_i})\}$ to $T_p$

\lastcon{Transformation result ${q_i}'$.}
\end{algorithmic}
\end{algorithm}

\begin{algorithm}[h]
\caption{RECORD\_REWARD}
\label{alg:record-reward}
\begin{algorithmic}[1]
\footnotesize
\ENSURE{Original network structure $q_i$; RMSE evaluation result $\widetilde{M_i}$; RL trajectories of networks in netpool $T_p$;.} 
\STATE Find unfinished record $\{s_{q_i},a_{q_i},r,\mu_{q_i}(\cdot|s_{q_i})\}$ of $q_i$ from $T_p$;
\STATE $r_{q_i} = \frac{1}{\widetilde{M_i}}$
\STATE Update record in $T_p$ by $\{s_{q_i},a_{q_i},r_{q_i},\mu_{q_i}(\cdot|s_{q_i})\}$
\end{algorithmic}
\end{algorithm}

\begin{algorithm}[h]
\caption{RL\_UPDATE}
\label{alg:update-rl}
\begin{algorithmic}[1]
\footnotesize
\ENSURE{Historical RL trajectories $T$; RL trajectories of networks in the net pool $T_p$; Selector Actor $A_s$; Wider Actor $A_w$; Deeper Actor $A_d$; Parameters $\theta_s, \theta_w, \theta_d$ of $A_s, A_w$ and $A_d$; Max training trajectory number $N_{Tmax}$.}
\STATE $\theta = \{\theta_s, \theta_w, \theta_d\}$ 
\STATE $T_{train} = T_p$
\IF{$size(T_{train})<N_{Tmax}$}
\STATE Sample $\max(N_{Tmax}-size(T_{train}),size(T))$ trajectories and add it to $T_{train}$
\ENDIF
\FOR{Trajectory $t \in T_{train}$}
    \STATE  Calculate ACER gradient ${\Hat{g_t}}^{acer}$ and Critic gradient ${\Hat{g_u}}^{critic}$ using Eq. (\ref{eq:ACER-gradient}) and Eq. (\ref{eq:critic-gradient});
    \STATE  Update $\theta$ by ACER gradient ${\Hat{g_t}}^{acer}$ and Critic gradient ${\Hat{g_u}}^{critic}$;
\ENDFOR
\end{algorithmic}
\end{algorithm}
}

\newpage
\section{Detailed Data Description}\label{appx:Data details}

In each electricity load subset, the time step of load data is set to an hour, thereby leading to 24 data points per day, and thus approximately 2160 data points per season. Following \cite{jalali2021novel}, we split each independent subset into a training dataset with 75\% data points and a test dataset with 25\% data points. Moreover, the data in the training dataset is chronically earlier than that in the test dataset. When forecasting the data points in the load demand subsets, apart from using historical load demand data, we incorporate their corresponding time information as inputs, namely $\widetilde{h}^{th}$ hour of a day and $\widetilde{w}^{th}$ day of a week, where $\widetilde{h}$ and $\widetilde{w}$ are restricted to ranges of $[1,24]$ and $[1,7]$, respectively. For instance, if the load demand is 50 MW at 8:00 am on a Sunday, we describe such an input vector as $(50, 8, 7)'$. To improve forecasting accuracy, we use one-week data with the size of $168\times 3$ as one input for the forecasting model to predict the 24h-ahead load demand point. 

In each wind power subset, the time step of wind power data is set as an hour. Similar to the settings in \cite{zhang2020short}, the last five-day data of each season are treated as the test dataset, and the remaining data of that season are the training dataset. As for the forecast of data points in the wind power subsets, numerical weather prediction (NWP) data are widely utilized following the day-ahead wind power literature (see \citealt{zhang2020short,chen2023novel}). We can notice that the NWP data contains five dimensions, namely, humidity, wind speed, wind direction, temperature and air pressure, which are weather simulation data generated from NWP computer models of the atmosphere and oceans \citep{rabier2005overview}. Consistent with \cite{zhang2020short}, we use the sine and cosine values of wind direction to signify the real wind direction. Moreover, we use the historical wind speed, wind power data, and hourly timestamp information in a day as additional input variables. Therefore, we have an input vector with nine dimensions. We use three-day data with a size of ${72}\times {9}$ as one input for the forecasting model to predict the 24h-ahead wind power point. Unlike the load input data, NWP data at the forecasting point is known in advance. For instance, if we are at 8 am on Jun 9 and forecast the wind power generation at 8 am on Jun 10, we can use the NWP data from 9 am on Jun 8 to 8 am on Jun 10 as input. 

\begin{figure}[h]
    \begin{subfigure}[b]{0.5\textwidth}
        \centering
        \includegraphics[width=\textwidth]{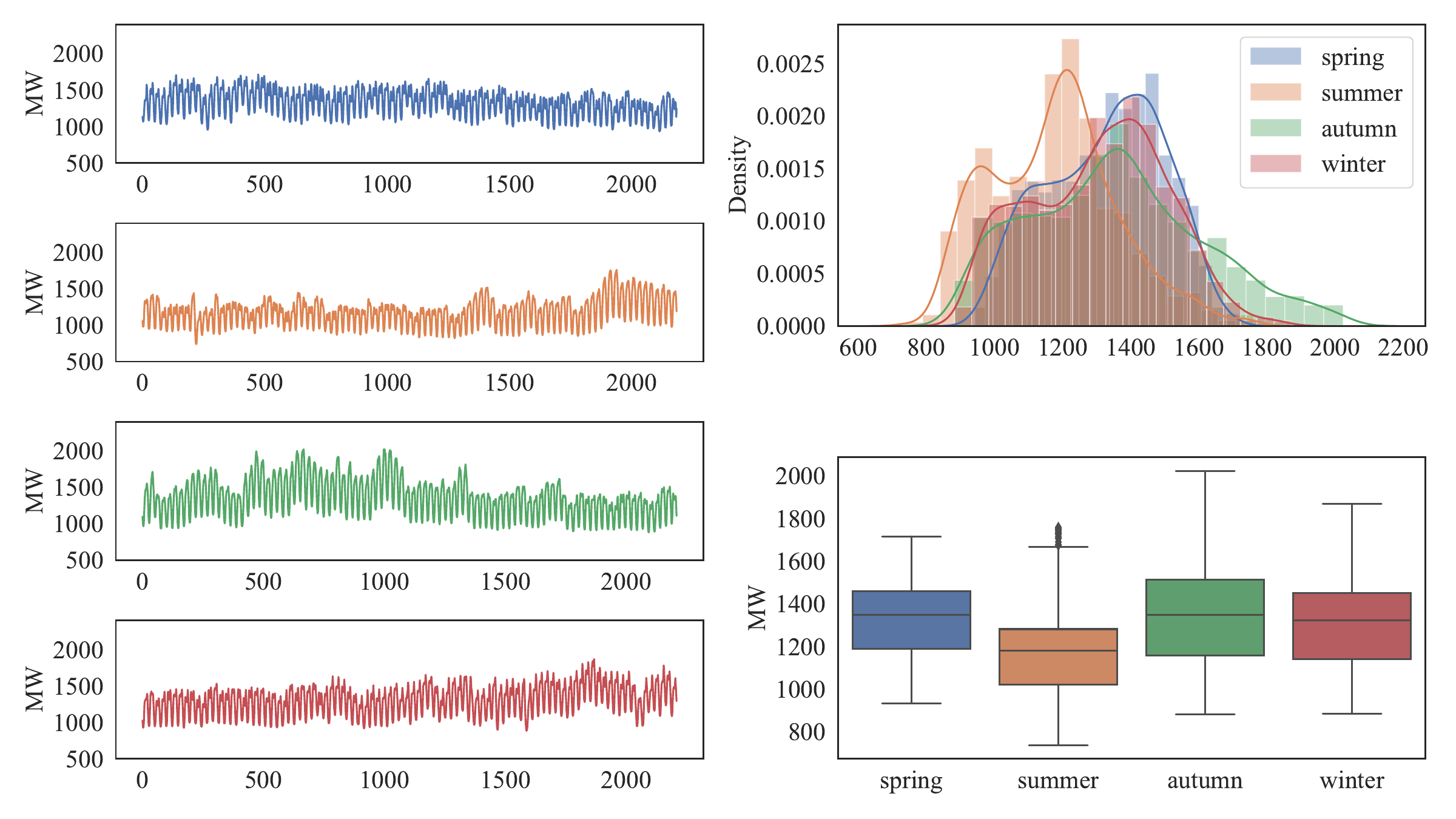 }
        \caption*{\footnotesize{(a) ME load dataset}}
    \end{subfigure}
    \hfill
    \begin{subfigure}[b]{0.5\textwidth}  
        \centering
        \includegraphics[width=\textwidth]{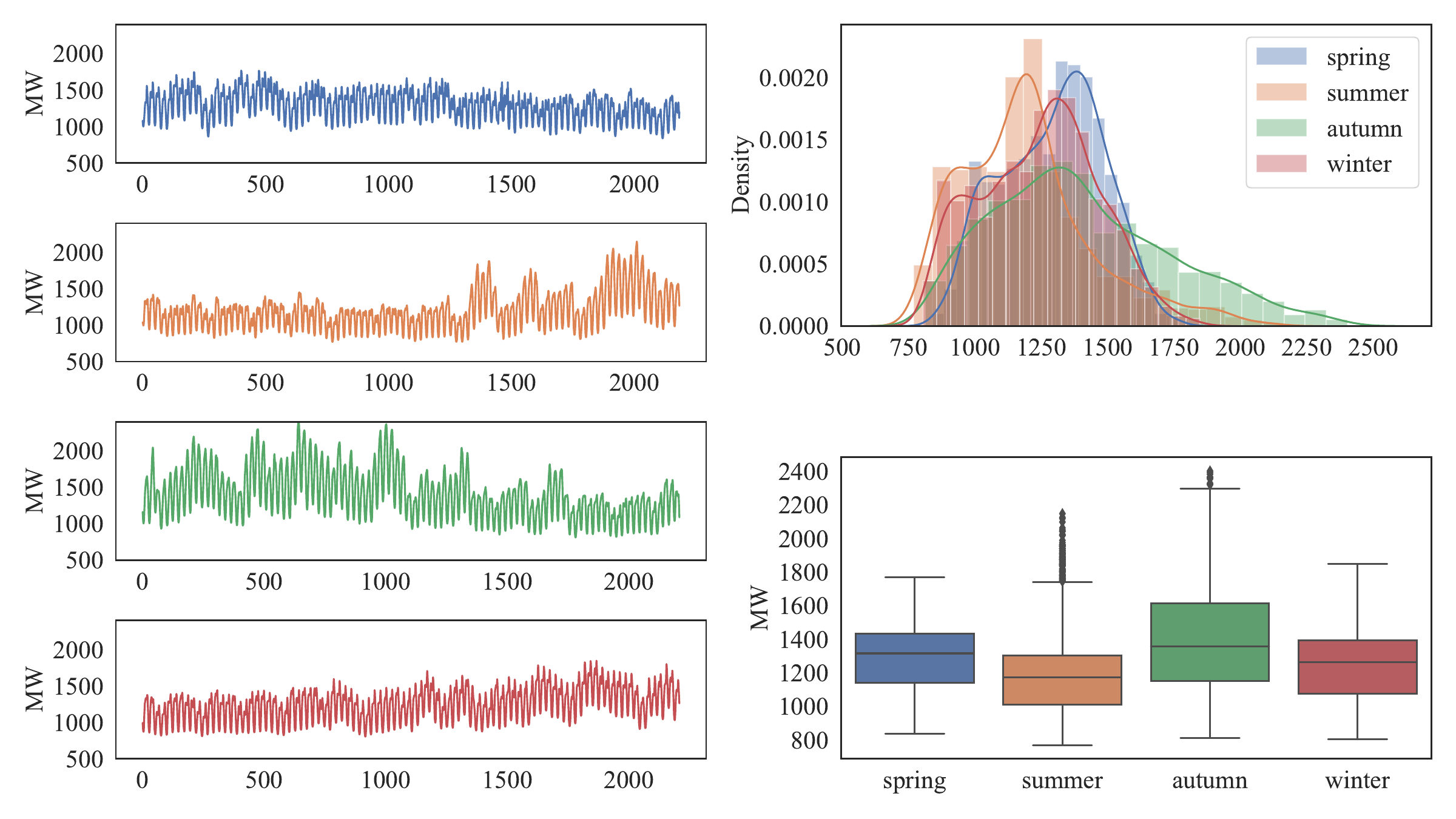}
        \caption*{\footnotesize{(b) NH load dataset}}
    \end{subfigure}
    \vskip\baselineskip
    \begin{subfigure}[b]{0.5\textwidth}   
        \centering
        \includegraphics[width=\textwidth]{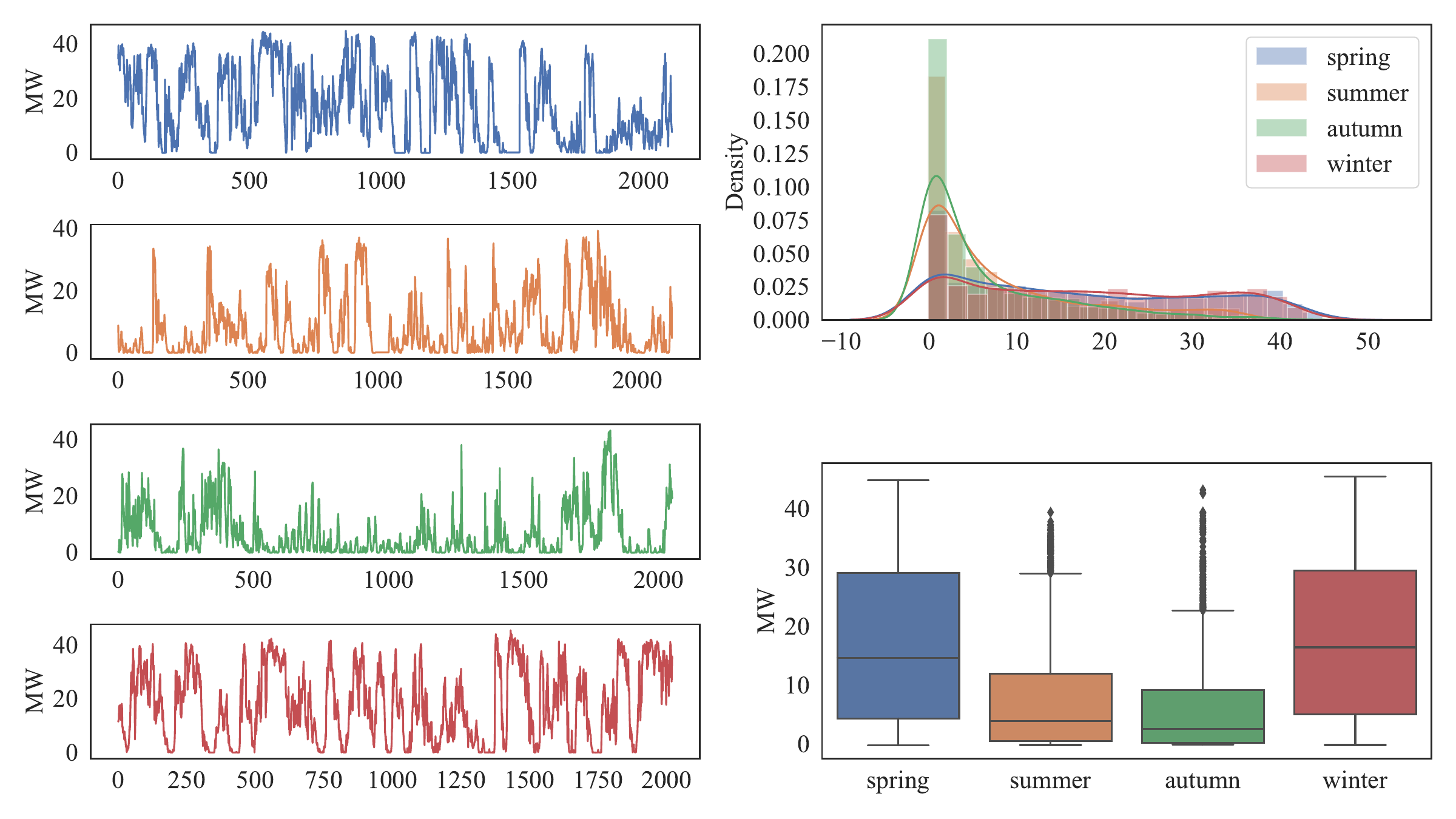}
        \caption*{\footnotesize{(c) WF1 wind power dataset}}
    \end{subfigure}
    \hfill
    \begin{subfigure}[b]{0.5\textwidth}   
        \centering
        \includegraphics[width=\textwidth]{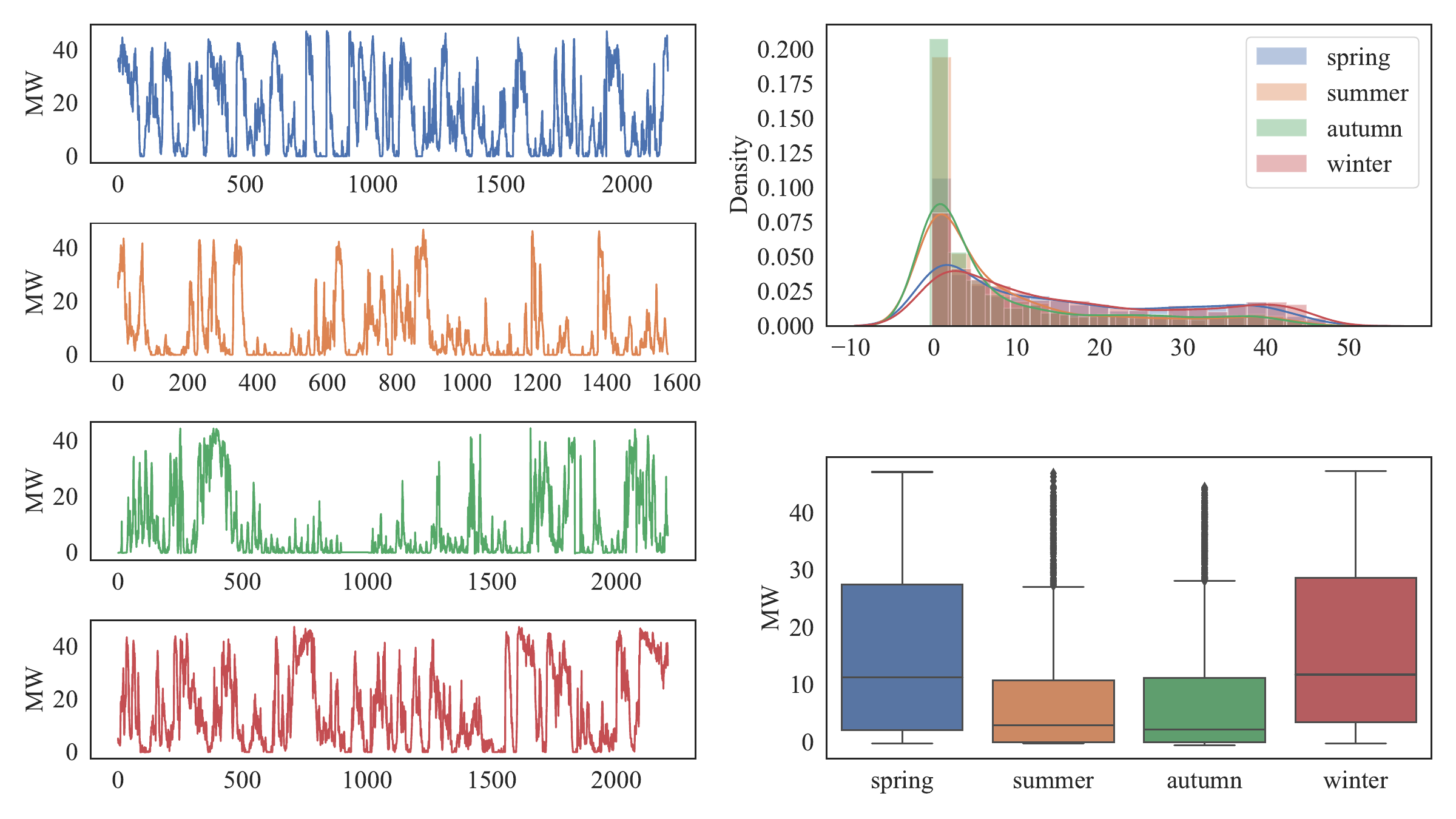}
        \caption*{\footnotesize{(d) WF2 wind power dataset}}
    \end{subfigure}
    \caption{
    Data description of four cases \label{fig:data-description}
    }
\end{figure}

We plot the data value, empirical density, and boxplot by seasons in each sub-figure of Fig. \ref{fig:data-description}. We can observe that the load data is quite different from wind power data; however, both the load data and wind power data have seasonal patterns. For instance, the ME-spring load data has similar patterns to the NH-spring load data, and WF1-winter wind power generation data is also akin to WF2-winter wind power generation data. However, despite these similarities, they also have some differences from the further observations. We select these datasets for the numerical experiments for the following reasons. First, two types of electricity data, i.e., load and wind power data, are utilized to demonstrate the generalizations of the proposed IAAS framework. Second, the two identical types of data within some similarities are leveraged to display the stability of this framework. 

\clearpage
\section{Experimental Settings}\label{appx:experimental-setting}

\begin{table}[ht]
\caption
{Experimental settings for the baselines and IAAS framework\label{tab:Experimental-settings}}
{
\begin{tabularx}{\textwidth}{XXX}
\hline
Model & Hyperparameter & Value \\ 
\hline
\multirow{5}{*}{IAAS} & Activation function & ReLU \\
 & Batch Size & 256 \\
 & Optimizer & Adam \\
 & Epoch number of   each episode & 50 \\
 & Max. search   episodes & 200 \\ 
\hline
\multirow{7}{*}{SNAS} & Activation function & ReLU \\
 & Batch size & {[}64-1024], step size 64 \\
 & Hidden channels & 32 \\
 & Layer per cell & {[}1,2,3,4] \\
 & Search epoch & {[}50-100] step size 10 \\
 & Search   iteration per epoch & {[}64-256] step size 64 \\
 & Training epoch & 500 \\ 
\hline
\multirow{4}{*}{DAIN} & Activation   function & ReLU \\
 & Mean learning   rate & {[}1e-5,1e-4,1e-3,1e-2] \\
 & Gate learning   rate & {[}1e-5,1e-4,1e-3,1e-2] \\
 & Scale learning   rate & {[}1e-5,1e-4,1e-3,1e-2] \\ 
\hline
\multirow{6}{*}{\begin{tabular}[c]{@{}l@{}}CNN+LSTM\\CNN\\LSTM \\ResNet\\ResNetPlus\end{tabular}} & Size of each layer & {[}4-128], step size 4 \\
 & Number of layers & {[}2-10], step size 1 \\
 & Activation function & ReLU, SELU, LeakyReLU \\
 & Batch size & {[}64-1024], step size 64 \\
 & Learning rate & {[}0.001-0.1] , step size 0.001 \\
 & Dropout rate & {[}0.2-0.8] , step size 0.1 \\ 
\hline
\multirow{4}{*}{SVR} & Kernel & RBF, Sigmoid~, Poly \\
 & Polynomial order & {[}1-12] , step size 1 \\
 & Regularization & 1,10,100,1000 \\
 & epsilon & {[}0.1-1] , step size 0.1 \\ 
\hline
\multirow{2}{*}{RF~} & Split measure function & squared, absolute, Poisson \\
 & Number of trees & {[}16-128] , step size 4 \\ 
\hline
RR & penalty alpha & {[}0-4] , step size 0.1 \\
\hline
\end{tabularx}
}
{}
\end{table}

The experimental settings, including all optimized hyperparameters for the baseline models (or methods) and our proposed IAAS framework, are listed in Table \ref{tab:Experimental-settings}. It should be mentioned that for the CNN, LSTM, and CNN+LSTM, we use the grid search algorithm to optimize the number of layers, size of each layer, activation functions, batch size, learning rate (LR), and dropout rate. Note that we make use of three activation functions, namely ReLU, scaled exponential linear units (SELU), and LeakyReLU, as the optimization candidates. For ResNet and ResNetPlus, we first examine their performance based on the reference article. Then, to obtain their best performances, we also utilize the grid search algorithm to optimize their hyperparameters, such as the number of layers and LR. For DAIN, we use the multilayer perceptron (MLP) to build the forecasting model since MLP has achieved better forecasting accuracy than RNN in \cite{passalis2019deep}. We follow the hyperparameter settings from \cite{passalis2019deep} to conduct the experiments. And then, we make use of the trial-and-error simulation method to optimize the mean LR, gate LR, and scale LR to achieve a better result. As to the SNAS, we follow \cite{chen2021scale} to set the hyperparameters since it is a NAS method to develop the time-series model, and these hyperparameter settings have helped SNAS obtain a better time-series forecasting accuracy on various datasets. Moreover, we also utilize the trial-and-error simulation method to optimize the hyperparameters, such as batch size, the number of layers per cell, the number of search iterations per epoch, and the number of search epochs. Note that since both \cite{passalis2019deep} and \cite{chen2021scale} only leveraged the ReLU activation function, we follow this setting in these two baseline models. In terms of SVR, RF, and RR, we use the grid search algorithm to optimize the hyperparameters as shown in Table \ref{tab:Experimental-settings} as well. 

\newpage
\section{Sensitivity Analysis Regarding the Number of Search Episodes}\label{sensitivity}
Considering the importance of the number of search episodes, we perform a sensitivity analysis to investigate its effect on model performance in this section. Specifically, we use five different numbers of episodes, namely 40, 80, 120, 160, 200, 240 and 280 for the experiments with the usage of ME load and WF1 datasets. We present the results in Table \ref{tab:Episode-performance} and bold the first best value from left to right in each row. Observing all eight cases, the RMSE values with more episodes are never larger than those with fewer episodes. This is because of the heuristic screening algorithm, which always preserves the network structures with good performance and kick off the bad ones in the net pool. The RMSLE values in all four load cases have the same changing pattern as their corresponding RMSE values. As to the MAE values, they have some differences. For example, in the case of ME-spring, when using 40 episodes and 160 episodes, the MAE values are 48.366 and 48.879, respectively. This means that the increase of the number of search episodes leads to the increase of the MAE value. The reason for this is that we use the RMSE as the loss function, which directly reflects the change in the RMSE forecasting accuracy but not the MAE. Note that the calculation of MAE is different from that of RMSE; hence, we regard this phenomenon as normal. 

 \begin{table}[ht]
\caption
{Sensitivity analysis regarding the number of search episodes\label{tab:Episode-performance}}
{
\resizebox{\linewidth}{!}{

\begin{tabular}{ccccccccc}
\hline
Case & Metrics & 40 episodes & 80 episodes & 120 episodes & 160 episodes & 200 episodes & 240 episodes & 280 episodes \\ \cline{1-9}
\multirow{3}{*}{ME-spring} & RMSE & 61.978 & 61.978 & 61.978 & \textbf{61.807} & 61.807 & 61.807 & 61.807 \\ \cline{2-9}
 & MAE & \textbf{48.366} & 48.366 & 48.366 & 48.879 & 48.879 & 48.879 & 48.879 \\ \cline{2-9}
 & RMSLE & 0.05 & 0.05 & 0.05 & \textbf{0.049} & 0.049 & 0.049 & 0.049 \\ \cline{1-9}
\multirow{3}{*}{ME-summer} & RMSE & 73.258 & 73.258 & \textbf{69.154} & 69.154 & 69.154 & 69.154 & 69.154 \\ \cline{2-9}
 & MAE & \textbf{57.29} & 57.29 & 58.187 & 58.187 & 58.187 & 58.187 & 58.187 \\ \cline{2-9}
 & RMSLE & 0.054 & 0.054 & \textbf{0.052} & 0.052 & 0.052 & 0.052 & 0.052 \\ \cline{1-9}
\multirow{3}{*}{ME-autumn} & RMSE & \textbf{38.483} & 38.483 & 38.483 & 38.483 & 38.483 & 38.483 & 38.483 \\ \cline{2-9}
 & MAE & \textbf{30.526} & 30.526 & 30.526 & 30.526 & 30.526 & 30.526 & 30.526 \\ \cline{2-9}
 & RMSLE & \textbf{0.031} & 0.031 & 0.031 & 0.031 & 0.031 & 0.031 & 0.031 \\ \cline{1-9}
\multirow{3}{*}{ME-winter} & RMSE & 85.92 & \textbf{85.386} & 85.386 & 85.386 & 85.386 & 85.386 & 85.386 \\ \cline{2-9}
 & MAE & \textbf{66.123} & 68.773 & 68.773 & 68.773 & 68.773 & 68.773 & 68.773 \\ \cline{2-9}
 & RMSLE & 0.064 & \textbf{0.062} & 0.062 & 0.062 & 0.062 & 0.062 & 0.062 \\ \cline{1-9}
\multirow{2}{*}{WF1-spring} & RMSE & 3.982 & 3.982 & 3.982 & \textbf{3.901} & 3.901 & 3.901 & 3.901 \\ \cline{2-9}
 & MAE & 3.054 & 3.054 & 3.054 & \textbf{2.996} & 2.996 & 2.996 & 2.996 \\ \cline{1-9}
\multirow{2}{*}{WF1-summer} & RMSE & 3.375 & 3.375 & 3.375 & 3.348 & \textbf{3.29} & 3.29 & 3.29 \\ \cline{2-9}
 & MAE & \textbf{2.31} & 2.31 & 2.31 & 2.474 & 2.614 & 2.614 & 2.614 \\ \cline{1-9}
\multirow{2}{*}{WF1-autumn} & RMSE & 2.587 & 2.498 & 2.498 & 2.498 & \textbf{2.237} & 2.237 & 2.237 \\ \cline{2-9}
 & MAE & 1.769 & 1.91 & 1.91 & 1.91 & \textbf{1.633} & 1.633 & 1.633 \\ \cline{1-9}
\multirow{2}{*}{WF1-winter} & RMSE & \textbf{4.5} & 4.5 & 4.5 & 4.5 & 4.5 & 4.5 & 4.5 \\ \cline{2-9}
 & MAE & \textbf{3.427} & 3.427 & 3.427 & 3.427 & 3.427 & 3.427 & 3.427 \\ \hline
\end{tabular}
}}
{}
\end{table}

All in all, we have the following five findings: (i) when using 200 episodes, we can obtain the best RMSE accuracy in all cases and best RMSLE accuracy in all load cases; (ii) in three out of eight cases, we obtain the best MAE accuracy when using 200 episodes; (iii) the forecasting accuracies using 240 episodes and 280 episodes are the same as those using 200 episodes; (iv) in most cases, the forecasting accuracies between using 200 episodes and others do not show large differences; (v) in the case of WF1-autumn, the best forecasting accuracy is from using 200 episodes and is better than that from using 160 episodes with improvement of 10.4\% (RMSE) and 14.5\% (MAE). Therefore, we can conclude that the determination of using 200 episodes in this study is reasonable. Others may consider using 40 episodes if they pay more attention to the computational time.


\newpage

\section{Efficiency Performance Examination of the Selected RL Algorithm}\label{time analysis}

\begin{figure}[htb]
    \begin{subfigure}[b]{0.515\textwidth}
        \centering
	    \includegraphics[width=\textwidth]{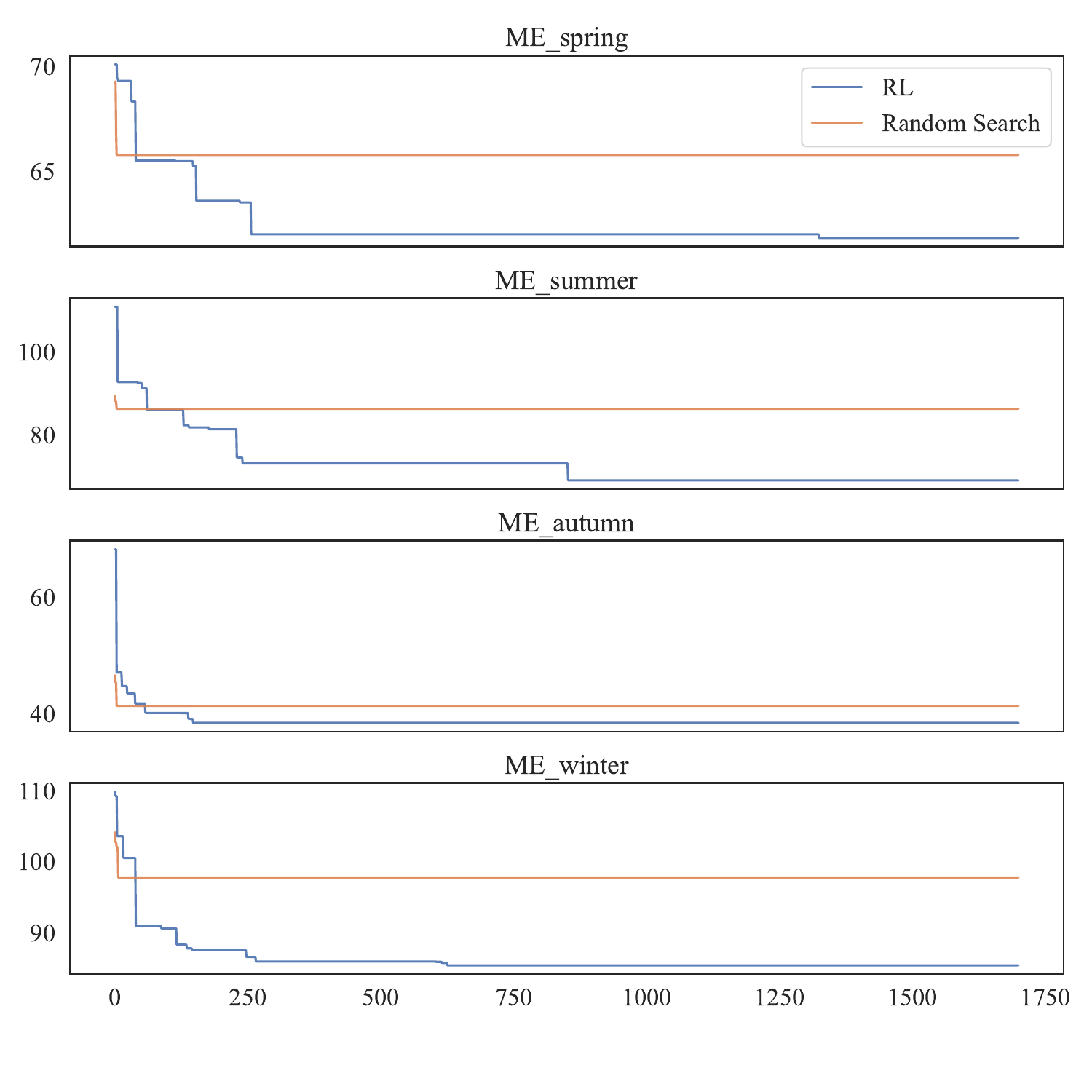}
        \caption*{\footnotesize{(a) Load case}}
    \end{subfigure}
    \hfill
    \begin{subfigure}[b]{0.515\textwidth}
        \centering
		\includegraphics[width=\textwidth]{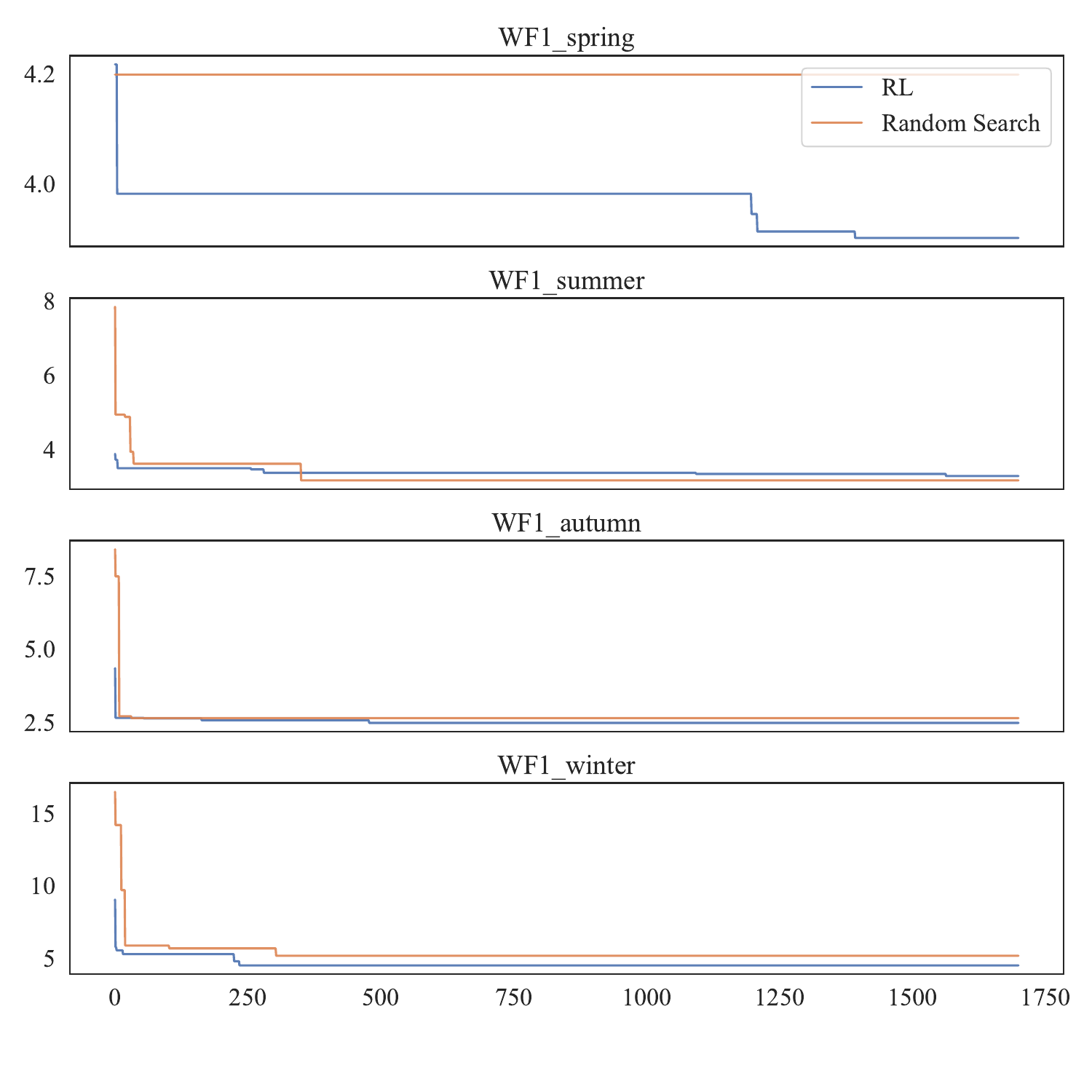}
        \caption*{\footnotesize{(b) Wind power case}}
    \end{subfigure}
\caption{Top RMSE forecasting accuracy in the search process.
    \label{fig:accuracy-compare}}
\end{figure}

In this study, we utilize RL to implement the intelligent control for the network transformation. To demonstrate the efficiency of using RL in the IAAS framework, we follow \cite{cai2018efficient} and make use of the random search algorithm as a benchmark. We randomly select the ME load dataset and WF1 dataset from the four datasets to demonstrate the forecasting performance of the selected RL and random search algorithms. For the random search operation, we use the same pool size and randomly generate network structures at each search episode instead of using RL-guided network transformation operation. To manage the networks in the pool, we utilize the proposed heuristic screening algorithm so that we can iterativley search a neural architecture with good performance after completing total search episodes. In this experiment, we set each algorithm to iteratively select 600 network structures. Fig. \ref{fig:accuracy-compare} shows the change of the best forecasting performance along with the increase number of the searched network structures. As seen, after searching a certain number of network structures, both of these two algorithms can achieve the best network structure among the 600 ones in each case. However, the RL algorithm can efficiently find the network structure with better performance in seven out of eight cases. Although the network performance from RL is lower than that from random search in WF1\_autumn case, their performances are very close as shown in Fig. \ref{fig:accuracy-compare}(b).

\begin{figure}[htb]
    \begin{subfigure}[b]{0.515\textwidth}
        \centering
	    \includegraphics[width=\textwidth]{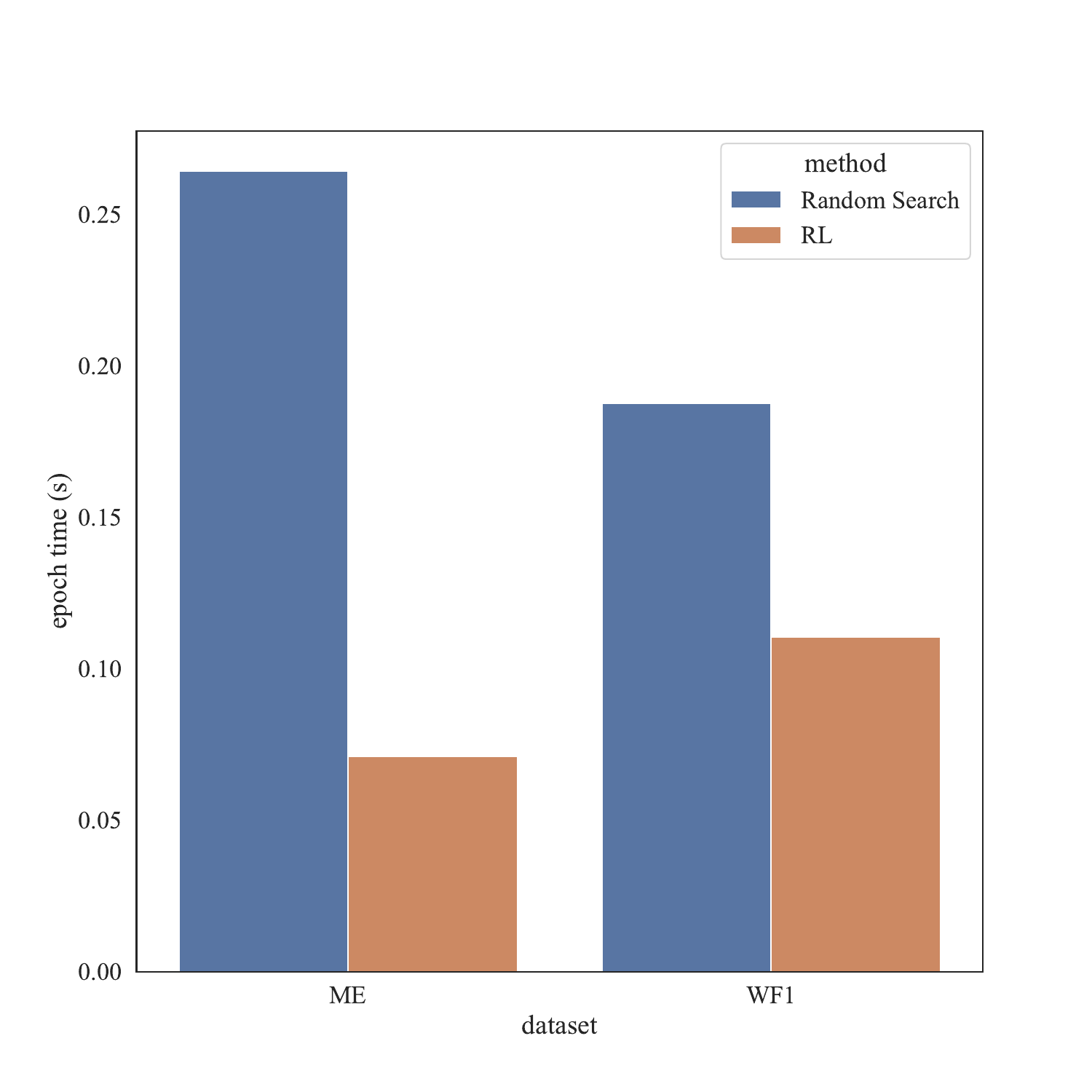}
        \caption*{\footnotesize{(a)}}
    \end{subfigure}
    \hfill
    \begin{subfigure}[b]{0.515\textwidth}
        \centering
		\includegraphics[width=\textwidth]{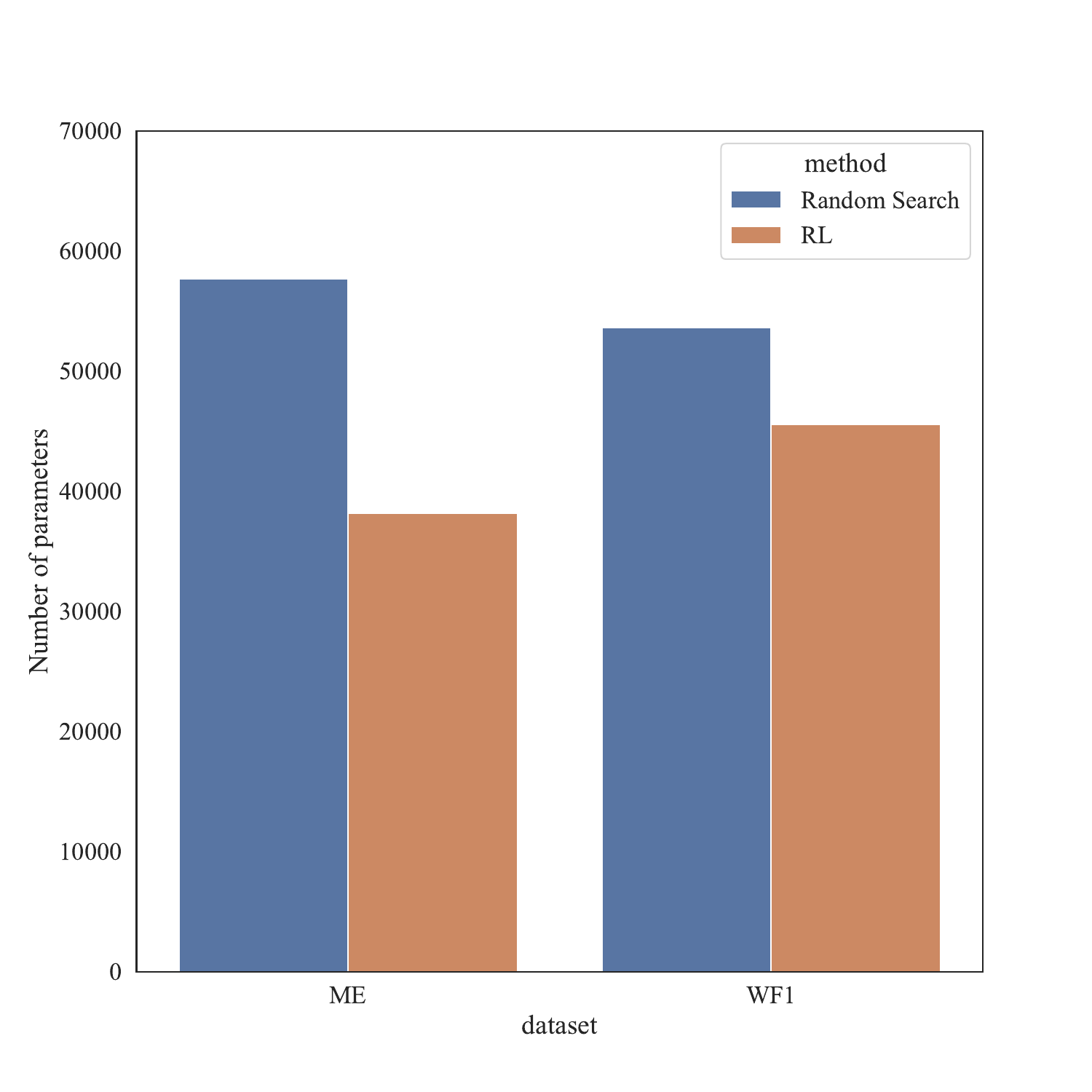}
        \caption*{\footnotesize{(b)}}
    \end{subfigure}
\caption{Performance comparison between the selected RL and random search algorithm: (a) average training time per epoch and (b) average parameter number of each searched model.
    \label{fig:analysis-compare}}
\end{figure}

Moreover, we use the average training time of the generated networks over episodes, and the average number of parameters over the generated networks to showcase the performance of these two algorithms. As seen in Fig. \ref{fig:analysis-compare}(a), using RL-guided network transformation reduces the average training time when comparing with random search. As seen in Fig. \ref{fig:analysis-compare}(b), RL tends to generate network with fewer parameters. From these three perspectives, we can conclude the selected RL is more suitable to the IAAS framework than the random search algorithm. 

\clearpage

\section{Seq2Seq Forecasting Accuracy Comparison}\label{TransformerCompare}

In Section 5.1, we segment the dataset quarterly, and use the small portion of the data as the testing dataset. However, in practice, the electricity company always make the forecasting for the next several days at each time step. Therefore, the rolling-window technique is usually adopted for such a Seq2Seq forecasting task. Considering the excellent performance of transformer models in natural language processing area, we conduct a further experiment between the IAAS and the transformer models under the Seq2Seq setting. Specifically, we select two well-known transformer models, Transformer \citep{vaswani2017attention} and Informer \citep{zhou2021informer} as benchmarks. Following the above experiments, we also use RMSE, MAE and RMSLE as error measurements for load forecasting accuracy evaluation, and use RMSE and MAE for wind power forecasting accuracy evaluation. In this experiment, we randomly select ME load dataset and WF1 dataset from Section 4, and conduct one-day ahead forecasting and one-week ahead forecasting, respectively. 

Intuitively, the one-day ahead forecasting experiment is designed to use the hourly historical data in past days to predict the data in the future one day (24 time steps). The input sequence sizes for the load and wind data are $168\times3$ (seven days) and $72\times9$ (three days), respectively. The output sequence sizes for these two cases are both $24\times1$, which means we forecast the load or wind power in the next day. 
For example, on Jan 1, we use the IAAS framework to search a Seq2Seq model to forecast the load or wind power at Jan 2. After that, we use this Seq2Seq model at Jan 2 to forecast them on Jan 3 with the past 7 days' real data including that of Jan 1 as input. Following this way, we will use this model to forecast the load or wind power at Jan 4-6. In total, we make use of this Seq2Seq model five times for one-day ahead forecasting. Considering the correlation of weather in consecutive days, we update this model at Jan 6. We dynamically update the Seq2Seq model every five days, aiming to provide a quality forecasting.

\begin{figure}[htb]
    \begin{subfigure}[b]{0.32\textwidth}
        \centering
	    \includegraphics[width=\textwidth]{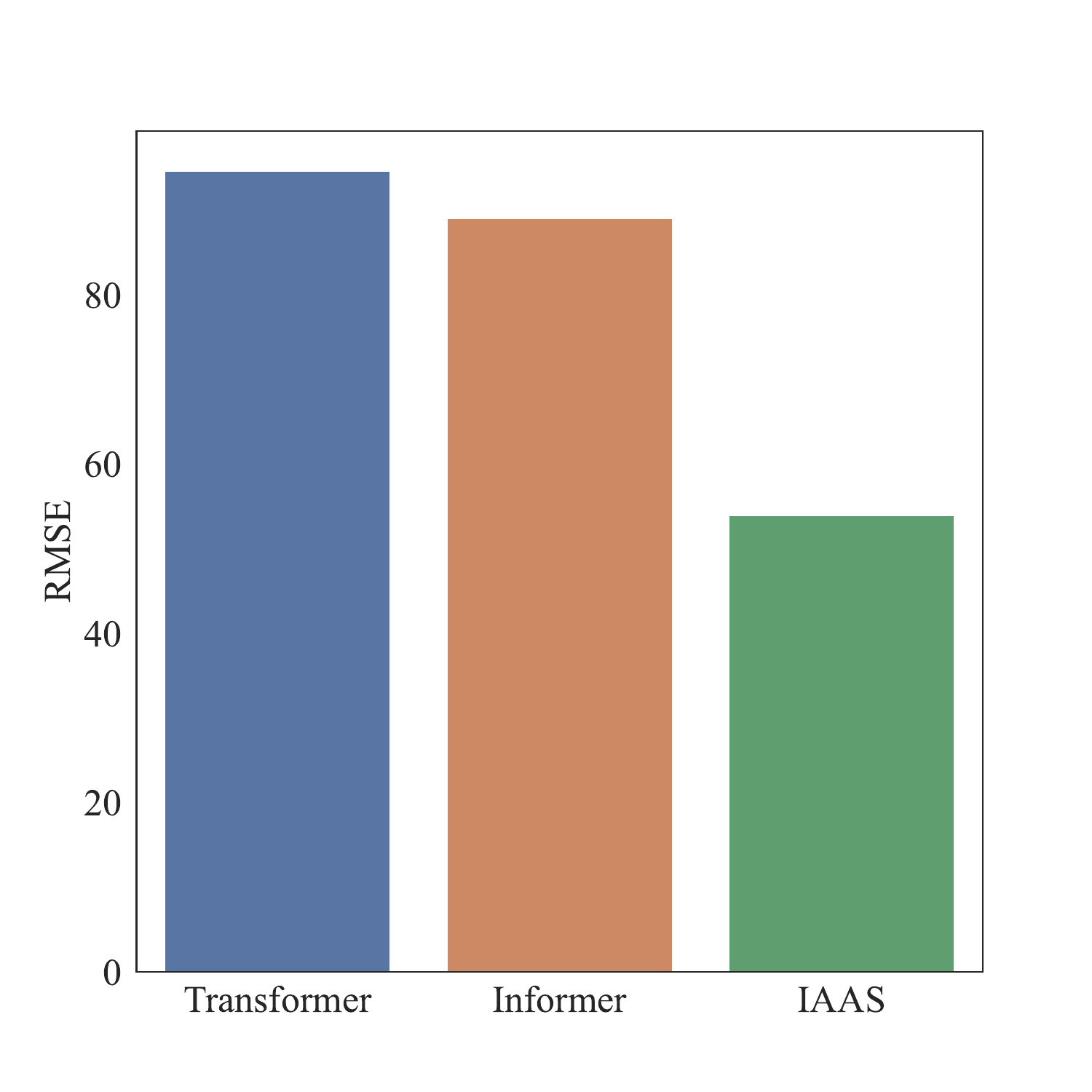}
        \caption*{\footnotesize{(a) RMSE}}
    \end{subfigure}
    \hfill
    \begin{subfigure}[b]{0.32\textwidth}
        \centering
		\includegraphics[width=\textwidth]{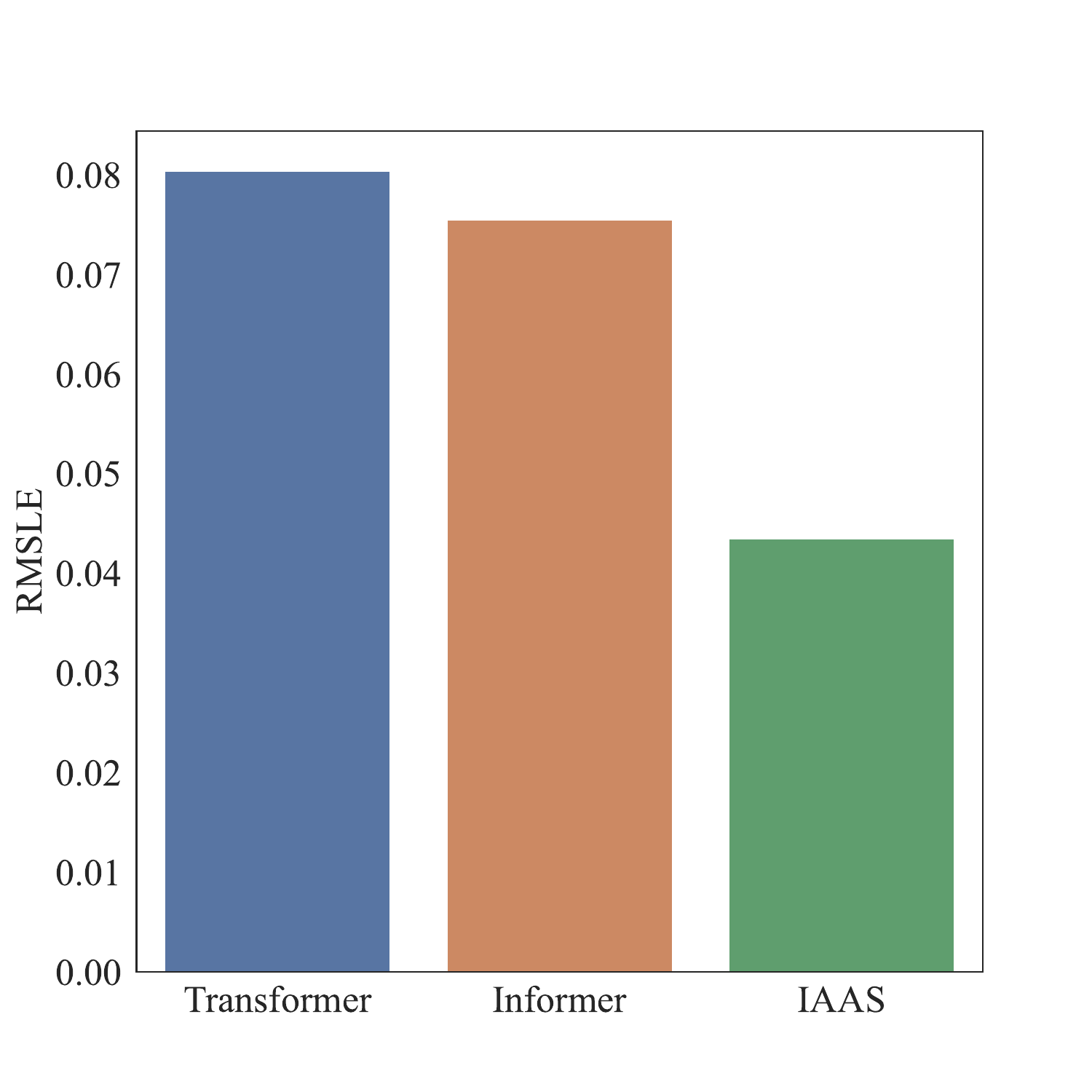}
        \caption*{\footnotesize{(b) RMSLE}}
    \end{subfigure}
    \hfill
    \begin{subfigure}[b]{0.32\textwidth}
        \centering
		\includegraphics[width=\textwidth]{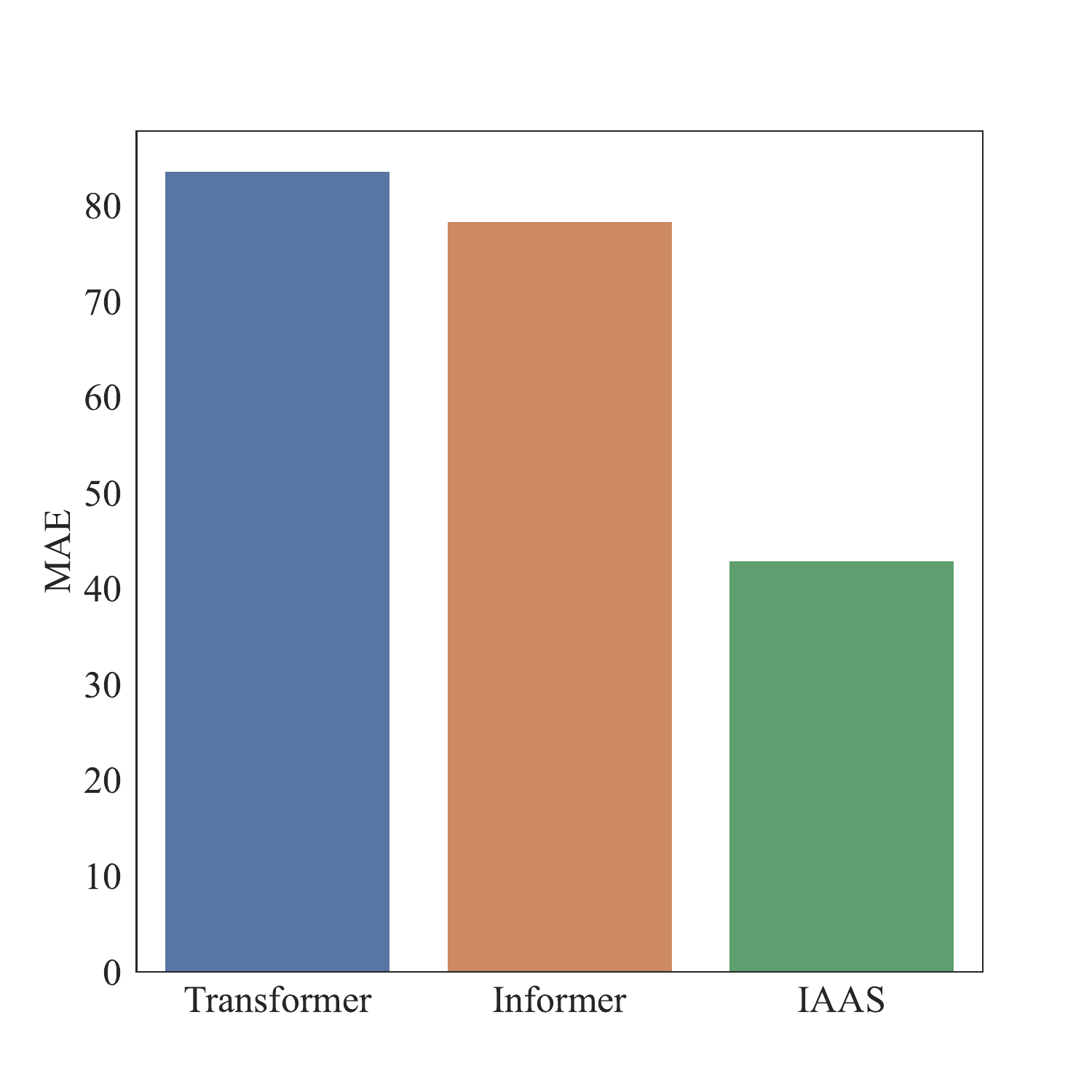}
        \caption*{\footnotesize{(C) MAE}}
    \end{subfigure}
\caption{One-day ahead load forecasting accuracy results under the Seq2Seq setting using Transformer, Informer, and IAAS. 
    \label{fig:seq2seq-ME}}
\end{figure}

\begin{figure}[htb]
    \centering
    \begin{subfigure}[b]{0.32\textwidth}
        \centering
	    \includegraphics[width=\textwidth]{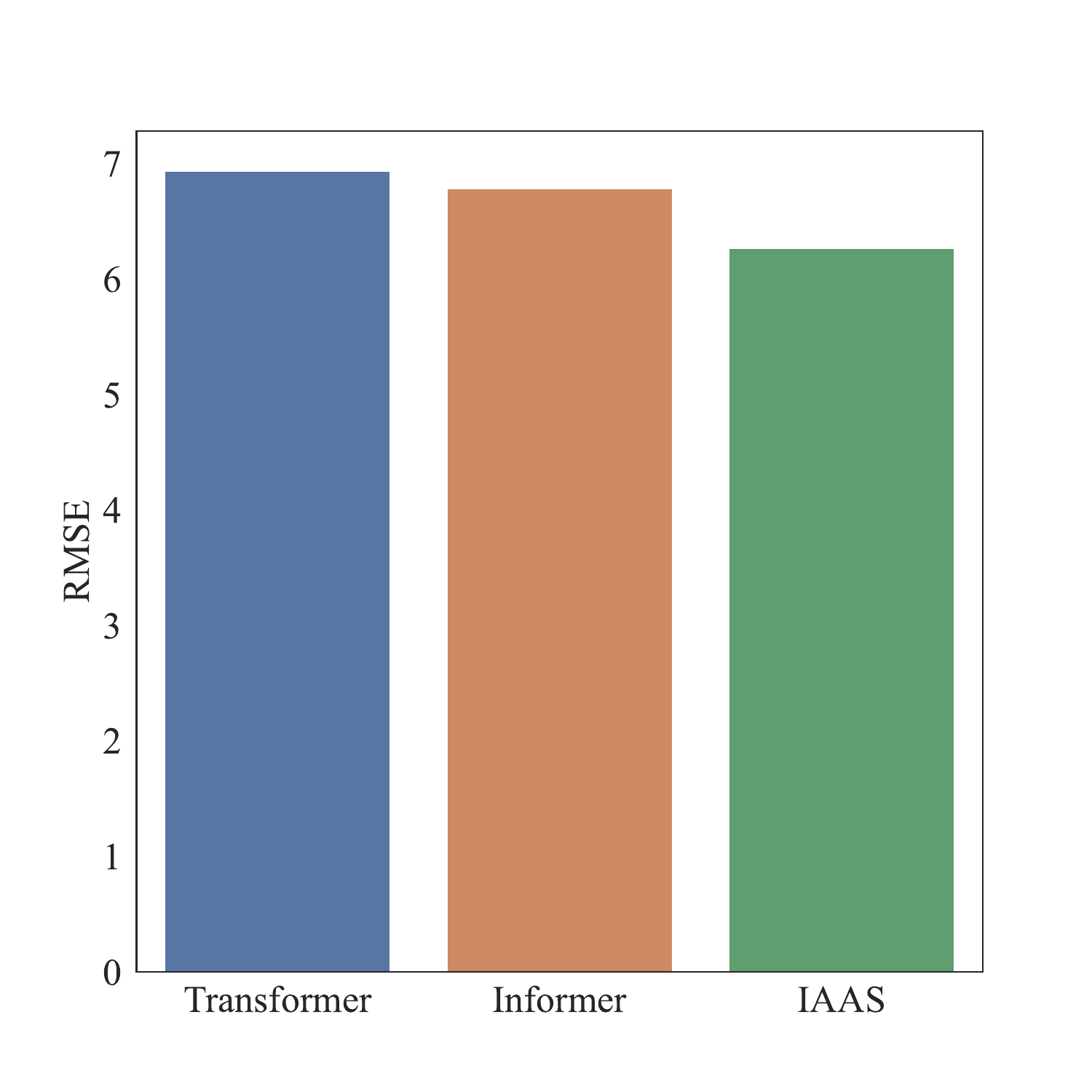}
        \caption*{\footnotesize{(a) RMSE}}
    \end{subfigure}
    \begin{subfigure}[b]{0.32\textwidth}
        \centering
	\includegraphics[width=\textwidth]{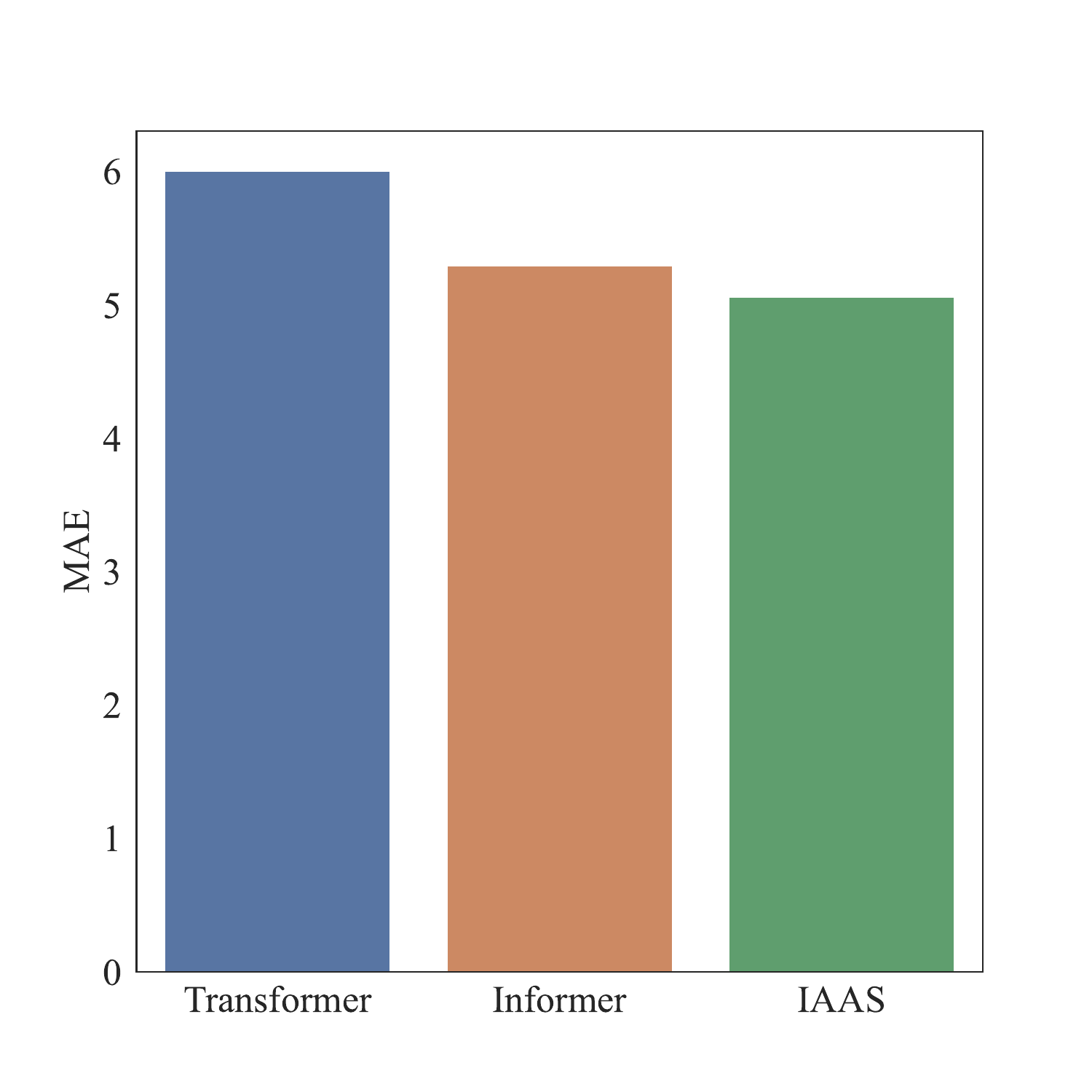}
        \caption*{\footnotesize{(b) MAE}}
    \end{subfigure}
\caption{One-day ahead wind power forecasting accuracy results using Transformer, Informer, and IAAS. 
    \label{fig:seq2seq-WF1}}
    
\end{figure}

We use three-month data for training the Seq2Seq model and the rest nine-month data for testing. The rolling window scheme is used to dynamically update the training and testing data for the one-day ahead forecasting case. To reuse the knowledge learned in the previous window, we do not train the Transformer and Informer models from the scratch. Instead, we train them based on their optimized parameters from last window. For IAAS, we keep the initial settings same for the RL agents and net pool to search the neural structures for each rolling-window process. We design the Transformer and Informer structures through the trial-and-error simulation method, and the best structures of these two models both contain two encoder layers and two decoder layers with 512 model dimensions and 8 heads. Following the hyperparameter settings from Informer, which is designed for the time-series forecasting task, we train these two models for 1000 epochs using Adam optimizer, ReLU activation function, and RMSE loss function. Moreover, we use the grid search algorithm to determine the batch size from [128, 256, 512] and the weight decay rate from [0.1, 0.01, 0.001, 0.0001, 0.00001]. The overall testing forecasting accuracy results are presented in Figs. \ref{fig:seq2seq-ME} and \ref{fig:seq2seq-WF1}. As seen, in the load forecasting case, the IAAS model performs much better than Transformer and Informer regarding the three evaluation metrics; in the wind power forecasting case, the IAAS model has slightly better performance than Transformer and Informer models. All in all, the IAAS framework is likely to generate better electricity time-series forecasting models than the existing transformer models under the Seq2Seq setting. In other words, the IAAS framework is better for the practical applications in power system.


\begin{figure}[htb] 
    \begin{subfigure}[b]{0.32\textwidth}
        \centering
	    \includegraphics[width=\textwidth]{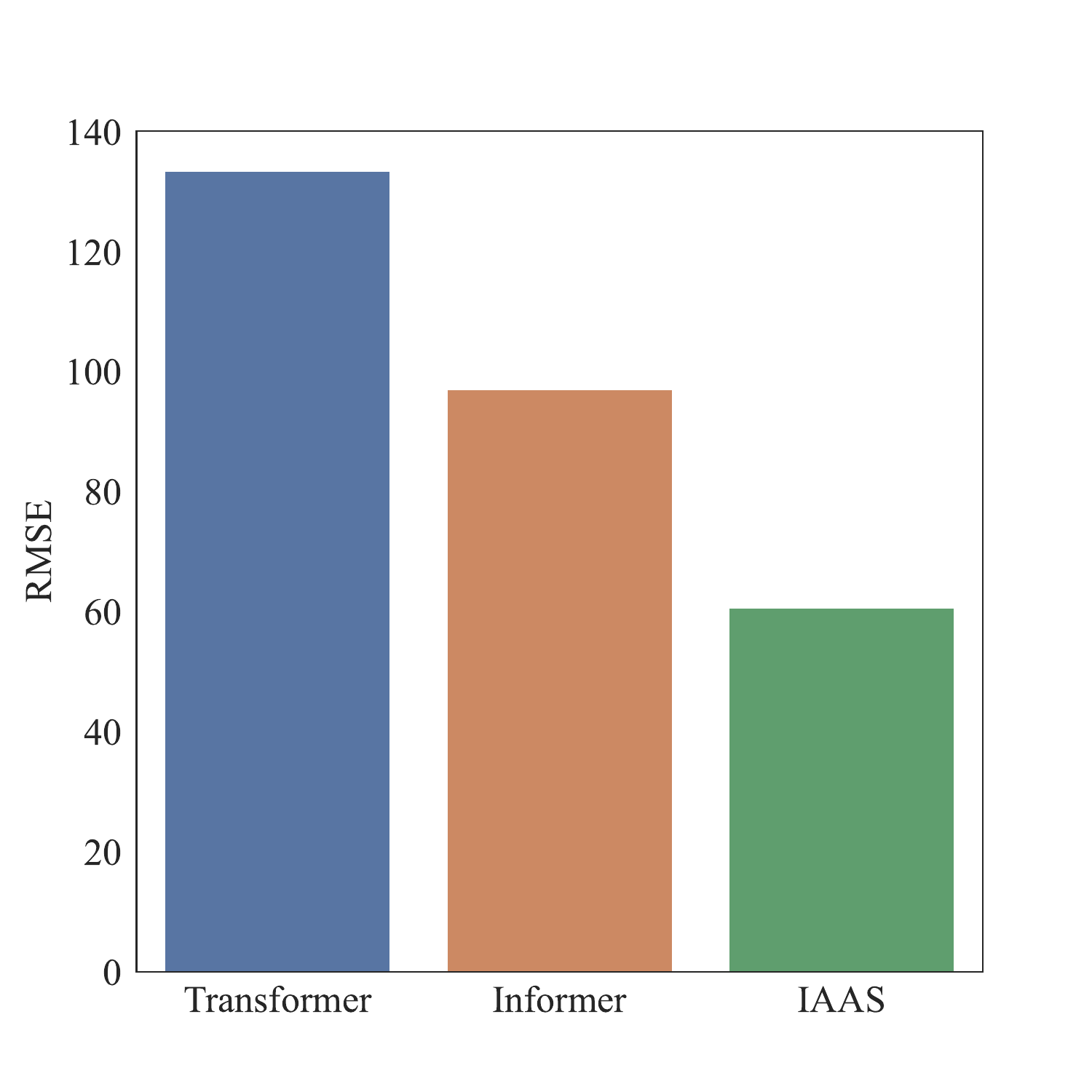}
        \caption*{\footnotesize{(a) RMSE}}
    \end{subfigure}
    \hfill
    \begin{subfigure}[b]{0.32\textwidth}
        \centering
		\includegraphics[width=\textwidth]{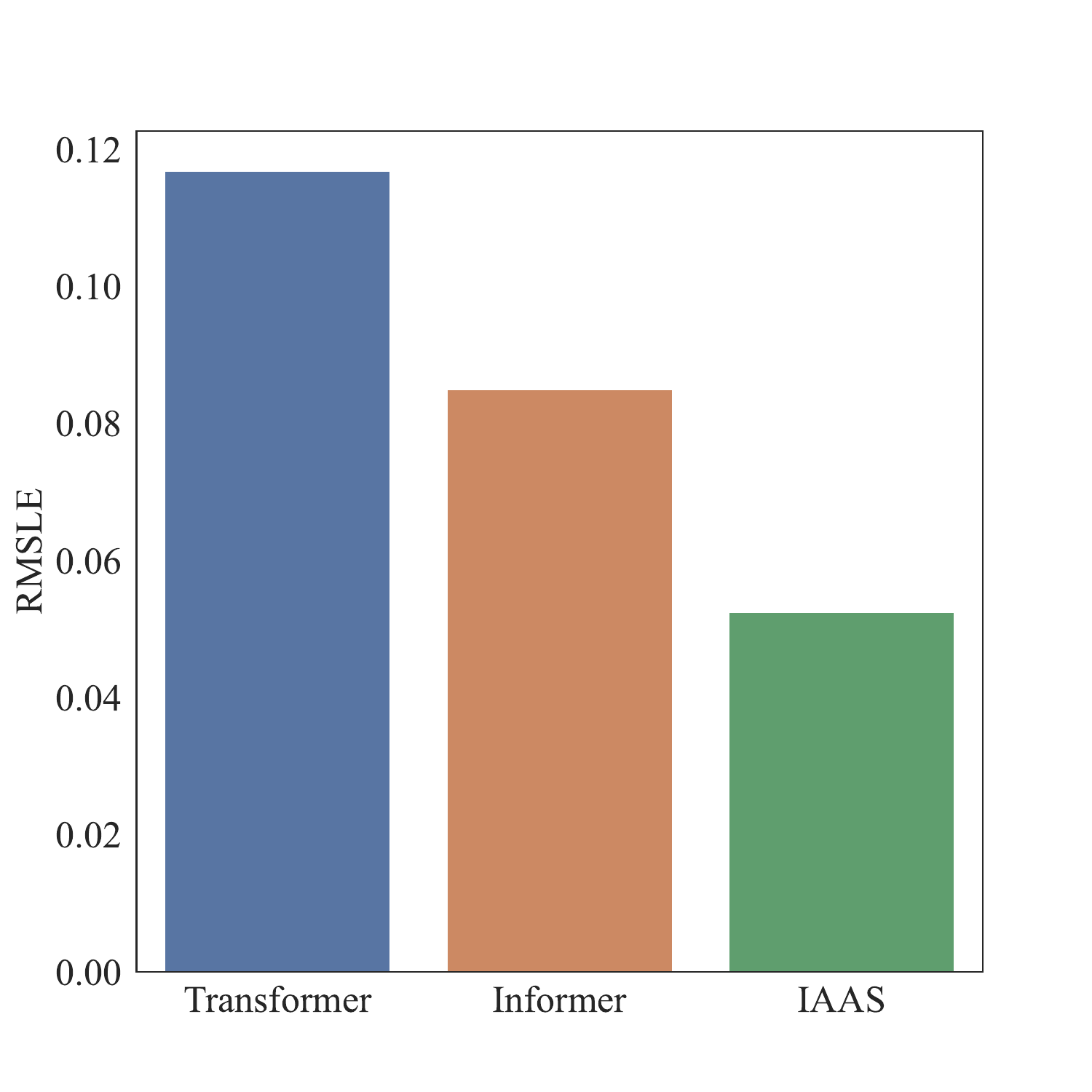}
        \caption*{\footnotesize{(b) RMSLE}}
    \end{subfigure}
    \hfill
    \begin{subfigure}[b]{0.32\textwidth}
        \centering
		\includegraphics[width=\textwidth]{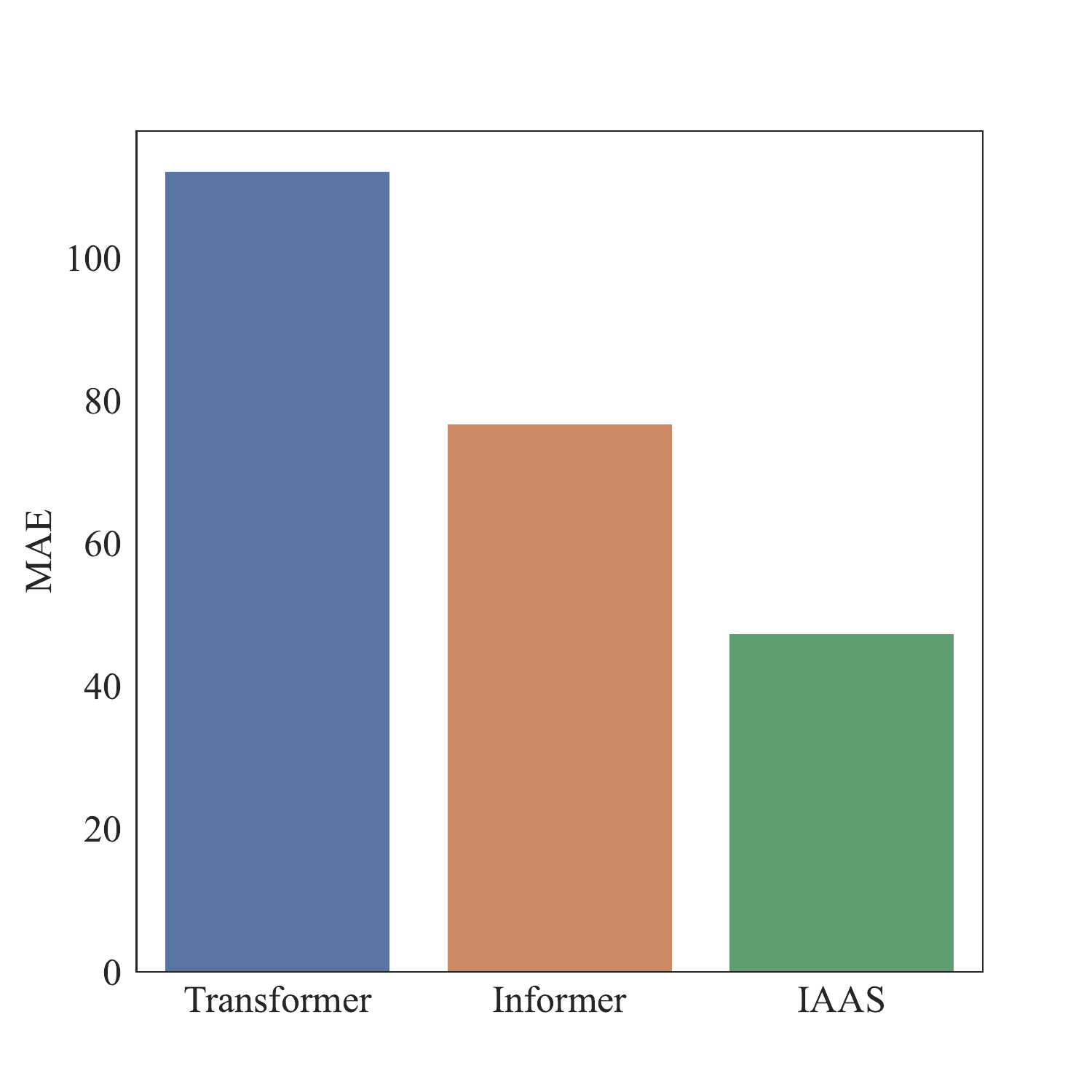}
        \caption*{\footnotesize{(C) MAE}}
    \end{subfigure}
\caption{One-week ahead load forecasting accuracy results under the Seq2Seq setting using Transformer, Informer, and IAAS. 
    \label{fig:seq2seq-ME1}}
\end{figure}

\begin{figure}[htb]\label{seq2seq wind}
    \centering
    \begin{subfigure}[b]{0.32\textwidth}
        \centering
	    \includegraphics[width=\textwidth]{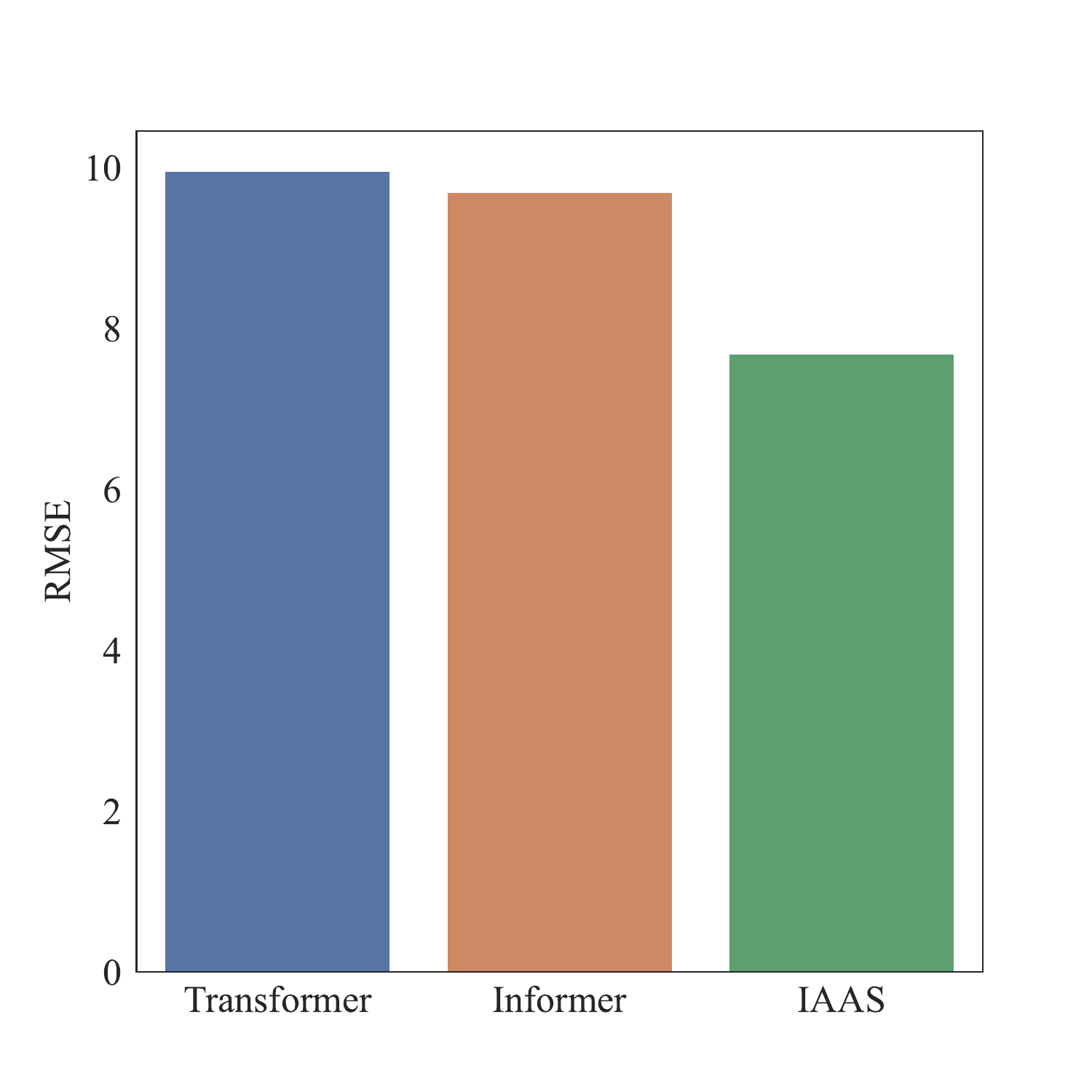}
        \caption*{\footnotesize{(a) RMSE}}
    \end{subfigure}
    \begin{subfigure}[b]{0.32\textwidth}
        \centering
	\includegraphics[width=\textwidth]{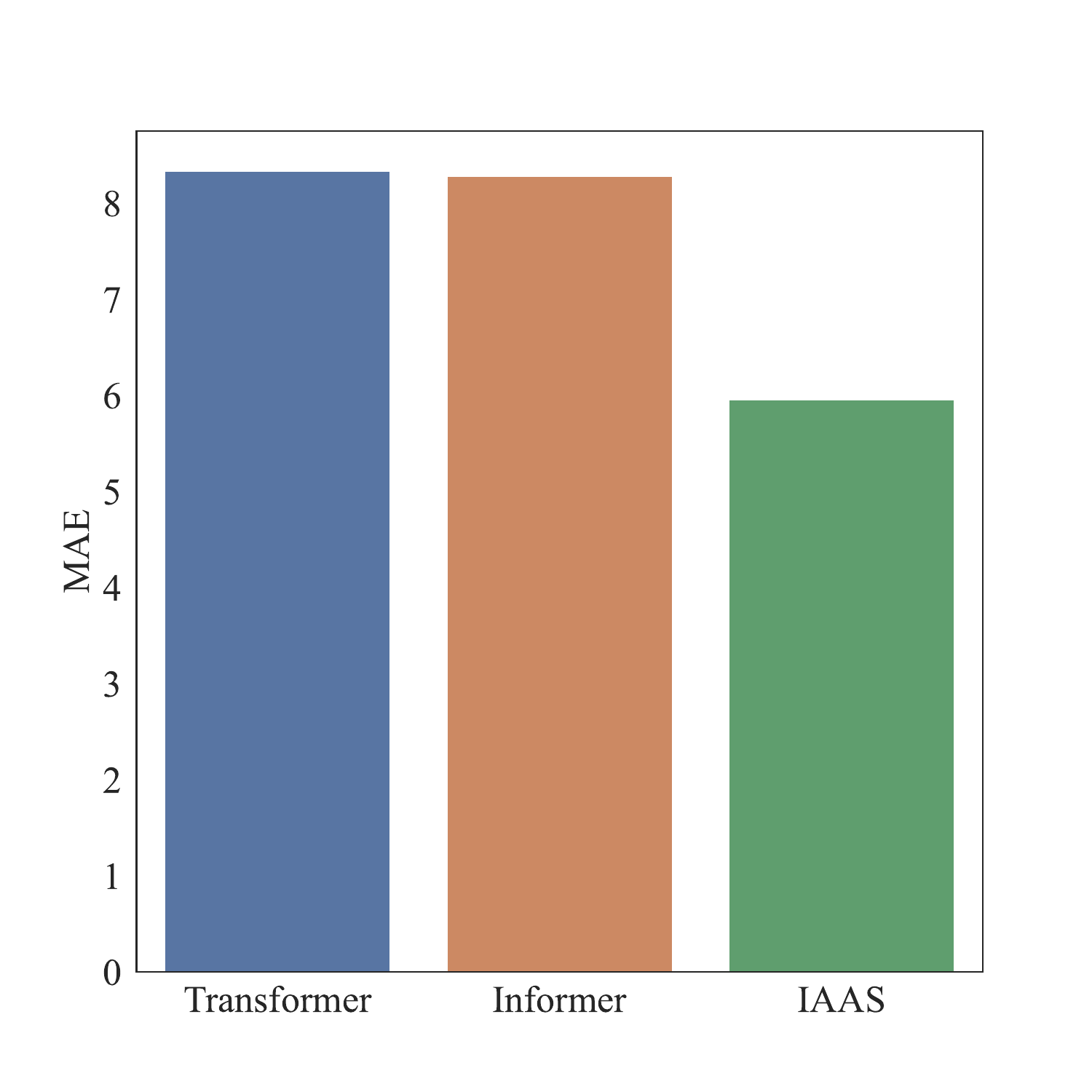}
        \caption*{\footnotesize{(b) MAE}}
    \end{subfigure}
\caption{One-week ahead wind power forecasting accuracy results using Transformer, Informer, and IAAS. 
    \label{fig:seq2seq-WF11}}
\end{figure}

The one-week ahead forecasting experiment is designed to use hourly historical data in the past 14 days (336 steps) to predict the data in the future seven days (168 steps). Hence, the input sequence sizes for the load and wind data are $336\times3$ and $336\times9$, respectively. The output sequence sizes for these two cases are both $168\times1$. Similar to one-day ahead forecasting experiment, we also use 3-month data for training but use the following 7 days for testing. We move the rolling window every seven days for the next round of training and testing. Therefore, for this case, we only obtain one forecasting sequence with the size of $168\times1$ for the corresponding week. In other words, for one-week ahead load forecasting, the Seq2Seq setting is $336\times3$ to $168\times1$, which is a many-to-many setting. Similarly, for one-week ahead wind power forecasting, the Seq2Seq setting is $336\times9$ to $168\times1$. This means that we predict the next one-week load or wind power using the previous two-week real data. For example, on Jan 1, we forecast the load or wind power in the following week (i.e., from Jan 2 to Jan 8). Hence, we will use the IAAS framework to search a Seq2Seq neural architecture for the load or wind power forecasting. This model will be only used once. On Jan 8, we update this Seq2Seq model by using IAAS with another search to forecast the load or wind power in the following week (i.e., from Jan 9 to Jan 15).  We put the experiment results in Figs. \ref{fig:seq2seq-ME1} and \ref{fig:seq2seq-WF11}. As seen, IAAS still achieves better accuracy in the week-ahead forecasting case than the other two transformer models under the seq2seq setting. All in all, the IAAS framework is likely to generate better electricity time-series forecasting models than the existing transformer models under the seq2seq setting. In other words, the IAAS framework is better for the practical applications in power systems.


\newpage

\section{Heuristic Mechanism Exploration}\label{appendix I}
In this section, we explore the heuristic mechanism with an ablation experiment to demonstrate how it works. Specifically, we develop three variants of IAAS framework considering this mechanism. 

The first one is without the heuristic approach, named IAAS\_h. Specifically, we initialize five networks in the net pool and transform them with RL. The five transformed networks will replace their parent networks in the net pool and go to the next-round transformation. Such an experiment is similar to running an EAS framework \citep{cai2018efficient}, which uses the transformed network directly for the next-round transformation. This setting follows the traditional paradigm of RL that uses equal fixed-length trajectories to collect the samples. 

The second one is without the random network generation in the heuristic algorithm, named IAAS\_r. Using this one, we can demonstrate the importance of randomly generating a network for each update of the net pool. Particularly, this setting in the heuristic algorithm is to provide more samples at the early stages of an MDP trajectory.  

The third one is based on the first one but randomly deleting five networks from the net pool. We name this one IAAS\_d. In particular, if there are five networks in the net pool, there will be ten networks in the net pool after transformation. At this time, we randomly eliminate five networks. Such a variant is a version after using a simple trajectory truncation strategy. With this, we can demonstrate whether our elimination strategy is appropriate.

As noted, all these three variants also use the same computational budget as ours, i.e., 1000 action times. Therefore, for the first variant, the length of each trajectory is 200. As to the second and third ones, due to the trajectory termination, the lengths of trajectories are not the same. We use the ME load dataset and the WF1 wind power dataset for this experiment. We present the results in the following table with eight scenarios. We then have the following findings:
\begin{enumerate}
    \item Comparing IAAS with IAAS\_h, our IAAS achieves the better performance in seven scenarios than IAAS\_h. This indicates the importance of the heuristic mechanism in IAAS, compared to the traditional method using fixed-length trajectories. 
    \item Comparing IAAS with IAAS\_r, our IAAS achieves better performance in seven scenarios than IAAS\_r. This demonstrates the importance of designing a random network generation component in the heuristic algorithm.
    \item Comparing IAAS with IAAS\_d, our IAAS achieves a better performance than IAAS\_d in all scenarios in terms of all metrics except the ME\_summer scenario. As seen, IAAS\_d achieves the lowest MAE value in ME\_summer scenario, but our IAAS has the better performance in this scenario than the other three variants regarding the RMSE and RMSLE values. This result clearly demonstrates the importance of the network elimination strategy in the heuristic mechanism.
   
\end{enumerate}

In addition to these three, we also find out that IAAS\_r performs better than IAAS\_h in five out of seven scenarios in terms of three metrics. As to the ME\_winter scenario, we believe IAAS\_r has a similar performance to IAAS\_h since in this scenario, IAAS\_r has a better RMSE value, IAAS\_h has a better MAE value, and these two have an equal RMSLE value. Although IAAS\_r does not have the component of randomly generating one network in each episode, this result also indicates the superiority of our proposed heuristic algorithm.

\begin{table}[ht]

\caption
{The ablation experiment result of heuristic mechanism exploration \label{tab:Heuristic-ablation}}
\centering
{
\small
\begin{tabular}{cccccc}
\hline
Scenario & Metrices & IAAS & IAAS\_h & IAAS\_r & IAAS\_d \\ \hline
\multirow{3}{*}{ME\_spring} & RMSE & 61.807 & 67.807 & \textbf{61.277} & 67.220 \\ \cline{2-6} 
 & MAE & 48.879 & 54.089 & \textbf{48.715} & 54.079 \\ \cline{2-6} 
 & RMSLE & \textbf{0.049} & 0.053 & \textbf{0.049} & 0.053 \\ \hline
\multirow{3}{*}{ME\_summer} & RMSE & \textbf{69.154} & 96.909 & 74.734 & 75.200 \\ \cline{2-6} 
 & MAE & 58.187 & 71.701 & 57.757 & \textbf{56.529} \\ \cline{2-6} 
 & RMSLE & \textbf{0.052} & 0.070 & 0.055 & 0.055 \\ \hline
\multirow{3}{*}{ME\_autumn} & RMSE & \textbf{38.483} & 42.983 & 39.765 & 43.342 \\ \cline{2-6} 
 & MAE & \textbf{29.526} & 33.153 & 29.925 & 32.892 \\ \cline{2-6} 
 & RMSLE & \textbf{0.031} & 0.034 & 0.032 & 0.035 \\ \hline
\multirow{3}{*}{ME\_winter} & RMSE & \textbf{85.386} & 97.573 & 97.071 & 102.469 \\ \cline{2-6} 
 & MAE & \textbf{68.773} & 77.845 & 80.431 & 82.645 \\ \cline{2-6} 
 & RMSLE & \textbf{0.062} & 0.071 & 0.071 & 0.075 \\ \hline
\multirow{2}{*}{WF1\_spring} & RMSE & \textbf{3.901} & 4.389 & 3.963 & 4.143 \\ \cline{2-6} 
 & MAE & \textbf{2.996} & 3.497 & 3.012 & 3.360 \\ \hline
\multirow{2}{*}{WF1\_summer} & RMSE & 3.290 & \textbf{3.064} & 3.381 & 3.149 \\ \cline{2-6} 
 & MAE & 2.614 & \textbf{2.204} & 2.553 & 2.313 \\ \hline
\multirow{2}{*}{WF1\_autumn} & RMSE & \textbf{2.237} & 2.680 & 2.633 & 2.798 \\ \cline{2-6} 
 & MAE & \textbf{1.633} & 1.983 & 1.882 & 2.124 \\ \hline
\multirow{2}{*}{WF1\_winter} & RMSE & \textbf{4.500} & 4.655 & 5.245 & 5.865 \\ \cline{2-6} 
 & MAE & \textbf{3.427} & 3.775 & 4.078 & 4.664 \\ \hline
\end{tabular}
}
{}
\end{table}

\clearpage
\section{Details of Ablation Studies}\label{appx:Ablation study details}

Observing Table \ref{tab:ablation-experiments}, IAAS\_LoadNet has better performance in all cases as compared to IAAS\_n\_LoadNet and IAAS\_sn\_LoadNet, in terms of the RMSE, MAE and RMSLE values. Averagely, IAAS\_LoadNet performs better than IAAS\_n\_LoadNet with 13.5\% (RMSE), 13.5\% (MAE), and 13.4\% (RMSLE), whereas IAAS\_LoadNet is also better than IAAS\_sn\_LoadNet with 15.3\% (RMSE), 14.8\% (MAE), and 14.7\% (RMSLE). In addition, IAAS\_LoadNet achieves a better performance than IAAS\_s\_LoadNet in six out of eight cases in terms of three evaluation metrics. However, observing the NH-summer and NH-autumn cases, the forecasting accuracies from IAAS\_LoadNet are close to those from IAAS\_s\_LoadNet. Averagely, IAAS\_LoadNet has a better performance than IAAS\_s\_LoadNet with 10.4\% (RMSE), 10.0\% (MAE) and 9.4\% (RMSLE).

\begin{table}[H]
\caption
{Load forecasting accuracy results in ablation experiments\label{tab:ablation-experiments}}
{
    \resizebox{\textwidth}{!}{
    \scriptsize{
\begin{tabular}{cccccc}
\hline
Case & Metrices & IAAS\_LoadNet & IAAS\_s\_LoadNet & IAAS\_n\_LoadNet & IAAS\_sn\_LoadNet \\ \hline
\multirow{3}{*}{ME-spring} & RMSE & \textbf{61.807} & 69.398 & 68.646 & 65.031 \\ \cline{2-6} 
 & MAE & \textbf{48.879} & 56.255 & 55.988 & 49.814 \\ \cline{2-6} 
 & RMSLE & \textbf{0.049} & 0.055 & 0.054 & 0.052 \\ \hline
\multirow{3}{*}{ME-summer} & RMSE & \textbf{69.154} & 93.573 & 83.095 & 83.754 \\ \cline{2-6} 
 & MAE & \textbf{58.187} & 70.889 & 67.327 & 64.173 \\ \cline{2-6} 
 & RMSLE & \textbf{0.052} & 0.067 & 0.061 & 0.06 \\ \hline
\multirow{3}{*}{ME-autumn} & RMSE & \textbf{38.483} & 61.95 & 46.891 & 43.282 \\ \cline{2-6} 
 & MAE & \textbf{30.526} & 47.82 & 35.718 & 33.441 \\ \cline{2-6} 
 & RMSLE & \textbf{0.031} & 0.049 & 0.038 & 0.035 \\ \hline
\multirow{3}{*}{ME-winter} & RMSE & \textbf{85.386} & 102.263 & 105.682 & 102.728 \\ \cline{2-6} 
 & MAE & \textbf{68.773} & 79.342 & 81.106 & 84.5 \\ \cline{2-6} 
 & RMSLE & \textbf{0.062} & 0.075 & 0.077 & 0.075 \\ \hline
\multirow{3}{*}{NH-spring} & RMSE & \textbf{68.749} & 73.36 & 80.832 & 80.54 \\ \cline{2-6} 
 & MAE & \textbf{53.759} & 59.191 & 63.534 & 62.833 \\ \cline{2-6} 
 & RMSLE & \textbf{0.058} & 0.062 & 0.067 & 0.068 \\ \hline
\multirow{3}{*}{NH-summer} & RMSE & 144.016 & \textbf{140.237} & 152.769 & 157.385 \\ \cline{2-6} 
 & MAE & 114.45 & \textbf{113.796} & 123.907 & 129.272 \\ \cline{2-6} 
 & RMSLE & 0.096 & \textbf{0.093} & 0.101 & 0.106 \\ \hline
\multirow{3}{*}{NH-autumn} & RMSE & 62.468 & \textbf{60.077} & 71.592 & 95.843 \\ \cline{2-6} 
 & MAE & 45.985 & \textbf{45.005} & 54.309 & 69.45 \\ \cline{2-6} 
 & RMSLE & 0.052 & \textbf{0.049} & 0.06 & 0.08 \\ \hline
\multirow{3}{*}{NH-winter} & RMSE & \textbf{82.911} & 83.442 & 98.854 & 95.489 \\ \cline{2-6} 
 & MAE & \textbf{65.506} & 67.772 & 80.264 & 76.723 \\ \cline{2-6} 
 & RMSLE & \textbf{0.062} & \textbf{0.062} & 0.074 & 0.072 \\ \hline
\multirow{3}{*}{Average} & RMSE & \textbf{76.622} & 85.537 & 88.545 & 90.506 \\ \cline{2-6} 
 & MAE & \textbf{60.758} & 67.509 & 70.269 & 71.276 \\ \cline{2-6} 
 & RMSLE & \textbf{0.058} & 0.064 & 0.067 & 0.068 \\ \hline
\end{tabular}

}
}
}
{}
\end{table}

Observing Table \ref{tab:wind-power-ablation-experiments}, IAAS\_WindNet also provides much better performance than IAAS\_n\_WindNet and IAAS\_sn\_WindNet in terms of the forecasting accuracy. On average, IAAS\_WindNet obtains the better forecasting accuracy than IAAS\_n\_WindNet with 16.7\% (RMSE) and 17.2\% (MAE), and obtains the better forecasting accuracy than IAAS\_sn\_WindNet with 17.5\% (RMSE) and 18.6\% (MAE). Moreover, IAAS\_WindNet performs better than IAAS\_s\_WindNet in most of cases regarding the two evaluation metrics. On average, IAAS\_WindNet performs better than IAAS\_s\_WindNet with 2.4\% (RMSE) and 2.4\% (MAE).

\begin{table}
\caption
{Wind power forecasting accuracy results in ablation experiments\label{tab:wind-power-ablation-experiments}}
{
    \resizebox{\textwidth}{!}{
    \scriptsize{
\begin{tabular}{cccccc}
\hline
Case & Metrices & IAAS\_WindNet & IAAS\_s\_WindNet & IAAS\_n\_WindNet & IAAS\_sn\_WindNet \\ \hline
\multirow{2}{*}{WF1-spring} & RMSE & \textbf{3.901} & 3.953 & 4.845 & 4.473 \\ \cline{2-6} 
 & MAE & \textbf{2.996} & 3.095 & 3.731 & 3.402 \\ \hline
\multirow{2}{*}{WF1-summer} & RMSE & 3.29 & \textbf{3.148} & 3.649 & 3.866 \\ \cline{2-6} 
 & MAE & 2.614 & \textbf{2.286} & 2.754 & 3.246 \\ \hline
\multirow{2}{*}{WF1-autumn} & RMSE & \textbf{2.237} & 2.411 & 2.658 & 3.138 \\ \cline{2-6} 
 & MAE & \textbf{1.633} & 1.817 & 1.911 & 2.298 \\ \hline
\multirow{2}{*}{WF1-winter} & RMSE & \textbf{4.5} & 4.555 & 5.291 & 5.249 \\ \cline{2-6} 
 & MAE & \textbf{3.427} & 3.582 & 4.331 & 4.146 \\ \hline
\multirow{2}{*}{WF2-spring} & RMSE & \textbf{4.922} & 5.147 & 5.762 & 5.967 \\ \cline{2-6} 
 & MAE & \textbf{3.953} & 4.027 & 4.147 & 4.704 \\ \hline
\multirow{2}{*}{WF2-summer} & RMSE & 4.87 & \textbf{4.854} & 5.193 & 5.077 \\ \cline{2-6} 
 & MAE & \textbf{3.591} & 3.774 & 4.095 & 4.067 \\ \hline
\multirow{2}{*}{WF2-autumn} & RMSE & 6.031 & \textbf{5.824} & 7.655 & 7.22 \\ \cline{2-6} 
 & MAE & 4.439 & \textbf{4.386} & 6.164 & 5.284 \\ \hline
\multirow{2}{*}{WF2-winter} & RMSE & \textbf{3.104} & 3.777 & 4.287 & 4.84 \\ \cline{2-6} 
 & MAE & \textbf{2.402} & 2.711 & 3.136 & 3.651 \\ \hline
\multirow{2}{*}{Average} & RMSE & \textbf{4.107} & 4.209 & 4.917 & 4.979 \\ \cline{2-6} 
 & MAE & \textbf{3.132} & 3.210 & 3.784 & 3.850 \\ \hline
\end{tabular}

}
}
}
{}
\end{table}

\clearpage
\section{Further Discussion}\label{appx:Further Discussion}
The IAAS framework essentially contains two basic operations (i.e., wider transformation and deeper transformation) for network enlargement, and one pruning operation to shrink the network. Literature works usually use some other techniques in the forecasting model development, such as residual connections in \cite{chen2018short} and dropout strategy in \cite{hossain2021forecasting}. In this section, we will discuss whether these techniques can be integrated with IAAS. 

In IAAS, we limit the use of the ReLU activation function considering the identity mapping. However, other activation functions (e.g., SeLU and Sigmoid) can also be applied if we use residual connections and set the output of the newly inserted layer to produce a zero tensor as the output. In other words, we can also use residual connections in IAAS for the model performance improvement. Note that in EAS, \cite{cai2018efficient} has already used residual connections in NAS. Moreover, we can also consider the dropout strategy in IAAS since it does not conflict with the network enlargement and shrinkage transformations. Furthermore, we can add a regularization term to the loss function for the model generalization improvement and exploit the early stopping strategy instead of the fixed number of episodes to shorten the computational time. Note that pruning, dropout, regularization, and early stopping techniques can all mitigate the overfitting issues. Therefore, one of the future research areas may consider the integration of these four techniques to address the problem of overfitting in NAS. 

In addition to these techniques, we can also consider other network structures as search candidates. In this study, we exploit the RNN and LSTM to extract features from the electricity time-series data. However, the LSTM we use is variant of the vanilla LSTM \citep{graves2005framewise}, named ``no-peepholes" (NP) LSTM. Besides NP LSTM, \cite{greff2016lstm} also derived the other seven variants (e.g., no-input-gate LSTM, no-forget-gate LSTM, etc.) by separately deleting one component from the vanilla LSTM. Since the main structures of these LSTM are similar, the proposed function-preserving network transformation and the movement pruning can be applied to all of them with minor changes. For example, when using vanilla LSTM, we need to initialize the parameters of peepholes at the transformation process since our formulation in Section 3 is based on NP LSTM. In the future, we will consider these existing techniques and the LSTM variants in IAAS so that we can build a more accurate forecasting model for power systems.

\end{document}